\documentclass[12pt,a4paper,notitlepage]{article}

\usepackage[section]{placeins}
\usepackage{amsfonts}
\usepackage{amsmath}
\usepackage{amssymb}
\usepackage{mathtools}
\usepackage{amsthm}
\usepackage{bm}
\usepackage{color}
\usepackage[T1]{fontenc}
\usepackage{graphicx,psfrag}
\usepackage{hyperref}
\usepackage{mathrsfs}
\usepackage{multirow}
\usepackage{multicol}
\usepackage[square]{natbib}
\usepackage[format=hang]{subcaption}
\usepackage{times}
\usepackage{upgreek}
\usepackage[table]{xcolor}
\AtBeginDocument{}

\date{}

\graphicspath{{./images/}}

\usepackage{pifont}
\makeatletter
\newcommand\Pimathsymbol[3][\mathord]{%
  #1{\@Pimathsymbol{#2}{#3}}}
\def\@Pimathsymbol#1#2{\mathchoice
  {\@Pim@thsymbol{#1}{#2}\tf@size}
  {\@Pim@thsymbol{#1}{#2}\tf@size}
  {\@Pim@thsymbol{#1}{#2}\sf@size}
  {\@Pim@thsymbol{#1}{#2}\ssf@size}}
\def\@Pim@thsymbol#1#2#3{%
  \mbox{\fontsize{#3}{#3}\Pisymbol{#1}{#2}}}
\makeatother
\input{utxmia.fd}
\DeclareFontShape{U}{txmia}{m}{n}{<->ssub * txmia/m/it}{}
\DeclareFontShape{U}{txmia}{bx}{n}{<->ssub * txmia/bx/it}{}

\usepackage{tikz}
\pgfrealjobname{iCANN} 
\usepackage{tikz-3dplot}
\usetikzlibrary{patterns,decorations.pathmorphing}
\usepackage{environ}
\makeatletter
\newsavebox{\measure@tikzpicture}
\NewEnviron{scaletikzpicturetowidth}[1]{%
  \def\tikz@width{#1}%
  \begin{lrbox}{\measure@tikzpicture}%
  \BODY
  \end{lrbox}%
  \pgfmathparse{#1/\wd\measure@tikzpicture}%
  \BODY
}
\makeatother
\usetikzlibrary{positioning}

\usepackage{pgfplots}
\usepgfplotslibrary{groupplots}
\usepackage{xcolor}

\definecolor{rwth1}{RGB}{0,84,159}
\definecolor{rwth2}{RGB}{246,168,0}
\definecolor{rwth3}{RGB}{0,97,101}
\definecolor{rwth4}{RGB}{204,7,30}
\definecolor{rwth5}{RGB}{87,171,39}
\definecolor{rwth6}{RGB}{161,16,53}
\definecolor{rwth7}{RGB}{0,152,161}
\definecolor{rwth8}{RGB}{122,111,172}

\definecolor{red}{HTML}{FF0000}
\definecolor{orange}{HTML}{FF8C00}
\definecolor{yellow}{HTML}{FFD700}
\definecolor{green}{HTML}{009900}
\definecolor{cyan}{HTML}{00CED1}
\definecolor{blue}{HTML}{0000CD}
\definecolor{purple}{HTML}{9400D3}
\definecolor{black}{HTML}{000000}

\tikzstyle{dashpattern0} = [dash pattern = ]
\tikzstyle{dashpattern1} = [dash pattern = on 4.25pt off 0.75pt]
\tikzstyle{dashpattern2} = [dash pattern = on 1.5pt off 0.5pt]
\tikzstyle{dashpattern3} = [dash pattern = on 0.75pt off 0.4pt]
\tikzstyle{dashpattern4} = [dash pattern = on 3pt off 1pt on 1pt off 1pt]
\tikzstyle{dashpattern5} = [dash pattern = on 3.75pt off 0.5pt on 0.75pt off 0.5pt on 0.75pt off 0.5pt]
\tikzstyle{dashpattern6} = [dash pattern = on 3.25pt off 0.5pt on 0.75pt off 0.5pt on 0.75pt off 0.5pt on 0.75pt off 0.5pt]
\tikzstyle{dashpattern7} = [dash pattern = on 3.25pt off 0.5pt on 0.75pt off 0.5pt on 0.75pt off 0.5pt on 0.75pt off 0.5pt on 0.75pt off 0.5pt]
\tikzstyle{dashpattern8} = [line cap=round, dash pattern = on 3.25pt off 2.75pt]
\tikzstyle{dashpattern9} = [line cap=round, dash pattern = on 0.01pt off 2pt]
\tikzstyle{dashpattern10}= [line cap=round, dash pattern = on 3.25pt off 2pt on 0.01pt off 2pt]
\tikzstyle{dashpattern11}= [line cap=round, dash pattern = on 3.5pt off 1.75pt on 0.01pt off 1.75pt on 0.01pt off 1.75pt]
\tikzstyle{dashpattern12}= [line cap=round, dash pattern = on 3.5pt off 1.75pt on 0.01pt off 1.75pt on 0.01pt off 1.75pt on 0.01pt off 1.75pt]
\tikzstyle{dashpattern13}= [line cap=round, dash pattern = on 3.5pt off 1.75pt on 0.01pt off 1.75pt on 0.01pt off 1.75pt on 0.01pt off 1.75pt on 0.01pt off 1.75pt]

\pgfplotscreateplotcyclelist{rwth}{
{rwth1},{rwth2},{rwth3},{rwth4},{rwth5},{rwth6},{rwth7},{rwth8}
}
\pgfplotscreateplotcyclelist{rwthLine}{
{rwth1, dash pattern={on 5pt off 2pt}},{rwth2, dash pattern={on 3pt off 1pt}},{rwth3, dash pattern={on 1pt off 1pt}},{rwth4, dash pattern={on 5pt off 1pt on 1pt off 1pt}},{rwth5, dash pattern={on 3pt off 1pt on 1pt off 1pt}},{rwth6, dash pattern={on 5pt off 1pt on 1pt off 1pt on 1pt off 1pt}},{rwth7, dash pattern={on 3pt off 1pt on 1pt off 1pt on 1pt off 1pt}},{rwth8}
}
\pgfplotscreateplotcyclelist{TBcolor}{
{red, dash pattern={on 5pt off 2pt}},{orange, dash pattern={on 3pt off 1pt}},{yellow, dash pattern={on 1pt off 1pt}},{green, dash pattern={on 5pt off 1pt on 1pt off 1pt}},{cyan, dash pattern={on 3pt off 1pt on 1pt off 1pt}},{blue, dash pattern={on 5pt off 1pt on 1pt off 1pt on 1pt off 1pt}},{purple, dash pattern={on 3pt off 1pt on 1pt off 1pt on 1pt off 1pt}},{black}
}
\pgfplotscreateplotcyclelist{HCcolor}{
{red, dash pattern={on 5pt off 2pt}},{orange, dash pattern={on 3pt off 1pt}},{blue, dash pattern={on 5pt off 1pt on 1pt off 1pt}},{purple, dash pattern={on 3pt off 1pt on 1pt off 1pt}},{green, dash pattern={on 5pt off 1pt on 1pt off 1pt on 1pt off 1pt}},{black}
}

\DeclareMathOperator{\trace}{tr}
\newcommand{\tr}[1]{{\trace \left(#1 \right)}}
\DeclareMathOperator{\determinante}{det}
\newcommand{\dete}[1]{{\determinante \left(#1 \right)}}
\DeclareMathOperator{\symmetricpart}{sym}
\newcommand{\symp}[1]{{\symmetricpart \left(#1 \right)}}

\DeclareMathOperator{\deviatoric}{dev}
\newcommand{\dev}[1]{{\deviatoric\left(#1 \right)}}

\newcommand{\pfrac}[2]{{\frac{\partial #1}{\partial #2}}}

\newboolean{color_white}
\setboolean{color_white}{true}
\newcommand{\rb}[1]{\ifthenelse{\boolean{color_white}}
{\textcolor{blue}{#1}}
{#1}}

\usepackage{algorithm}
\usepackage{algpseudocode}
\algnewcommand{\LineComment}[1]{\State \(\triangleright\) #1}

\makeatletter
\renewcommand*\env@matrix[1][c]{\hskip -\arraycolsep
  \let\@ifnextchar\new@ifnextchar
  \array{*\c@MaxMatrixCols #1}}
\makeatother

\begin{document}

\author{\large {Hagen Holthusen$^1$, Lukas Lamm$^1$, Tim Brepols$^1$, Stefanie Reese$^1$}\\[0.25mm]
{and Ellen Kuhl$^2$}\\[0.25mm]
\hspace*{-0.1cm}
\normalsize{\em $^1$Institute of Applied Mechanics, RWTH Aachen
  University,}\\[-0.1cm]
\normalsize{\em Mies-van-der-Rohe-Str. 1, 52074 Aachen, Germany}\\[0.25cm]
\normalsize{\em $^2$Department of Mechanical Engineering, Stanford University,}\\[-0.1cm]
\normalsize{\em Stanford, CA 94305, United States}\\[0.25cm]
}

\title{\LARGE Theory and implementation of\\inelastic Constitutive Artificial Neural Networks}

\maketitle

%
%

\vspace{-1cm}
\small
{\bf Abstract}\\
Nature has always been our greatest inspiration in the research, design and development of novel materials and has driven us to gain a deep understanding of the underlying mechanisms that characterize anisotropy and inelastic behavior.
All this knowledge has been accumulated in the principles of thermodynamics.
Deduced from these principles, the multiplicative decomposition combined with pseudo potentials are powerful and universally valid concepts for inelastic materials.
Simultaneously, the tremendous increase in computational performance enabled us to investigate and rethink our history-dependent material models to make the most of our predictions.
Today, we have reached a point where materials and their models are becoming increasingly sophisticated.
This raises the question: How do we find the best model that includes all inelastic effects to explain our complex data? 
Constitutive Artificial Neural Networks (CANN) may answer this question.
Here, we extend the CANNs to inelastic materials (iCANN).
Rigorous considerations of objectivity, rigid motion of the reference configuration, multiplicative decomposition and its inherent non-uniqueness, choice of appropriate stretch tensors, restrictions of energy and pseudo potential, and consistent inelastic evolution guide us towards the general architecture of the iCANN satisfying thermodynamics per design.
We combine feed-forward networks of the Helmholtz free energy and pseudo potential with a recurrent neural network approach to take time dependencies into account.
Specializing the general iCANN to visco-elasticity, we demonstrate that the iCANN is capable of autonomously discovering models for artificially generated data, the response of polymers at different stretch rates for cyclic loading as well as the relaxation behavior of muscle data.
As the design of the network is not limited to visco-elasticity, our vision is that the iCANN will reveal to us new ways to find the various inelastic phenomena hidden in the data and to understand their interaction.
Our source code, data, and examples are available at \cite{iCANN_code} (\url{https://doi.org/10.5281/zenodo.10066805}).

\vspace*{-0.1cm}


\normalsize

\newpage
\section*{Nomenclature}
\begingroup
\allowdisplaybreaks
\begin{alignat*}{4}
&a, A &&\text{Scalar} &&\text{$\tr{\bm{A}}$} &&\text{Trace of $\bm{A}$} \\
&\text{$\bm{a}$} &&\text{First order tensor} \quad &&\text{$\dete{\bm{A}}$} &&\text{Determinant of $\bm{A}$} \\
&\text{$\bm{A}$} &&\text{Second order tensor} \quad &&\text{$\dev{\bm{A}}$} &&\text{$\bm{A}-\frac{\tr{\bm{A}}}{3}\bm{I}$} \\
&\text{$\bm{I}$} &&\text{Identity tensor} \quad &&\text{$\exp\left(\bm{A}\right)$} \qquad &&\text{Exponential of $\bm{A}$} \\
&\text{$\mathrm{SO}(\bullet)$} \qquad &&\text{Special orthogonal group} \qquad &&\text{$I_1^{\bm{A}}$} &&\text{$\tr{\bm{A}}$} \\
&\text{$\bm{A}^T$} &&\text{Transpose of $\bm{A}$} \quad &&\text{$I_2^{\bm{A}}$} &&\text{$\frac{1}{2}\left(\tr{\bm{A}}^2-\tr{\bm{A}^2}\right)$} \\
&\text{$\bm{A}^{-1}$} &&\text{Inverse of $\bm{A}$} \quad &&\text{$I_3^{\bm{A}}$} &&\text{$\dete{\bm{A}}$} \\
&\text{$\mathrm{sym}\left(\bm{A}\right)$} \qquad &&\text{$\frac{1}{2}\left(\bm{A}+\bm{A}^T\right)$} &&\text{$J_2^{\bm{A}}$} &&\text{$\frac{1}{2}\,\tr{\dev{\bm{A}}^2}$} \\
&\phantom{a} &&\phantom{a} &&\text{$J_3^{\bm{A}}$} &&\text{$\frac{1}{3}\,\tr{\dev{\bm{A}}^3}$}
\end{alignat*}
\endgroup
\section{Motivation}
\label{sec:motivation}
Identifying the relationship between inputs and outputs of complex data sets is often tedious. 
Artificial neural networks are a promising solution to this challenge.
Inspired by how the brain is solving problems, the architecture of neural networks is designed to learn these relations by extracting information from the data provided.
Much like the human brain, neural networks are able to form more neurons as we provide more data; and thus, understand the input-output dependency even better.
However, in contrast to a neural network, we -- as humans -- do not start from scratch when learning something new.
The way we learn -- and more importantly, our ability to predict future outcomes based on what we have learned -- is shaped by a powerful characteristic of our brain that we constantly benefit from: Experience.
Experience allows us to intuitively extrapolate and, one day, may become knowledge. 
In a mechanical context, the study of materials and their behavior has led us to the universal laws of thermodynamics.
As classical neural networks lack from this knowledge, it is not surprising that they may learn the relation between strains and stresses well, but fail to predict the material behavior outside the training regime.
This is already the case for hyperelastic materials (see e.g. \cite{linka2023}) where the
stress strain relation is bijective, and is even worse for inelastic
materials with a non-unique stress strain relation
(see e.g. \cite{wang2023}). Another issue of traditional neural
  networks is that they typically require a rather huge amount of solution data of the
  underlying problem for the training process. This can be an obstacle
  whenever the acquisition of data from real experiments or realistic micromechanical
  simulations is either not feasible or too costly.

For this reason, neural networks were developed in
  which -- in one way or the other -- prior knowledge about the
  underlying physics of the problem is incorporated. This is done with
  the aim of requiring less data and improving the predictive capabilities of the trained network. One prominent approach in this direction are Physics-Informed Neural Networks, also called
  PINNs (\cite{RaissiPerdikarisEtAl2019}). PINNs\footnote{For simplicity, other neural network approaches based on very similar ideas will also be referred to by this abbreviation, even if
  they are not explicitly named as such in the corresponding papers.} incorporate physical constraints to the formulation by adding additional terms to the loss
  function. This strategy facilitates the training process by
  narrowing the admissible solution space and assists the optimizer in
  finding a network behaving physically more
  reasonably, i.e.\ showing better extrapolation abilities. Typically, PINNs
  are used for solving partial differential equations, i.e.\ whole (initial-)boundary value
  problems, as an alternative to, e.g., the finite element
  method (for several recent examples in the context of solid mechanics, see
  \cite{RaoSunEtAl2020}, \cite{ZhangYinEtAl2020},
  \cite{AminiNiakiHaghighatEtAl2021}, \cite{CaiWangEtAl2021},
  \cite{HaghighatRaissiEtAl2021}, \cite{VahabHaghighatEtAl2021},
  \cite{HenkesWesselsEtAl2022}, \cite{RezaeiHarandiEtAl2022},
  \cite{HarandiMoeineddinEtAl2023}, \cite{NiuZhangEtAl2023}). Much less
  frequently, PINNs are also used to replace constitutive
  models. For instance, \cite{HaghighatAboualiEtAl2023} present a PINN framework which is able to characterize and
  discover yield surfaces and flow rules in geometrically linear
  elasto-plasticity, showing how to deal with complex inequality
  constraints in the loss function. \cite{EghbalianPouraghaEtAl2023}
  develop a surrogate model approximating classical elasto-plastic constitutive relations that can be used for large
  deformations via integration into a hypoelastic-based plasticity
  framework. However, it can generally be criticized
  that in PINN-like approaches physical laws are only
  weakly enforced during the `offline' training phase and not anymore during `online' computations. For this reason, outputs of the network may
  still violate basic physical principles.

  Going in an entirely different direction, so-called (material) model-free approaches have been
  developed which completely
  dispense with the formulation of constitutive models and their
  characterization for solving boundary value problems,
  relying instead only on experimental data 
  (\cite{KirchdoerferOrtiz2016}, \cite{IbanezBorzacchielloEtAl2017}). Such purely
  data-driven methods have the clear advantage that only minimal and
  generally accepted laws (such as basic kinematic relationships and
  force equilibrium in a mechanical context) are assumed as given, but no
  complicated (and possibly error-prone) constitutive equations need
  to be formulated at all. On the downside, the large amount of data required for the
  success of such methods can be disadvantageous -- a
  property shared with traditional neural
  network approaches. These methods have already been extended to
  inelasticity (see e.\,g.\ \cite{EggersmannKirchdoerferEtAl2019}, \cite{IbanezAbisset-ChavanneEtAl2018}). Nevertheless, their successful applicability
  in the case of complex multi-dimensional and history-dependent problems or problems with
  non-unique solutions still needs further clarification, the reason
  of which they are subject of intensive research in various
  directions (see e.\,g.\ \cite{EggersmannStainierEtAl2021},
  \cite{CiftciHackl2022}, \cite{ZschockeLeichsenringEtAl2022}, \cite{KuangBaiEtAl2023}, to name only a few).

  As a very promising alternative to PINNs described above,
  neural networks have been developed that enforce physical constraints in a
  strict sense even in `online' computations. This can be realized by encoding them a priori into the network architecture
  itself, as done e.\,g.\ by the Thermodynamics-based Artificial Neural
  Networks developed by \cite{MasiStefanouEtAl2021}. Another way of proceeding is to work
  with traditional neural networks in the first place but to enforce
  the constraints a posteriori, as suggested e.\,g.\ by
  \cite{KalinaLindenEtAl2022}. In any case, the outputs of such networks do by design
  fulfill those natural requirements that we have been placing on
  classically formulated material models ever since, as e.g.\ the second law of thermodynamics. These approaches are therefore very much in line with our goal of not wanting to
  design neural networks that completely discard all the knowledge
  of physical and thermodynamical principles having been accumulated over
  centuries. Instead, it is our strong belief that we should better teach networks
  the mentioned principles already beforehand, such that they can
  solely concentrate during training on revealing all the interesting
  (possibly inelastic) material phenomena for us in the given data.

  Recently, these considerations have led to a most fascinating
  development, which may lead to a true paradigm shift
  in constitutive modeling in the future. The basic idea is as follows: instead of laboriously deducing which specific phenomena are actually
  taking place in a material by manually analyzing experimental data and setting up a corresponding biased material
  model (which is likely either too simplistic or too
  complex, depending on the developer's level of experience), a
  constitutive model is automatically discovered that optimally only
  reflects the material phenomena truly present in the
  data. One representative following this idea is the framework EUCLID
  developed by \cite{FlaschelKumarEtAl2021, FlaschelKumarEtAl2022}. Basically, EUCLID is an intelligent
  algorithm using unlabeled data and sparse regression techniques to select
  and link only those candidates from a large pre-formulated library
  of constitutive models (generalized standard materials) which are
  actually necessary to describe the data. In \cite{FlaschelKumarEtAl2023}, it has been shown (for the case of small deformations) that such an
  approach can be used to successfully identify a model for a material
  whose material class was previously completely unknown.

A somehow related, but in its implementation completely different and
even more general strategy in this regard is pursued by Constitutive Artificial
Neural Networks, also known as CANNs
(\cite{linka2021}, \cite{abdolazizi2023arxiv}, \cite{LinkaPierreEtAl2023}). In the
first place, these are neural networks in which kinematical,
thermodynamical, and physical constraints (as given by the second law of thermodynamics,
material objectivity, material symmetry, etc.) are strictly
incorporated. Additionally, and in contrast to most other neural
networks in the literature, CANNs aim to formulate a physically
reasonable, yet extremely general expression for the Helmholtz free
energy by an intelligent construction of the network
architecture. Consequently, through appropriate training, the network
will autonomously recognize which components of this parameterized
energy are actually needed to describe the data. The main novelty is
that, in special cases, the network will be able to exactly represent
known constitutive models from the literature; in general, however, it
will generate a previously completely unknown material model that describes the
data even more accurately and leads to better predictions (\cite{linka2023}). Important in this context is also the fact that the network
learns a set of physically interpretable parameters, which is clearly advantageous.   

The aforementioned idea has taken the community by storm. Since the first publication on Constitutive
Artificial Neural Networks less than three years ago (\cite{linka2021}),
dozens of research groups have adopted and refined the initial method
to align with general physical requirements and thermodynamic
constraints. 
Constitutive Artificial Neural Networks for hyperelastic materials are now relatively well understood for both isotropic and anisotropic materials. 
In fact, the approach is so common by now that it has even been integrated directly into the finite element workflow to translate into engineering practice (see \cite{peirlinck2024}). 
However, a critical missing link is to expand the general concept to inelastic materials.

Today, there are nearly as many approaches to model inelastic material behavior as there are inelastic phenomena. 
Some of them are only useful for individual effects, such as visco-elasticity.
Therefore, we should not ask ourselves how we can fuse all these approaches into one model; instead, what is the most general approach?
What are the essential steps to describe inelastic effects?
Once these general mechanisms are found, the question of inelasticity, e.g. plasticity, is merely a question of specializing the approach and not a question of changing the model.
For solids, we identify the concepts of internal state variables (\cite{rice1971}) and pseudo/dissipation potentials of stress (\cite{kerstin1969}) as universal and perfectly complementary approaches.
The former concept allows to distinguish between elastic (reversible) strains associated with the amount of elastically stored energy and inelastic (irreversible) strains related to the dissipation.
In the finite strain theory, the assumption of a multiplicative decomposition of the deformation gradient into elastic and inelastic parts is advantageous, well-known, widely accepted in the continuum mechanics community, and able to capture a broad range of material behaviors (see \cite{kroener1959}, \cite{lee1969}, \cite{mandel1973}, \cite{sidoroff1974}, \cite{rodriguez1994}, \cite{lion1997}, \cite{haupt2001}, \cite{menzel2005dam}, \cite{bamann2010}, and \cite{latorre2015}, among many others).
It is worth noting that none of these parts generally satisfies the compatibility condition, so they should not be misunderstood as gradients.
The latter concept follows from sound thermodynamic considerations that are valid for rate-dependent (see e.g. \cite{lion1996a,lion1997b}, \cite{reese1998}, and \cite{besson2009}) and -independent (see e.g. \cite{miehe1998} and \cite{badreddine2010}) materials and can be easily extended to induced anisotropy (see e.g. \cite{hill1948} and \cite{benzerga2001}).
Similar to the Helmholtz free energy, which gives us the relationship
between strains and stresses, the pseudo potential
reveals how the thermodynamic driving force
influences the microstructure and, thus, determines the inelastic rate (see \cite{maugin1994}).
The charm of this combination lies in the fact that the potential can
be interpreted mechanically in a way similar to the
volumetric-isochoric split (\cite{flory1961}), and further, that restricting the
potential to be zero-valued, positive, and convex satisfies thermodynamical requirements a priori (see \cite{germain1983}).
Actually, the approach is in line with that for
  generalized standard materials (see \cite{halphen1975}), with the difference that the
  dissipation potential is formulated directly in terms of
  stress-like quantities.

\textbf{Objective and outline.} In this contribution, we present a new
framework for model discovery that extends Constitutive Artificial
Neural Networks (CANNs) to general inelastic
material behavior (from now on called iCANNs).
Our new iCANN assumes a multiplicative decomposition of the
deformation gradient and is, thus, capable to capture finite deformations and deformation rates.
To satisfy thermodynamics a priori, we derive the inelastic rates from a pseudo potential that depends on the thermodynamic driving force.
In Section~\ref{sec:general}, we briefly review the fundamentals of thermodynamics for inelastic materials in general.
To this end, we introduce universally valid restrictions on the Helmholtz free energy as well as the pseudo potential.
Our considerations lead us to the architecture of the network in Section~\ref{sec:iCANN}.
In analogy to CANNs, we use a feed-forward network for the pseudo potential, the underlying concept of which is inherently generic and holds for any time of inelasticity.
We subsequently embed the feed-forward networks of the energy and pseudo potential into a recurrent neural network environment.
Thus, we obtain a modular structure for iCANNs, which we specialize to visco-elasticity for illustrative purposes in Section~\ref{sec:special}.
In Section~\ref{sec:results}, we investigate the performance of our iCANN using artificially generated data, the visco-elastic behavior of polymers subjected to cyclic loading, and the relaxation behavior of muscle data.
We continue with a discussion of our approach and
  mention its limitations in Section~\ref{sec:limits}, before
  providing concluding remarks and an outlook in Section~\ref{sec:conclusion}.


\section{Generalization: Constitutive framework for inelastic materials}
\label{sec:general}
To ensure a thermodynamically consistent and objective formulation of our inelastic constitutive network, we will briefly recall the basic principles of continuum mechanics.
Our two main model ingredients, from which all constitutive relations are derived, are the Helmholtz free energy, $\psi_0$, and the pseudo potential, $g_0$, which must satisfy all these principles.
Therefore, we introduce a motion, $\bm{\varphi}$, that relates material points with position vector $\bm{x}_0$ in the reference configuration defined at a fixed moment in time to points in the current configuration, $\bm{x}=\bm{\varphi}\left(\bm{x}_0,t\right)$, at time $t$ (see \cite{marsden1994_book}). Since $\bm{\varphi}$ is a bijective function, the motion is invertible, i.e., $\bm{x}_0=\bm{\varphi}^{-1}\left(\bm{x},t\right)$.
By linearizing the (inverse) motion, we obtain the deformation gradient as well as its inverse (see \cite{holzapfel})
\begin{equation}
	\bm{F} \coloneqq \pfrac{\bm{\varphi}\left(\bm{x}_0,t\right)}{\bm{x}_0}, \quad \bm{F}^{-1} = \pfrac{\bm{\varphi}^{-1}\left(\bm{x},t\right)}{\bm{x}} \quad \text{with} \quad I_3^{\bm{F}} > 0.
\end{equation}
Note that we may also include higher gradients of motion to describe the relative change between two points, but this is not considered in this work (see \cite{bertram2020_book}).

\textbf{Objectivity.} The principle of objectivity or frame indifference restricts the choice of constitutive equations so that they must be independent of the observer.
Mathematically, this is equivalent to a rigid body motion of the current configuration (see \cite{ogdenbook}).
Thus, we can introduce a second motion, $\bm{\varphi}^{+}\left(\bm{x}_0,t\right)$, and its inverse, $\bm{\varphi}^{+^{-1}}\left(\bm{x}^{+},t\right)$, as well as the translation and rigid rotation of the current position $\bm{x}^{+}\left(\bm{x}_0,t\right)=\bm{Q}^{+}(t)\bm{x}\left(\bm{x}_0,t\right)+\bm{c}^+(t)$ with $\bm{Q}^{+} \in \mathrm{SO}(3)$.
Since the reference configuration remains the same for both motions, we conclude that $\bm{\varphi}^{-1}=\bm{\varphi}^{+^{-1}}$ and differentiate both with respect to $\bm{x}$
\begin{equation}
	\underbrace{\pfrac{\bm{\varphi}^{+^{-1}}\left(\bm{x}^{+},t\right)}{\bm{x}^{+}}}_{\eqqcolon\bm{F}^{+^{-1}}} \pfrac{\bm{x}^{+}}{\bm{x}} = \pfrac{\bm{\varphi}^{{-1}}\left(\bm{x},t\right)}{\bm{x}} \quad \Rightarrow \quad \bm{F}^{+} = \bm{Q}^{+}\bm{F}
\label{eq:objectivity}
\end{equation}
which gives us the transformation of the deformation gradient when the observer changes.

\textbf{Rigid motion of the reference configuration.} Analogous to objectivity, a rigid motion of the reference configuration discusses under which alternative choices of reference configuration the response of the material must remain unchanged (see \cite{chadwick1998_book}).
If we assume a rigid body motion of the reference configuration and subsequently apply the same deformation, i.e., the relative distance between all points is changed the same as without the rigid motion, the behavior of the material must be invariant.
Consequently, we can introduce a motion, $\bm{\varphi}^{\#}\left(\bm{x}_0^{\#},t\right)$, which accounts for a translated and rotated reference configuration, i.e., $\bm{x}_0^{\#} = \bm{Q}^{\#}\bm{x}_0 + \bm{c}^{\#}$ with $\bm{Q}^{\#} \in \mathrm{SO}(3)$.
Since both $\bm{\varphi}$ and $\bm{\varphi}^{\#}$ yield to the same current configuration, we find the following (cf. \cite{book_compu_inelas})
\begin{equation}
	\underbrace{\pfrac{\bm{\varphi}^{\#}\left(\bm{x}_0^{\#},t\right)}{\bm{x}_0^{\#}}}_{\eqqcolon\bm{F}^{\#}} \pfrac{\bm{x}_0^{\#}}{\bm{x}_0} = \pfrac{\bm{\varphi}\left(\bm{x}_0,t\right)}{\bm{x}_0} \quad \Rightarrow \quad \bm{F}^{\#} = \bm{F}\bm{Q}^{\#^{T}}.	
	\label{eq:materialsymmetry}
\end{equation}

\begin{figure}[t]
	\centering
	\input{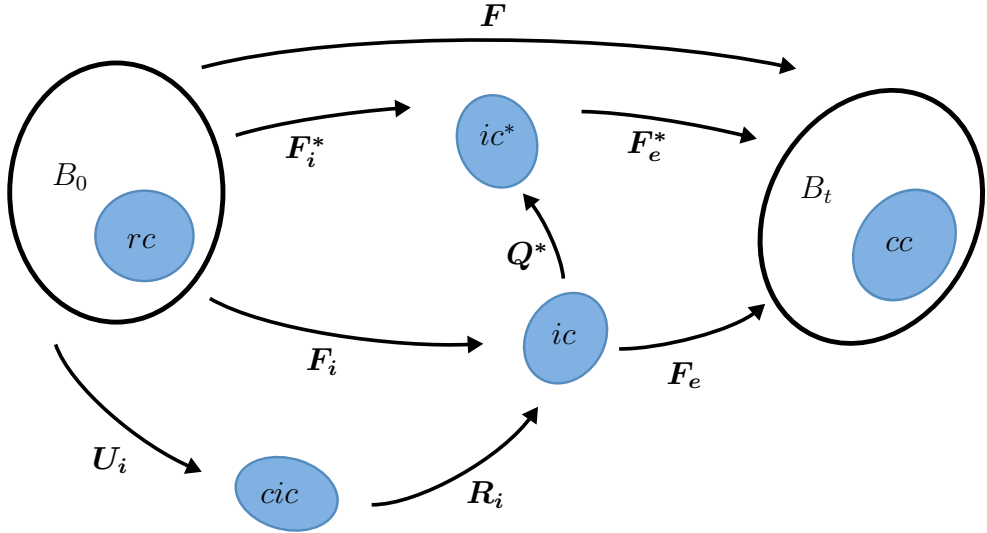}
	\caption{Multiplicative decomposition of the deformation gradient, $\bm{F}$, into elastic, $\bm{F}_e$, and inelastic, $\bm{F}_i$, parts as well as their arbitrarily rotated counterparts $\bm{F}_e^*$ and $\bm{F}_i^*$. Further, $\bm{F}_i=\bm{R}_i\bm{U}_i$ possesses its polar decomposition. Configurations: $rc$ -- reference configuration, $ic$ -- intermediate configuration, $ic^*$ -- arbitrarily rotated intermediate configuration, $cic$ -- co-rotated intermediate configuration, $cc$ -- current configuration.}
	\label{fig:mapping}
\end{figure}

\textbf{Multiplicative decomposition and rotational non-uniqueness.} Regardless of whether the material exhibits visco-elastic behavior, undergoes elasto-plastic deformations, or grows and remodels itself, all these phenomena have in common that the (elastic) stresses arise from the elastically stored energy.
Therefore, we utilize the multiplicative decomposition
\begin{equation}
	\bm{F} = \bm{F}_e \bm{F}_i, \quad \text{with} \quad I_3^{\bm{F}_e},I_3^{\bm{F}_i} > 0
\label{eq:FeFi}
\end{equation}
of the deformation gradient into elastic, $\bm{F}_e$, and inelastic, $\bm{F}_i$, parts.
This decomposition is applied in various fields of mechanics of inelastic materials, for instance, visco-elasticity (see \cite{sidoroff1974} and \cite{lubliner1985}), elasto-(visco)plasticity (see \cite{eckart1948}, \cite{kroener1959}, and \cite{lee1969}), damage (see e.g. \cite{schuette2002} and \cite{dorn2021}) as well as growth and remodeling (see \cite{rodriguez1994} and \cite{lubarda2002}).
However, since none of these parts can be derived from motion in general, they should not be misinterpreted as gradients.
Unfortunately, due to the fact that the introduced intermediate configuration is fictitious, we find that the decomposition \eqref{eq:FeFi} suffers from an inherent rotational non-uniqueness, i.e.,
\begin{equation}
	\bm{F} = \bm{F}_e \bm{Q}^{*^T}\bm{Q}^{*} \bm{F}_i \eqqcolon \bm{F}_e^{*}\bm{F}_i^{*} \quad \text{with} \quad \bm{Q}^{*} \in \mathrm{SO}(3) 
\label{eq:nonunique}
\end{equation}
is equivalent to \eqref{eq:FeFi}.
As a consequence, the multiplicative decomposition into elastic and inelastic parts is not unique, which is illustrated in Figure~\ref{fig:mapping}.
Thus, neither $\bm{F}_e$ nor $\bm{F}_i$ can be obtained in general, as there is an infinite number of possible combinations (see, for instance, the brief but precise discussion by \cite{casey2017}).
However, any constitutive law must be independent of this non-uniqueness.

\textbf{Elastic and inelastic stretch tensors.} In the case of elastic materials, objectivity constrains the Helmholtz free energy to be a function of the right stretch tensor $\bm{U}=+\sqrt{\bm{C}}$, where $\bm{C}=\bm{F}^T\bm{F}$ is the right Cauchy Green tensor, while a rigid motion of the reference configuration may lead to a dependence on the left stretch tensor $\bm{V}=+\sqrt{\bm{B}}$, where $\bm{B}=\bm{F}\bm{F}^T$ is the left Cauchy Green tensor.
If we superimpose both considerations, we find $\psi_0$ to be a scalar-valued isotropic function of its argument.
In order to find suitable stretch measures in the case of inelasticity, we superimpose the findings of Equations~\eqref{eq:objectivity}, \eqref{eq:materialsymmetry}, and \eqref{eq:nonunique}, i.e.,
\begin{equation}
	\bm{F}^{+*\#} = \underbrace{\bm{Q}^+ \bm{F}_e \bm{Q}^{*^T}}_{\eqqcolon \bm{F}_e^{+*}} \underbrace{\bm{Q}^{*} \bm{F}_i \bm{Q}^{\#^T}}_{\eqqcolon \bm{F}_i^{*\#}}.
\end{equation}
At this point, we assume $\psi_0$ to be a function of elastic and inelastic arguments.
However, analogously to elasticity, it remains to find suitable elastic and inelastic stretch tensors.
Therefore, we employ the polar decompositions of $\bm{F}_e=\bm{R}_e\bm{U}_e$ and $\bm{F}_i=\bm{V}_i\bm{R}_i$ with $\bm{R}_e, \bm{R}_i \in \mathrm{SO}(3)$ into their rotational and symmetric stretch parts.
We insert these relations into the definitions of $\bm{F}_e^{+*}$ and $\bm{F}_i^{*\#}$ and find the following (cf. \cite{green1971})
\begin{equation}
	\bm{U}_e^{+*} = \bm{Q}^{*} \bm{U}_e \bm{Q}^{*^T}, \quad \bm{V}_i^{*\#} = \bm{Q}^{*} \bm{V}_i \bm{Q}^{*^T}
\label{eq:UeVi}
\end{equation}
where $\bm{U}_e^{+*}$ and $\bm{V}_i^{*\#}$ are the stretch parts of $\bm{F}_e^{+*}$ and $\bm{F}_i^{*\#}$, respectively.
Consequently, the Helmholtz free energy as an isotropic function of $\bm{U}_e$ and $\bm{V}_i$ satisfies the requirements of objectivity, rigid motion of the reference configuration as well as rotational non-uniqueness, since both $\bm{U}_e$ and $\bm{V}_i$ transform in a similar way if subjected to these three requirements. 
Nevertheless, contrary to their invariants, it is important to note that both tensors are still not unique.

\textbf{Thermodynamically consistent evaluation.} The above considerations have shown us how to design the Helmholtz free energy as a function of stretches. 
Now, in order to obtain a thermodynamically consistent material response when subjected to arbitrary loading, we must satisfy the Clausius-Planck inequality $\mathcal{D} \coloneqq -\dot{\psi}_0 + \frac{1}{2} \bm{S}:\dot{\bm{C}} \geq 0$.
In the latter, $\bm{S}$ denotes the second Piola-Kirchhoff stress tensor.
In line with the discussion led to Equation~\eqref{eq:UeVi}, we can state the following general form of the Helmholtz free energy
\begin{equation}
	\psi_0 = \psi\left(\bm{C}_e,\bm{B}_i\right).
\label{eq:psi_general}
\end{equation}
For convenience, we expressed $\psi$ in terms of the elastic right Cauchy Green tensor, $\bm{C}_e=\bm{U}_e^2$, and the inelastic left Cauchy Green tensor, $\bm{B}_i=\bm{V}_i^2$.
We then evaluate the Clausius-Planck inequality for the Helmholtz free energy~\eqref{eq:psi_general}
\begin{equation}
	\left( \bm{S} - 2\, \bm{F}_i^{-1} \pfrac{\psi_0}{\bm{C}_e}\bm{F}_i^{-T} \right) : \frac{1}{2} \dot{\bm{C}} + \underbrace{\left( \overbrace{2\,\bm{C}_e \pfrac{\psi_0}{\bm{C}_e}}^{\eqqcolon \bm{\Sigma}} - \overbrace{2\,\pfrac{\psi_0}{\bm{B}_i}\bm{B}_i}^{\eqqcolon \bm{X}}\right)}_{\eqqcolon \bm{\Gamma}} : \bm{D}_i \geq 0
\label{eq:dissipation}
\end{equation}
and, under consideration of the arguments of \cite{coleman1961,coleman1963} and \cite{coleman1967}, end up with the state law $\bm{S}=2\, \bm{F}_i^{-1} \pfrac{\psi_0}{\bm{C}_e}\bm{F}_i^{-T}$.
Noteworthy, we employed the symmetry of the relative stress, $\bm{\Gamma}$, to reduce $\dot{\bm{F}}_i\bm{F}_i^{-1}$ to its symmetric part, $\bm{D}_i = \symp{\dot{\bm{F}}_i\bm{F}_i^{-1}}$ (see \cite{svendsen2001}).
In contrast to $\bm{\Gamma}$, neither the elastic Mandel-like stress, $\bm{\Sigma}$, nor the backstress, $\bm{X}$, are generally symmetric.
Moreover, we recognize that then subjected to the three requirements mentioned, neither the thermodynamically consistent driving forces $\bm{\Sigma}^{+*\#}=\bm{Q}^{*}\bm{\Sigma}\bm{Q}^{*^T}$, $\bm{X}^{+*\#}=\bm{Q}^{*}\bm{X}\bm{Q}^{*^T}$, and $\bm{\Gamma}^{+*\#}=\bm{Q}^{*}\bm{\Gamma}\bm{Q}^{*^T}$, nor the inelastic deformation rate $\bm{D}_i^{+*\#}=\bm{Q}^{*}\bm{D}_i\bm{Q}^{*^T}$ are unique in the sense of Equation~\eqref{eq:UeVi}.

\textbf{Evolution equation.} It remains to choose an evolution equation for $\bm{D}_i$ such that the remaining Inequality~\eqref{eq:dissipation}, $\bm{\Gamma} : \bm{D}_i \geq 0$, is fulfilled for arbitrary processes.
To this end, we introduce a pseudo potential in terms of the thermodynamically consistent driving force, $g_0=g\left(\bm{\Gamma}\right)$, which is also a scalar-valued isotropic function.
Next, the evolution equation is assumed to be (cf. \cite{kerstin1969})
\begin{equation}
	\bm{D}_i = \gamma\,\pfrac{g_0}{\bm{\Gamma}}
\label{eq:Di}
\end{equation}
where $\gamma\geq 0$ introduces a time-scale into the model.
Depending on the inelastic phenomenon modeled, $\gamma$ might be interpreted as a relaxation time (visco-elasticity) or Lagrange multiplier (elasto-plasticity).
In case of growth, the concept of homeostatic surfaces (see \cite{lamm2022}) allows the interpretation of $\gamma$.
For the time being, we constrain $g_0$ to be convex, non-negative and zero-valued at the origin, i.e., $g_0(\bm{0})=0$.
Hence, the dissipation inequality is satisfied (see \cite{germain1983}), which will be discussed in more detail in Section~\ref{sec:potential}.
In addition, it is worth mentioning that many potentials found in literature assume $g_0$ to be a positive homogeneous function of degree $\alpha>0$.
Thus, it is straightforward to prove that $\mathcal{D}\geq 0$, since $\bm{\Gamma} : \bm{D}_i=\alpha\gamma\,g\left(\bm{\Gamma}\right)$ (cf. \cite{hansen1994}).

\textbf{Co-rotated configuration.} We observed that none of the tensors defined with respect to the intermediate configuration are unique.
A consequence of this is that none of the partial derivatives with respect to $\bm{C}_e$, $\bm{B}_i$, and $\bm{\Gamma}$ can be obtained using algorithmic differentiation.
This is a pity, because in our neural network we only want to design general functions of both $\psi_0$ and $g_0$ that are capable of finding the best model to explain the experimental data without implementing additional pull-backs (see \cite{simo1992b} and \cite{dettmer2004}).
To this end, we use the co-rotated formulation introduced by \cite{holthusen2023}, which has proven advantageous when combined with algorithmic differentiation.
The fundamental idea is to rotate all non-unique tensors to their co-rotated counterparts, $\bar{(\bullet)}=\bm{R}_i^{-1}(\bullet)\bm{R}_i$ (see Figure~\ref{fig:mapping}),
\begin{equation}
	\bar{\bm{C}}_e = \bm{U}_i^{-1}\bm{C}\bm{U}_i^{-1},\quad \bar{\bm{B}}_i \equiv \bm{C}_i,\quad \bar{\bm{\Sigma}} = 2\,\bar{\bm{C}}_e\pfrac{\psi_0}{\bar{\bm{C}}_e},\quad \bar{\bm{X}} = 2\,\pfrac{\psi_0}{\bm{C}_i}\bm{C}_i, \quad \bar{\bm{D}}_i=\gamma\,\pfrac{g_0}{\bar{\bm{\Gamma}}}.
\label{eq:corotated}
\end{equation}
In the latter equation, $\bm{C}_i=\bm{F}_i^T\bm{F}_i$ is the inelastic right Cauchy Green tensor and $\bar{\bm{\Gamma}}=\bar{\bm{\Sigma}}-\bar{\bm{X}}$ is the co-rotated relative stress.
Since all tensors and their co-rotated counterparts are \textit{similar}, both $\psi(\bm{C}_e,\bm{B}_i)=\psi(\bar{\bm{C}}_e,\bm{C}_i)$ and $g(\bm{\Gamma})=g(\bar{\bm{\Gamma}})$ maintain the same functional structure.
It is interesting to note that $\bar{\bm{\Sigma}}$ and the Kirchhoff stress, $\bm{F}\bm{S}\bm{F}^T$, share the same eigenvalues (see \cite{dettmer2004}).
Hence, the pseudo potential expressed in terms of $\bar{\bm{\Gamma}}$ is physically reasonable, which is considered an advantage.
Finally, the state law reads $\bm{S}=2\, \bm{U}_i^{-1} \pfrac{\psi_0}{\bar{\bm{C}}_e}\bm{U}_i^{-1}$ (cf. \cite{coleman1963}, \cite{coleman1967}, and \cite{rice1971}).

%
\subsection{Helmholtz free energy: How strains cause stresses}
\label{sec:helmholtz}
So far, we have seen that, in addition to kinematics, the Helmholtz free energy function is the essential function that tells us how to go from strains to all the stresses we need.
Further, objectivity, rigid motion of the reference configuration, and rotational non-uniqueness led us to the conclusion that $\psi_0$ must be an isotropic function of $\bar{\bm{C}}_e$ and $\bm{C}_i$.
Thus, we can express the Helmholtz free energy in terms of its irreducible integrity basis (see \cite{spencer1971}, \cite{boehler1979}, and \cite{zheng1994})
\begin{equation}
	\psi_0 = \psi\left(\bar{\bm{C}}_e,\bm{C}_i\right) = \psi\left(I_1^{\bar{\bm{C}}_e},I_2^{\bar{\bm{C}}_e},I_3^{\bar{\bm{C}}_e},I_4,I_5,I_6,I_7,I_1^{\bm{C}_i},I_2^{\bm{C}_i},I_3^{\bm{C}_i}\right)
\label{eq:psi_general_invars}
\end{equation}
with the mixed invariants $I_4 = \tr{\bar{\bm{C}}_e\bm{C}_i}$, $I_5 = \tr{\bar{\bm{C}}_e^2\bm{C}_i}$, $I_6 = \tr{\bar{\bm{C}}_e\bm{C}_i^2}$, and $I_7 = \tr{\bar{\bm{C}}_e^2\bm{C}_i^2}$.
The constitutive artificial neural network that we design will learn the energy that best explains the experimental data by identifying the most valuable terms of a generic form for the Helmholtz free energy (see \cite{linka2023}).
The goal of this section is therefore to design such a generic function satisfying thermodynamics.

\textbf{Isotropy.} Within this contribution, we will only consider \textit{isotropic} materials, i.e., any directionally dependence is excluded.
As kinematic hardening shifts the pseudo potential in the principal stress space, we neglect any influence of $\bm{C}_i$, thus reducing the energy to $\psi\left(\bar{\bm{C}}_e\right) = \psi\left(I_1^{\bar{\bm{C}}_e},I_2^{\bar{\bm{C}}_e},I_3^{\bar{\bm{C}}_e}\right)$.
Furthermore, considering the energy as sufficiently smooth and infinitely differentiable, we can express it in terms of a Taylor series $\psi=\sum_{i,j,k=0}^\infty a_{ijk} (I_1^{\bar{\bm{C}}_e}-3)^i (I_2^{\bar{\bm{C}}_e}-3)^j (I_3^{\bar{\bm{C}}_e}-1)^k$ at the elastically undeformed state $\bar{\bm{C}}_e=\bm{I}$.
Since the free energy is zero in the undeformed state, we can conclude that $a_{000}=0$.
However, within the network that we will design, it is not possible, from a numerical point of view alone, to consider an infinite number of terms.
Unfortunately, this may violate fundamental conditions the energy must fulfill, namely
\begin{equation}
	\lim_{I_3^{\bar{\bm{C}}_e} \rightarrow\ \infty} \psi \rightarrow +\infty \quad \text{and} \quad \lim_{I_3^{\bar{\bm{C}}_e} \searrow \ 0} \psi \rightarrow +\infty.
\label{eq:require_inf}
\end{equation}
as well as
\begin{equation}
	\lim_{I_1^{\bar{\bm{C}}_e} \rightarrow\ \infty} \psi \rightarrow +\infty \quad \text{and/or} \quad \lim_{I_2^{\bar{\bm{C}}_e} \rightarrow\ \infty} \psi \rightarrow +\infty.
\label{eq:require_inf_2}
\end{equation}
\textbf{Volumetric-isochoric split.} To circumvent these issues, we exploit the volumetric-isochoric split (see \cite{flory1961}) and assume the elastic energy to take the form $\psi = \psi^{\mathrm{iso}}(\tilde{I}_1^{\bar{\bm{C}}_e},\tilde{I}_2^{\bar{\bm{C}}_e}) + \psi^{\mathrm{vol}}(I_3^{\bar{\bm{C}}_e})$ where $\tilde{I}_1^{\bar{\bm{C}}_e} = I_1^{\bar{\bm{C}}_e}/(I_3^{\bar{\bm{C}}_e})^{\frac{1}{3}}$ and $\tilde{I}_2^{\bar{\bm{C}}_e} = I_2^{\bar{\bm{C}}_e}/(I_3^{\bar{\bm{C}}_e})^{\frac{2}{3}}$.
This allows us to choose the following polynomial forms for the energy terms (cf. \cite{rivlin1951})
\begin{equation}
	\psi^{\mathrm{iso}} = \sum_{i=0}^n\sum_{j=0}^m c_{ij}\,(\tilde{I}_1^{\bar{\bm{C}}_e}-3)^i (\tilde{I}_2^{\bar{\bm{C}}_e}-3)^j, \quad \psi^{\mathrm{vol}} = \sum_{k=0}^p d_k (I_3^{\bar{\bm{C}}_e} - 1)^{2k} + W(I_3^{\bar{\bm{C}}_e})
\label{eq:vol-iso-split}
\end{equation}
where we introduced $W(I_3^{\bar{\bm{C}}_e})$ in order to ensure the requirements~\eqref{eq:require_inf}.
The isochoric response, $\psi^{\mathrm{iso}}$, addresses the second requirement~\eqref{eq:require_inf_2}.
Similar to $a_{000}$ and for mechanical convenience, we find that $c_{00}=d_0=0$ as well as $W(1)=0$.
In addition to the Helmholtz free energy being zero in the undeformed state, mechanics tells us that the same must hold for the stress.
For the isochoric response, this is naturally ensured by vanishing gradients of the modified invariants, i.e., $\pfrac{\tilde{I}_1^{\bar{\bm{C}}_e}}{\bar{\bm{C}}_e}\big\vert_{\bar{\bm{C}}_e=\bm{I}} = \pfrac{\tilde{I}_2^{\bar{\bm{C}}_e}}{\bar{\bm{C}}_e}\big\vert_{\bar{\bm{C}}_e=\bm{I}} = \bm{0}$.
Since the first term of the volumetric response, $(I_3^{\bar{\bm{C}}_e} - 1)^{2k}$, satisfies this condition as well, we conclude that $\pfrac{W(I_3^{\bar{\bm{C}}_e})}{I_3^{\bar{\bm{C}}_e}}$ must be zero for $I_3^{\bar{\bm{C}}_e}=1$.

\textbf{Polyconvexity.} Lastly, we consider polyconvexity for the design of our feed-forward network architecture of the free energy.
Therefore, we adjust the isochoric part of $\psi$ to exclude mixed products, i.e,
\begin{equation}
	\psi^{\mathrm{iso}} = \sum_{i=0}^n c_{i0}\,(\tilde{I}_1^{\bar{\bm{C}}_e}-3)^i + \sum_{j=0}^m c_{0j}\,(\tilde{I}_2^{\bar{\bm{C}}_e}-3)^j
\label{eq:psi_iso_sum}
\end{equation}
since it is generally easier to prove that the sum of convex functions is convex (cf. \cite{hartmann2003}).
Consequently, our network is considered \textit{not fully connected}.
Furthermore, in the context of our neural network, we will consider special convex functions of the power terms appearing in Equation~\eqref{eq:psi_iso_sum}.
This allows us to have a broader and more general basis of constitutive functions, while using less high numbers of exponents of the power series.
In the context of the network, these can be thought of as custom-designed activation functions (see \cite{asad2022}), which will be introduced in Section~\ref{sec:iCANN}.
%
%
\subsection{Pseudo potential: How elastic stresses lead to inelastic strains}
\label{sec:potential}
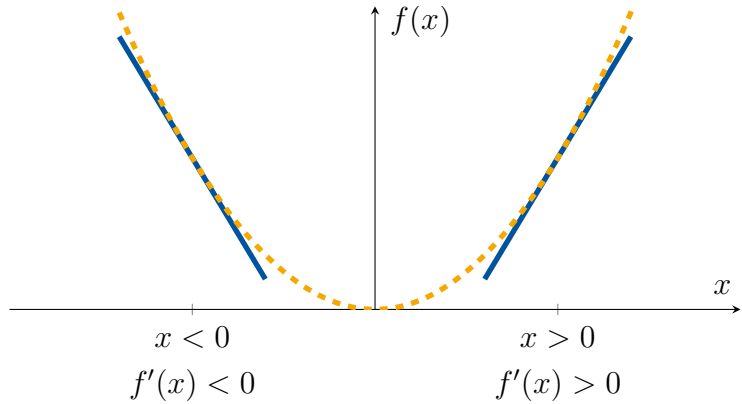
\begin{figure}[h]
	\centering
	\begin{tikzpicture}[plot/.style={very thick,mark=none,black}]
    \begin{axis}[
	height=0.35\textwidth,width=0.7\textwidth,
    ymin=0, ymax=2,
    xmin=-2, xmax=2,
    grid=none,
    axis y line=middle,
    axis x line=bottom,
    xtick={-1,1},
    ytick=\empty,
    xticklabels={{$x<0$\\$f'(x)<0$},{$x>0$\\$f'(x)>0$}},
    xticklabel style={align=center}
    ]
	\addplot[domain=0.6:1.4,samples=50,smooth,line width=2 pt,rwth1] {1+2*1*(x-1)};
	\addplot[domain=-1.4:-0.6,samples=50,smooth,line width=2 pt,rwth1] {1+2*(-1)*(x+1)};
	\addplot[domain=-1.5:1.5,samples=50,smooth,line width=2 pt,rwth2,dashed] {x^2};
	\node (A) at (390,15) {$x$};
	\node (A) at (225,190) {$f(x)$};
\end{axis}
\end{tikzpicture}
	\caption{A function, $f(x)$, that is \textit{convex}, \textit{non-negative}, and \textit{zero-valued} has the property that the sign of the derivative, $f'(x)\coloneqq \pfrac{f(x)}{x}$, evaluated at $x$ and $x$ itself coincide.}
\label{fig:convex_function}
\end{figure}
The previous section discussed general conditions that the free energy must satisfy and gave rise to the design of the feed-forward network of this very scalar quantity within the overall recurrent neural network.
Next, we will discuss the design of our second scalar function necessary to describe inelastic materials, the pseudo potential $g(\bar{\bm{\Gamma}})$.
As an isotropic function, we can express the potential in terms of its integrity basis as well, i.e., $g(\bar{\bm{\Gamma}})=g(I_1^{\bar{\bm{\Gamma}}},J_2^{\bar{\bm{\Gamma}}}, J_3^{\bar{\bm{\Gamma}}})$.

\textbf{Set of invariants.} The choice of stress invariants is by no means unique.
For instance, choosing the principal stresses, which is the case for the yield criterion by Tresca, is possible as well.
However, as will be discussed below, we will observe that this choice of invariants is vividly reflected in the evolution of inelastic strains.
For the subsequent discussions, it is useful to take the following into account
\begin{equation}
  \begin{alignedat}{3}
    &\pfrac{I_1^{\bar{\bm{\Gamma}}}}{\bar{\bm{\Gamma}}} = \bm{I},  &&\pfrac{I_1^{\bar{\bm{\Gamma}}}}{\bar{\bm{\Gamma}}} : \bar{\bm{\Gamma}} = I_1^{\bar{\bm{\Gamma}}} \\
    &\pfrac{J_2^{\bar{\bm{\Gamma}}}}{\bar{\bm{\Gamma}}} = \dev{\bar{\bm{\Gamma}}}, \quad &&\pfrac{J_2^{\bar{\bm{\Gamma}}}}{\bar{\bm{\Gamma}}} : \bar{\bm{\Gamma}} = 2\,J_2^{\bar{\bm{\Gamma}}} \\
    &\pfrac{J_3^{\bar{\bm{\Gamma}}}}{\bar{\bm{\Gamma}}} = \dev{\dev{\bar{\bm{\Gamma}}}^2}, \quad &&\pfrac{J_3^{\bar{\bm{\Gamma}}}}{\bar{\bm{\Gamma}}} : \bar{\bm{\Gamma}} = 3\,J_3^{\bar{\bm{\Gamma}}}.  
  \end{alignedat}
\label{eq:stress_invars}
\end{equation}
In the latter equation, the first invariant accounts for the hydrostatic pressure, the second takes the amount of shear stresses into account, and the third contains information about the direction and state of stress\footnote{In materials theory, Lode's angle is often used instead of the third stress invariant (cf. \cite{han1985}).}
Similar to the free energy, we expand the pseudo potential by a Taylor series at the origin $\bar{\bm{\Gamma}}=\bm{0}$, i.e., $g = \sum_{i,j,k}^{\infty} r_{ijk}(I_1^{\bar{\bm{\Gamma}}})^i (J_2^{\bar{\bm{\Gamma}}})^j (J_3^{\bar{\bm{\Gamma}}})^k$.

\textbf{Thermodynamic consistency.} As already mentioned in Section~\ref{sec:general}, the pseudo potential must be chosen such that the potential (cf. \cite{germain1983})

(i) is \textit{zero-valued} at the origin.\newline
Therefore, we choose $r_{000}=0$ in the Taylor series of $g$.
In practice, we could add any constant term to the potential and still get a thermodynamically consistent model. However, for mechanical convenience, and due to the fact that the potential can give rise to the dissipated energy, a zero-valued potential seems most reasonable.

(ii) is \textit{convex}.\newline
For similar argumentations as in the previous section, we therefore omit all mixed variant terms.
Consequently, the Taylor series reduces to
\begin{equation}
	g = \sum_{i=0}^\alpha r_{i00}\, (I_1^{\bar{\bm{\Gamma}}})^i + \sum_{j=0}^\beta r_{0j0}\,(J_2^{\bar{\bm{\Gamma}}})^j + \sum_{k=0}^\kappa r_{00k}\, (J_3^{\bar{\bm{\Gamma}}})^k \eqqcolon g_1(I_1^{\bar{\bm{\Gamma}}}) + g_2(J_2^{\bar{\bm{\Gamma}}}) + g_3(J_3^{\bar{\bm{\Gamma}}}).
\label{eq:Taylor_red_g}
\end{equation}
Hence, the feed-forward network architecture of the potential is \textit{not fully connected} as well.

(iii) is \textit{non-negative.}\newline
Since $J_2^{\bar{\bm{\Gamma}}} \geq 0$, $g_2$ in Equation~\eqref{eq:Taylor_red_g} always satisfies this condition, and further, also the requirements (i) and (ii).
Unfortunately, both $I_1^{\bar{\bm{\Gamma}}}$ and $J_3^{\bar{\bm{\Gamma}}}$ might be negative, and thus, all odd values of $j$ and $k$ can lead to negative sub-potentials $g_1$ and $g_3$.
Since we do not want to exclude odd values a priori, we need to ensure non-negativity as well as (i) and (ii) through special activation functions within the feed-forward network of our potential.
The fulfillment will therefore be guaranteed by design.
We will introduce our choice of functions in Section~\ref{sec:iCANN}.
Alternatively, one may use input convex neural networks (see \cite{amos2017}), as for instance used by \cite{asad2023}.

These three requirements together lead to a potential that has its global minimum at its origin.
In order to briefly illustrate why this leads to thermodynamically consistent, inelastic materials in general, we recall the reduced dissipation inequality, $\bar{\bm{\Gamma}} : \bar{\bm{D}}_i \geq 0$, the evolution Equation~\eqref{eq:Di}, the derivatives of the stress invariants~\eqref{eq:stress_invars}, and the choice of our potential~\eqref{eq:Taylor_red_g}, viz.
\begin{equation}
	\bar{\bm{\Gamma}} : \bar{\bm{D}}_i = \gamma \left( \pfrac{g_1(I_1^{\bar{\bm{\Gamma}}})}{I_1^{\bar{\bm{\Gamma}}}}\,I_1^{\bar{\bm{\Gamma}}} + 2\,\pfrac{g_2(J_2^{\bar{\bm{\Gamma}}})}{J_2^{\bar{\bm{\Gamma}}}}\,J_2^{\bar{\bm{\Gamma}}} + 3\,\pfrac{g_3(J_3^{\bar{\bm{\Gamma}}})}{J_3^{\bar{\bm{\Gamma}}}}\,J_3^{\bar{\bm{\Gamma}}} \right) \geq 0.
\end{equation}
Since the derivative of the sub-potentials with respect to their corresponding arguments, e.g. $\pfrac{g_1(I_1^{\bar{\bm{\Gamma}}})}{I_1^{\bar{\bm{\Gamma}}}}$ and $I_1^{\bar{\bm{\Gamma}}}$, always have the same sign due to the the requirements (i)-(iii), their product is always positive.
Figure~\ref{fig:convex_function} illustrates this property qualitatively.
Hence, we generally fulfill thermodynamics.

\textbf{Effect of stress invariants on inelastic evolution.} 
To motivate the choice of invariants as well as sub-potentials also from a more mechanical point of view, we additionally consider the inelastic volume, $I_3^{\bm{F}_i}$, or rather its evolution over time.
In this context, we found the analytical expression $	\dot{\overline{\left(I_3^{\bm{F}_i}\right)}} = I_3^{\bm{U}_i} \, \tr{\bar{\bm{D}}_i}$, revealing that the evolution over time can be expressed in terms of the evolution Equation~\eqref{eq:Di}.
With this expression at hand, it becomes obvious that the inelastic volume ratio is related to the hydrostatic pressure, $I_1^{\bar{\bm{\Gamma}}}$, which seems intuitive.
This means that as long as $\pfrac{g_1(I_1^{\bar{\bm{\Gamma}}})}{I_1^{\bar{\bm{\Gamma}}}}=0$, the volume ratio must remain constant.
For our specific choice made in Equation~\eqref{eq:Taylor_red_g}, we found that $g_1(I_1^{\bar{\bm{\Gamma}}})\neq \mathrm{const.}$ is strongly related to the volume ratio over time.
In addition, from a computational point of view, this guides our choice of time discretization schemes for the evolution equation, which will be discussed in Section~\ref{sec:special}.
\section{inelastic Constitutive Artificial Neural Networks}
\label{sec:iCANN}
The previous sections have set the physical boundaries for the constitutive equations to obtain thermodynamically consistent results. In the following, we will design an iCANN architecture that inherently satisfies these requirements. We will discuss the inputs and outputs in each time step, the general network architecture as well as the feed-forward architectures of the energy and potential networks including their individual activation functions.

\textbf{iCANN inputs and outputs.} Within a time step, the overall inputs for both networks are $\bm{C}$ and the time increment $\Delta t$. 
Moreover, since our aim is to discover history-dependent materials, the inelastic stretches, $\bm{U}_i$, are passed through the hidden states.
In addition, as we will discuss in detail in Section~\ref{sec:special}, we will use an explicit time integration scheme.
Thus, $\bm{C}$ corresponding to the last time step is passed through the hidden states as well.
While $\bar{\bm{C}}_e$ serves as the input for the energy network, the input for the potential network is $\bar{\bm{\Gamma}}$.
Furthermore, considering the constitutive requirements of objectivity, rigid motion of the reference configuration, and rotational non-uniqueness, the basis for both networks are the invariants of these tensors.

\begin{figure}[h]
	\centering
	\input{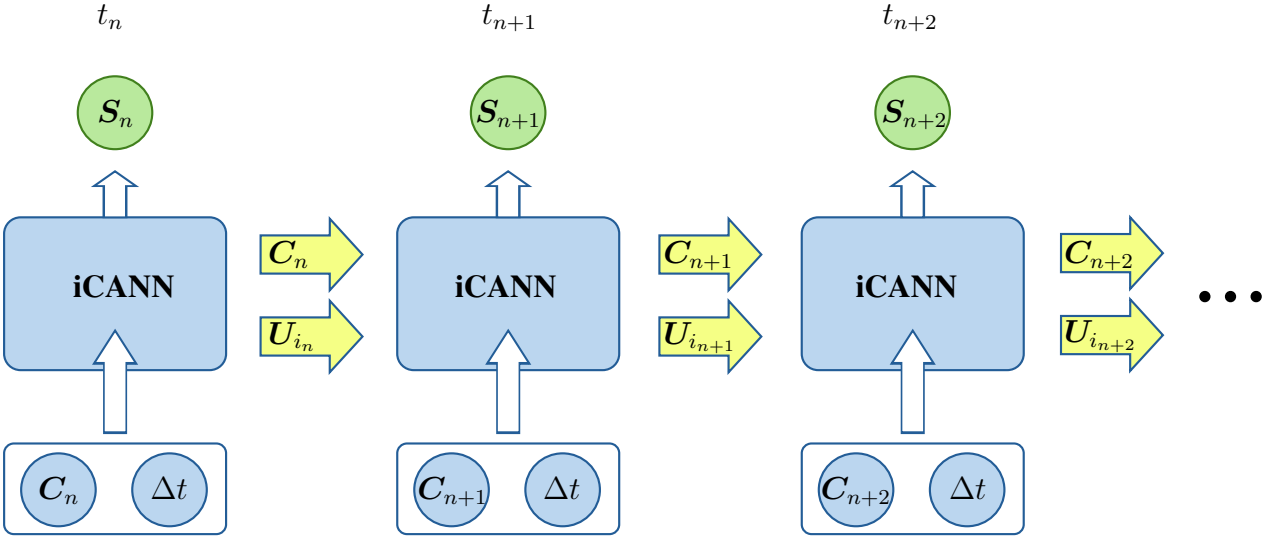}
	\caption{Schematic illustration of recurrent neural network for iCANN. The history dependence of the material is taken into account by the hidden state of the recurrent neural network. Thus, the inelastic stretches, $\bm{U}_i$, as well as the previous total stretches, $\bm{C}$, are propagated through time. Figure~\ref{fig:iCANN_architecture} illustrates the iCANN architecture in each time step.}
	\label{fig:hidden}
\end{figure}

\begin{figure}[h]
	\centering
	\input{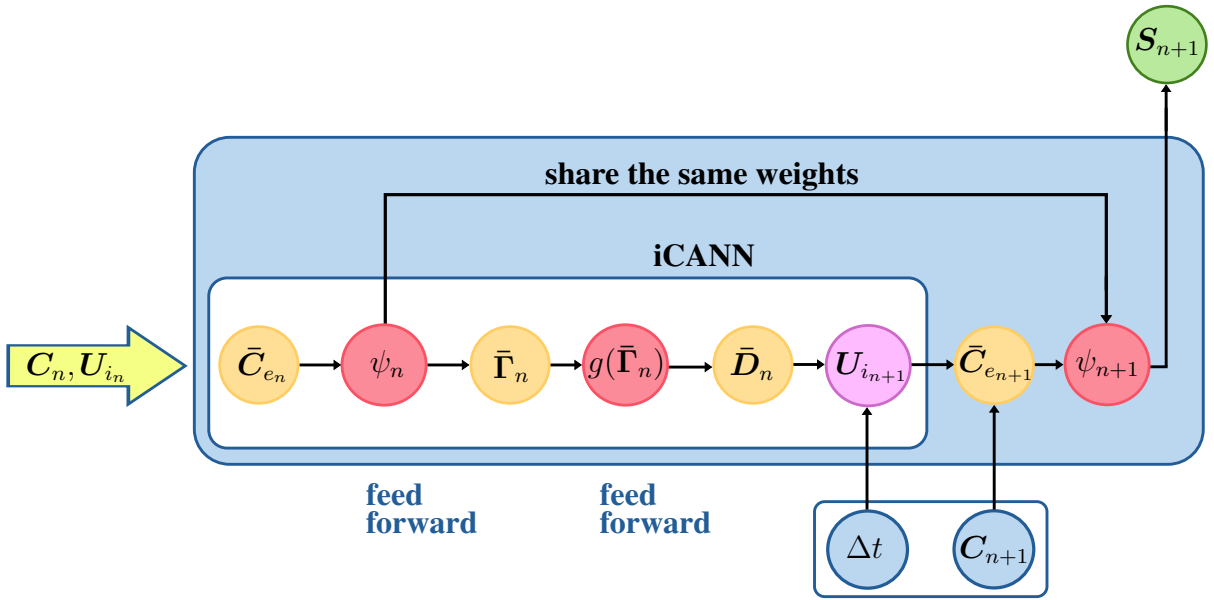}
	\caption{Schematic illustration of the iCANN architecture. Color code: blue -- current inputs; orange -- basic calculation; red -- feed-forward network; yellow -- hidden state variables; purple -- time discretization (Equation~\eqref{eq:time_discre}); green -- current output. It is important to note that the feed-forward networks for both $\psi_n$ and $\psi_{n+1}$ have the same weights. Thus, there is no double set of weights for the energy. The architecture of the energy network is illustrated in Figure~\ref{fig:network_psi}, Figure~\ref{fig:network_g} shows the architecture of the feed-forward network of the potential.}
	\label{fig:iCANN_architecture}
\end{figure}

\textbf{Network architecture.} The network architecture combines a recurrent neural network with two individual feed-forward networks for the Helmholtz free energy as well as the pseudo potential.
Both feed-forward networks are inspired by the classical CANN approach (see \cite{linka2021}).
As alluded in \cite{linka2023}, both networks are not fully connected in order to ensure polyconvexity and convexity, respectively.
Figure~\ref{fig:hidden} illustrates schematically the recurrent architecture at sequential time steps.
Noteworthy, $\Delta t$ does not have to be constant rather it is the increment between the previous and current time step, e.g., at $t_{n+1}$ we find $\Delta t = t_{n+1}-t_n$.
A more detailed illustration of the iCANN structure is provided in Figure~\ref{fig:iCANN_architecture}.
Here, we already exploited that we solve the evolution equation explicitly (cf. Equation~\eqref{eq:time_discre}).
Hence, we have to calculate the Helmholtz free energy twice, however, we must not use two feed-forward networks.
Otherwise, we would not learn a unique set of weights for $\psi$ explaining the overall material behavior.
If we would for example solve the evolution equation implicitly, we would need an additional Newton-Raphson iteration, which would certainly increase the numerical effort during training.

\textbf{Helmholtz free energy feed-forward network.} The feed-forward network for the energy that we are using is shown in Figure~\ref{fig:network_psi}.
We extend the CANN structure of \cite{linka2023} to compressible solids by decoupling the volumetric and isochoric response (cf. Equation~\ref{eq:vol-iso-split})\footnote{We kindly refer the interested reader to the aforementioned publication for a more detailed explanation.}.
Within this contribution, the explicit form we choose for $W(I_3^{\bar{\bm{C}}_e})$ is taken from \cite{ogden1972com}\footnote{For an overview of alternative choices, the reader is referred to \cite{hartmann2003}.}.
Thus, the equation for the Helmholtz free energy within our networks takes the form
\begin{equation}
\begin{split}
	\psi\left( \tilde{I}_1^{\bar{\bm{C}}_e}, \tilde{I}_2^{\bar{\bm{C}}_e}, I_3^{\bar{\bm{C}}_e} \right) &= w_{2,1}^{\psi}\,( \tilde{I}_1^{\bar{\bm{C}}_e} - 3 ) + w_{2,2}^{\psi}\,\left[ \exp(w_{1,1}^{\psi}\,(\tilde{I}_1^{\bar{\bm{C}}_e} - 3 ) ) - 1  \right] \\
&+ w_{2,3}^{\psi}\,( \tilde{I}_1^{\bar{\bm{C}}_e} - 3 )^2 + w_{2,4}^{\psi}\,\left[ \exp(w_{1,2}^{\psi}\,(\tilde{I}_1^{\bar{\bm{C}}_e} - 3 )^2 ) - 1  \right] \\
&+ w_{2,5}^{\psi}\,( \tilde{I}_2^{\bar{\bm{C}}_e} - 3 ) + w_{2,6}^{\psi}\,\left[ \exp(w_{1,3}^{\psi}\,(\tilde{I}_2^{\bar{\bm{C}}_e} - 3 ) ) - 1  \right] \\
&+ w_{2,7}^{\psi}\,( \tilde{I}_2^{\bar{\bm{C}}_e} - 3 )^2 + w_{2,8}^{\psi}\,\left[ \exp(w_{1,4}^{\psi}\,(\tilde{I}_2^{\bar{\bm{C}}_e} - 3 )^2 ) - 1  \right] \\
&+ w_{3,2}^{\psi} \underbrace{\left[ \left(I_3^{\bar{\bm{C}}_e}\right)^{-w_{3,1}^{\psi}} - 1 + w_{3,1}^{\psi}\,\ln(I_3^{\bar{\bm{C}}_e}) \right]}_{W(I_3^{\bar{\bm{C}}_e})} + \hdots
\end{split}
\label{eq:psi_network}
\end{equation}
As discussed in Section~\ref{sec:helmholtz}, we approximate the energy by a Taylor series within one layer of the network.
The subsequent layer applies specific activation functions.
In line with \cite{linka2023}, we utilize custom-designed activation functions: linear $(\bullet)$ and exponential $\exp(\bullet)-1$.
These functions are monotonic, continuous, continuous differentiable, and zero at the origin.\\
As shown in Figure~\ref{fig:network_psi}, we introduce two types of weights here.
The first, $w_{1,}^{\psi}$ and $w_{3,1}^{\psi}$, control the shape of the activation function, while the second, $w_{2,}^{\psi}$ and $w_{3,2}^{\psi}$, scale the individual contribution.
Note that due to the linearity of the first activation function, we can reduce the weights to one single weight for the linear activation, e.g., $w_{2,1}^{\psi}$.
In order to ensure physically reasonable results, all weights except $w_{3,1}^{\psi}$ are constraint to be greater or equal to zero.

\begin{figure}[h]
	\centering
	\input{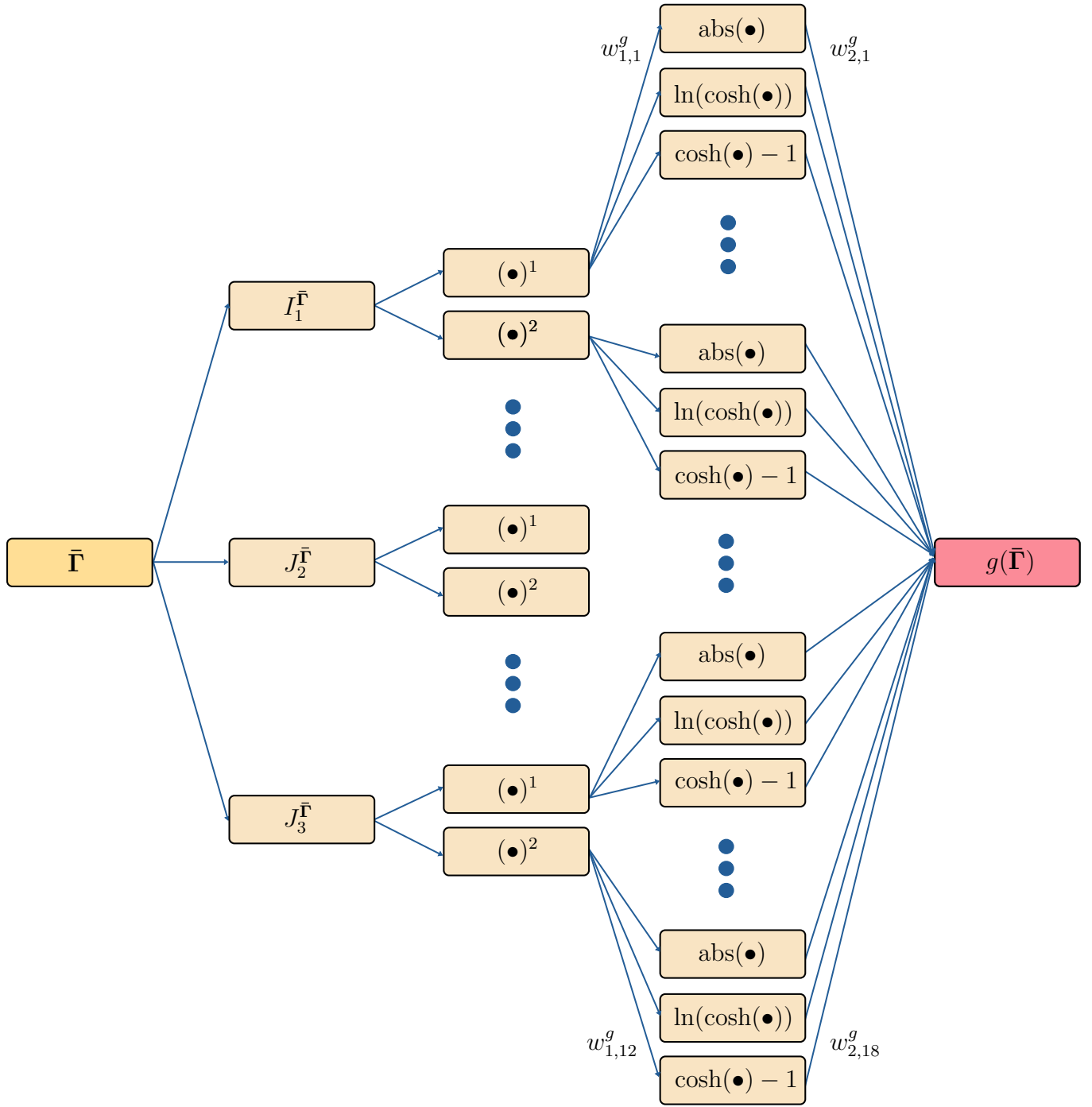}
	\caption{Schematic illustration of the feed-forward network architecture of the pseudo potential, $g$, within the recurrent neural network. The first layer computes the invariants, the second generates the powers $(\bullet)$ and $(\bullet)^2$, and the third applies the custom-designed activation functions absolute value $\mathrm{abs}(\bullet)$, logarithm/hyperbolic cosine $\mathrm{ln}(\mathrm{cosh}(\bullet))$, and the hyperbolic cosine $\mathrm{cosh}(\bullet)-1$. To satisfy convexity a priori, the network is not fully connected by design.}
	\label{fig:network_g}
\end{figure}

\textbf{Pseudo potential feed-forward network.} We use a similar structure to that of the CANN to design the feed-forward architecture of the pseudo potential.
The underlying idea is to find a general expression for the potential that is physically interpretable and that also satisfies the requirements given in Section~\ref{sec:potential} by design.
Therefore, we custom-design activation functions that satisfy a zero-value potential for zero stress, are convex for inputs of any power since we expand the potential by a Taylor series, and are non-negative for any positive and negative arguments.
One possible set of activation functions we are choosing is shown in Figure~\ref{fig:activation_func}: absolute value $\mathrm{abs}(\bullet)$, natural logarithm of the hyperbolic cosine $\mathrm{ln}(\mathrm{cosh}(\bullet))$, and the hyperbolic cosine $\mathrm{cosh}(\bullet) - 1$.
Like the activation functions for the energy, these functions are continuous and continuously differentiable.
Moreover, to guarantee convexity, we do not use a fully-connected feed-forward network, i.e., we do not include mixed products of the invariants (cf. Equation~\eqref{eq:Taylor_red_g}).\\
It remains that we introduce a set of possible weights.
To this end, we take inspiration from the CANN design, which means we introduce two types of weights. On the one hand, the shape of each activation function is controlled by a weight, $w_{1,}^g$, and, in addition, the individual contribution of each function is scaled by a second weight, $w_{2,}^g$.
Due to the properties of the absolute activation function, a weight controlling the shape is redundant.
Thus, we can again reduce the set of weights, since only one weight is needed for this very activation function.
We also restrict the range of values of the weights to be greater than or equal to zero for thermodynamic considerations.
All of these considerations lead us to the design of the potential network shown schematically in Figure~\ref{fig:network_g}, or its explicit form
\begin{equation}
\begin{split}
	g\left( I_1^{\bar{\bm{\Gamma}}}, J_2^{\bar{\bm{\Gamma}}}, J_3^{\bar{\bm{\Gamma}}} \right) &= w_{2,1}^g\, \mathrm{abs}(I_1^{\bar{\bm{\Gamma}}}) + w_{2,2}^g\, \mathrm{ln}(\mathrm{cosh}(w_{1,1}^g\,I_1^{\bar{\bm{\Gamma}}})) + w_{2,3}^g\, \left[\mathrm{cosh}(w_{1,2}^g\,I_1^{\bar{\bm{\Gamma}}}) - 1 \right] \\
&+w_{2,4}^g\, \mathrm{abs}((I_1^{\bar{\bm{\Gamma}}})^2) + w_{2,5}^g\, \mathrm{ln}(\mathrm{cosh}(w_{1,3}^g\,(I_1^{\bar{\bm{\Gamma}}})^2)) + w_{2,6}^g\, \left[\mathrm{cosh}(w_{1,4}^g\,(I_1^{\bar{\bm{\Gamma}}})^2) - 1 \right] \\
&+w_{2,7}^g\, \mathrm{abs}(J_2^{\bar{\bm{\Gamma}}}) + w_{2,8}^g\, \mathrm{ln}(\mathrm{cosh}(w_{1,5}^g\,J_2^{\bar{\bm{\Gamma}}})) + w_{2,9}^g\, \left[\mathrm{cosh}(w_{1,6}^g\,J_2^{\bar{\bm{\Gamma}}}) - 1 \right] \\
&+w_{2,10}^g\, \mathrm{abs}((J_2^{\bar{\bm{\Gamma}}})^2) + w_{2,11}^g\, \mathrm{ln}(\mathrm{cosh}(w_{1,7}^g\,(J_2^{\bar{\bm{\Gamma}}})^2)) + w_{2,12}^g\, \left[\mathrm{cosh}(w_{1,8}^g\,(J_2^{\bar{\bm{\Gamma}}})^2) - 1 \right] \\
&+w_{2,13}^g\, \mathrm{abs}(J_3^{\bar{\bm{\Gamma}}}) + w_{2,14}^g\, \mathrm{ln}(\mathrm{cosh}(w_{1,9}^g\,J_3^{\bar{\bm{\Gamma}}})) + w_{2,15}^g\, \left[\mathrm{cosh}(w_{1,10}^g\,J_3^{\bar{\bm{\Gamma}}}) - 1 \right] \\
&+w_{2,16}^g\, \mathrm{abs}((J_3^{\bar{\bm{\Gamma}}})^2) + w_{2,17}^g\, \mathrm{ln}(\mathrm{cosh}(w_{1,11}^g\,(J_3^{\bar{\bm{\Gamma}}})^2)) + w_{2,18}^g\, \left[\mathrm{cosh}(w_{1,12}^g\,(J_3^{\bar{\bm{\Gamma}}})^2) - 1 \right] \\
&+ \hdots
\end{split}
\label{eq:g_network}
\end{equation}
which can be interpreted as the sum of convex potentials.
In a more general context, this design is nothing more than a general convex yield criterion (cf. \cite{mollica2002}) with `p-norm' equal to one. 
Remarkably, changing the `p-norm' regularizes the corners of intersecting yield criteria/potentials.
Furthermore, varying the `p-norm' allows us to consider a wider range of pseudo potentials using the same set of activation functions, which is seen as an easy way to extend our approach.
From a network perspective, the `p-norm' could be seen as an additional weight.

The purpose of CANNs is to provide a simple network architecture, which is also physically sound and interpretable.
The extension made to iCANNs by the use of recurrent neural networks for the history dependence as well as the network of the pseudo potential follows exactly this way.
All quantities within the `overall' network are physically interpretable and can be clearly assigned to the influence of strains on stresses (Section~\ref{sec:helmholtz}) or to the influence of stresses on inelastic strains (Section~\ref{sec:potential}).
\begin{figure}[h]
	\centering
	\input{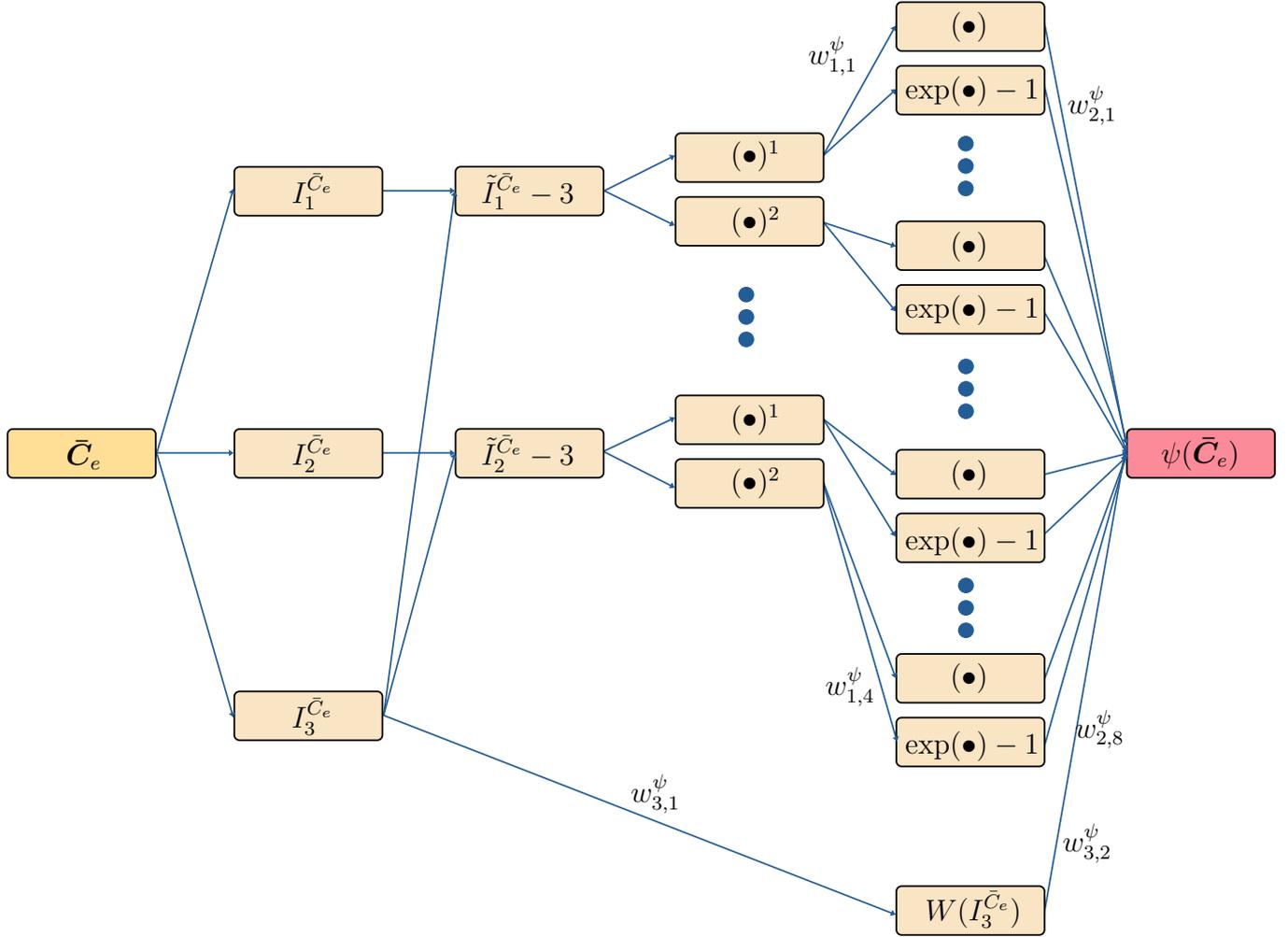}
	\caption{Schematic illustration of the feed-forward network architecture of the elastic and compressible Helmholtz free energy, $\psi$, within the recurrent neural network. The first layer computes the invariants, the second employs the volumetric-isochoric split, the third generates the powers $(\bullet)$ and $(\bullet)^2$, and the fourth applies the custom-designed activation functions identity $(\bullet)$ and the exponential $\exp(\bullet)-1$. To satisfy polyconvexity a priori, the network is not fully connected by design.}
	\label{fig:network_psi}
\end{figure}
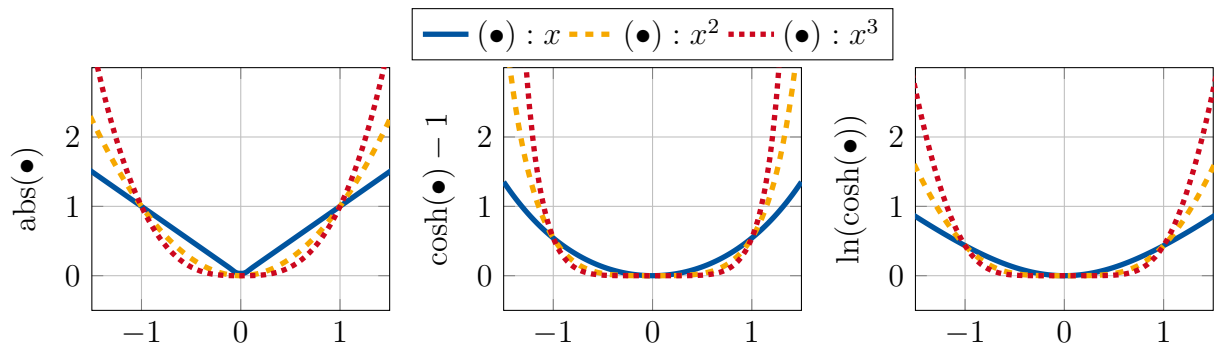
\begin{figure}[h]
	\centering
	\begin{tikzpicture}
\begin{groupplot}[group style={columns=3,rows=1,horizontal sep=1.5cm},height=4.8cm,width=5.5cm]
	\nextgroupplot[xmin =-1.5,
    			xmax=1.5,
    			ymin =-0.5,
    			ymax=3,
    			ylabel = {$\mathrm{abs}(\bullet)$},
			xlabel near ticks,    			
    			ylabel near ticks,
    			xtick={-1,0,1},
    			ytick={0,1,2},
    			grid=both]
\addplot[domain=-2:2,samples=50,smooth,line width=2 pt,rwth1] {abs(x)};
	\addplot[domain=-2:2,samples=50,smooth,line width=2 pt,rwth2,dashed] {abs(x^2)};
	\addplot[domain=-2:2,samples=50,smooth,line width=2 pt,rwth4,dotted] {abs(x^3)};
	\nextgroupplot[xmin =-1.5,
    			xmax=1.5,
    			ymin =-0.5,
    			ymax=3,
    			ylabel = {$\mathrm{cosh}(\bullet)-1$},
			xlabel near ticks,    			
    			ylabel near ticks,
    			xtick={-1,0,1},
    			ytick={0,1,2},
    			legend style={at={(0.5,1.03)},anchor=south}, 
    			legend columns=4,
    			grid=both]
	\addplot[domain=-1.5:1.5,samples=50,smooth,line width=2 pt,rwth1] {cosh(x)-1};
	\addplot[domain=-1.5:1.5,samples=50,smooth,line width=2 pt,rwth2,dashed] {cosh(x^2)-1};
	\addplot[domain=-1.5:1.5,samples=50,smooth,line width=2 pt,rwth4,dotted] {cosh(x^3)-1};
	\legend{$(\bullet): x$,$(\bullet): x^2$,$(\bullet): x^3$}
	\nextgroupplot[xmin =-1.5,
    			xmax=1.5,
    			ymin =-0.5,
    			ymax=3,
    			ylabel = {$\mathrm{ln}(\mathrm{cosh}(\bullet))$},
			xlabel near ticks,    			
    			ylabel near ticks,
    			xtick={-1,0,1},
    			ytick={0,1,2},
    			grid=both]
	\addplot[domain=-2:2,samples=50,smooth,line width=2 pt,rwth1] {ln(cosh(x))};
	\addplot[domain=-2:2,samples=50,smooth,line width=2 pt,rwth2,dashed] {ln(cosh(x^2))};
	\addplot[domain=-2:2,samples=50,smooth,line width=2 pt,rwth4,dotted] {ln(cosh(x^3))};
\end{groupplot}
\end{tikzpicture}
	\caption{Particular choices of custom-designed activation functions for the pseudo potential, $g$, with linear, quadratic, and cubic inputs. As required, all functions are \textit{convex}, \textit{non-negative}, and \textit{zero-valued}.}
	\label{fig:activation_func}
\end{figure}
\section{Specialization: Incompressibility, visco-elasticity and time discretization}
\label{sec:special}
\begin{figure}[h]
	\centering
	\input{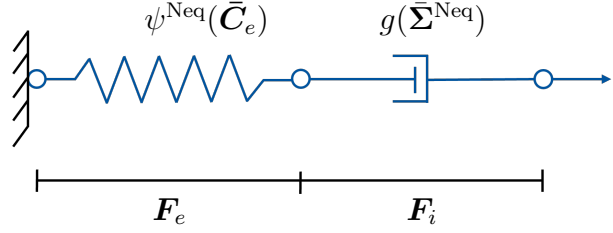}
	\caption{Schematic illustration of a single Maxwell element. Due to the history dependence and the inelastic evolution, the energy associated with the spring element, $\psi^{\mathrm{Neq}}$, is not instantaneously in equilibrium.}
\label{fig:Maxwell}
\end{figure}
In the previous chapters we have discussed the constitutive properties of general inelastic materials from a theoretical point of view as well as its particular implementation and incorporation into a recurrent neural network.
In the following, we will now specify the material equations and architecture of the network for the application of visco-elastic material behavior.

\textbf{Incompressibility and loadings.} For the time being, we restrict ourselves to the case of incompressible material behavior, i.e., $I_3^{\bm{C}} = 1$.
In order to ensure this condition within our studies, we add a Lagrange term, $-p(I_3^{\bm{C}}-1)=-p(I_3^{\bar{\bm{C}}_e}I_3^{\bm{C}_i}-1)$, to the overall Helmholtz free energy $\psi_0$.
In the latter, $p$ can be interpreted as the hydrostatic pressure and is calculated from the boundary conditions.
It is important to note that although the overall material behavior is incompressible, the elastic response might not.
Only if $I_3^{\bm{C}_i}=1$ remains constant, we obtain elastic incompressibility.
Furthermore, we only investigate coaxial loads, i.e., uniaxial loading ($\bm{C}=\mathrm{diag}\left[C_{11}, 1/\sqrt{C_{11}}, 1/\sqrt{C_{11}}\right]$), equibiaxial loading ($\bm{C}=\mathrm{diag}\left[C_{11}, C_{11}, 1/C_{11}^2\right]$), and pure shear ($\bm{C}=\mathrm{diag}\left[C_{11}, 1, 1/C_{11}\right]$) (see \cite{steinmann2012}).
Hence, the principal axes do not rotate during loading, and thus, both $\bar{\bm{C}}_e$ and $\bm{C}_i$ are diagonal tensors as well (cf. \cite{itskov2004}).
Finally, we calculate $p$ by the condition that $S_{33}=0$ must hold for each of these loading cases.

\textbf{Maxwell element.} In terms of rheology, visco-elastic material behavior can be represented as a spring and dashpot element in serial connection, which is schematically illustrated in Figure~\ref{fig:Maxwell}.
Additionally, since we consider incompressible behavior, we find the following overall energy
\begin{equation}
	\psi_0 = \psi^{\mathrm{Neq}}(\bar{\bm{C}}_e) - p\,(I_3^{\bar{\bm{C}}_e}I_3^{\bm{C}_i}-1)
\label{eq:psi_iCANN_Maxwell}
\end{equation}
where $\psi^{\mathrm{Neq}}$ belongs to the spring element.
Noteworthy, having Equation~\eqref{eq:corotated} in mind, we observe that $\bar{\bm{\Gamma}}=\bar{\bm{\Sigma}}-\bar{\bm{X}}\equiv2\,\bar{\bm{C}}_e\pfrac{\psi^{\mathrm{Neq}}(\bar{\bm{C}}_e)}{\bar{\bm{C}}_e}\eqqcolon \bar{\bm{\Sigma}}^{\mathrm{Neq}}$.
Thus, we realize that the driving force for inelastic effects, $\bar{\bm{\Sigma}}^{\mathrm{Neq}}$, is independent from the Lagrange multiplier, $p$, and express the reduced dissipation inequality as $\bar{\bm{\Sigma}}^{\mathrm{Neq}} : \bar{\bm{D}} \geq 0$ for the case of visco-elastic, incompressible material behavior.
In addition, we can rewrite the state law per Maxwell element as $\bm{S}^{\mathrm{Neq}} = 2\,\bm{U}_i^{-1}\pfrac{\psi^{\mathrm{Neq}}(\bar{\bm{C}}_e)}{\bar{\bm{C}}_e}\bm{U}_i^{-1} - p\,\bm{C}^{-1}$.\newline
Moreover, for discovering the viscous response, we need to specify the evolution Equation~\eqref{eq:Di}, where we identify $\gamma$ to be the reciprocal of the relaxation time.
In this regard, we might consider the relaxation time to be a function of $\bar{\bm{\Sigma}}^{\mathrm{Neq}}$, see \cite{ricker2023} for a comprehensive overview of suitable choices.
However, for the time being, we consider $\gamma$ to be constant.

\textbf{Reduction of Maxwell element.} For the studies we will conduct within this contribution, we design the feed-forward network of the Helmholtz free energy of the Maxwell element, $\psi^{\mathrm{Neq}}(\bar{\bm{C}}_e)$, to be the very same as shown in Equation~\eqref{eq:psi_network} without any further terms.
Consequently, the weights of this network are: 4 ($w_{1,}^{\psi}$) + 8 ($w_{2,}^{\psi}$) + 2 ($w_{3,}^{\psi}$) = 14 weights in total.
In contrast, since the amount of experimental data is limited, we reduce the form of the potential shown in Equation~\eqref{eq:g_network}, which was for example also used by \cite{tac2023}.
In addition, also in line with the latter mentioned work, we replace $J_2^{\bar{\bm{\Sigma}}^{\mathrm{Neq}}}$ by a scaled version $\tilde{J}_2^{\bar{\bm{\Sigma}}^{\mathrm{Neq}}}=3\,J_2^{\bar{\bm{\Sigma}}^{\mathrm{Neq}}}$, yielding the following potential
\begin{equation}
\begin{split}
	g\left( I_1^{\bar{\bm{\Sigma}}^{\mathrm{Neq}}}, \tilde{J}_2^{\bar{\bm{\Sigma}}^{\mathrm{Neq}}} \right) &= w_{2,1}^g\, \mathrm{abs}(I_1^{\bar{\bm{\Sigma}}^{\mathrm{Neq}}}) + w_{2,2}^g\, \mathrm{ln}(\mathrm{cosh}(w_{1,1}^g\,I_1^{\bar{\bm{\Sigma}}^{\mathrm{Neq}}})) \\
&+w_{2,4}^g\, \mathrm{abs}((I_1^{\bar{\bm{\Sigma}}^{\mathrm{Neq}}})^2) + w_{2,5}^g\, \mathrm{ln}(\mathrm{cosh}(w_{1,3}^g\,(I_1^{\bar{\bm{\Sigma}}^{\mathrm{Neq}}})^2)) \\
&+\tilde{w}_{2,7}^g\, \mathrm{abs}(\tilde{J}_2^{\bar{\bm{\Sigma}}^{\mathrm{Neq}}}) + w_{2,8}^g\, \mathrm{ln}(\mathrm{cosh}(\tilde{w}_{1,5}^g\,\tilde{J}_2^{\bar{\bm{\Sigma}}^{\mathrm{Neq}}})).
\end{split}
\label{eq:potential_iCANN_Maxwell}
\end{equation}
Thus, for the potential we have in total: 3 ($w_{1,}^{g}$) + 6 ($w_{2,}^{g}$) = 9 weights.
Noteworthy, in comparison to Equation~\eqref{eq:g_network}, the relations $w_{2,7}^g=3\,\tilde{w}_{2,7}^g$ and $w_{1,5}^g=3\,\tilde{w}_{1,5}^g$ hold.
At this point, an additional weight for the relaxation time, $\frac{1}{\gamma}$, would increase the overall number of weights in our network to 24.
However, considering Equation~\eqref{eq:Di}, we realize that this weight is obsolete if the relaxation time is constant, since the weights $w_{2,}^{g}$ can be considered to have the dimension `stiffness divided by time'.
Hence, the time-scale is included in the potential.
Although this might be unintuitive in terms of mechanics, we can reduce the numerical effort within the training process of our network without changing the underlying constitutive framework.

\begin{figure}[h]
	\centering
	\input{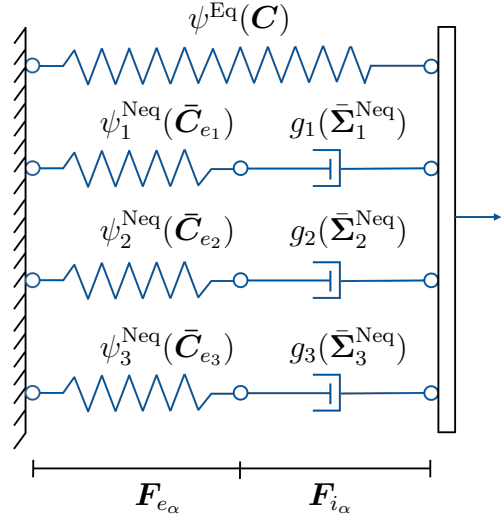}
	\caption{Schematic illustration of a generalized Maxwell element describing a visco-elastic solid. While the overall deformation, $\bm{F}$, is the same for the equilibrium, $\psi^{\mathrm{Eq}}$, and each of the non-equilibrium springs, $\psi^{\mathrm{Neq}}_\alpha$, the elastic parts, $\bm{F}_{e_\alpha}$, may very.}
\label{fig:Solid}
\end{figure}

\textbf{Visco-elastic solid.} A Maxwell element is not well suited to model visco-elastic solids.
Therefore, to learn the materials' responses of different solids in Section~\ref{sec:results}, we utilize a common approach in continuum mechanics.
Here, we connect three Maxwell elements and an equilibrium spring in parallel, where the equilibrium spring behaves purely elastically.
As a result, the Helmholtz free energy can be expressed as
\begin{equation}
	\psi_0 = \sum_{\alpha=1}^3 \left(\psi^{\mathrm{Neq}}_{\alpha}(\bar{\bm{C}}_{e_{\alpha}}) -p_{\alpha}(I_3^{\bar{\bm{C}}_{e_{\alpha}}} I_3^{\bm{C}_{i_{\alpha}}} - 1 ) \right) + \psi^{\mathrm{Eq}}(I_1^{\bm{C}},I_2^{\bm{C}}) - p (I_3^{\bm{C}}-1)
\label{eq:psi_eiCANN}
\end{equation}
together with the potential
\begin{equation}
	g_0 = \sum_{\alpha=1}^3 g_{\alpha}\left( I_1^{\bar{\bm{\Sigma}}^{\mathrm{Neq}}_{\alpha}}, \tilde{J}_2^{\bar{\bm{\Sigma}}^{\mathrm{Neq}}_\alpha} \right)
\label{eq:potential_eiCANN}
\end{equation}
with $\alpha$ denoting the contribution of each Maxwell element.
Figure~\ref{fig:Solid} schematically illustrates the iCANN used here.
Note that the energy, $\psi^{\mathrm{Eq}}$, depends solely on the right Cauchy Green tensor, $\bm{C}$.
Since the determinant of the latter is always equal to one here, we do not need to consider the invariant $I_3^{\bm{C}}$ within the equilibrium energy term.
Thus, $\psi^{\mathrm{Eq}}$ is designed in analogy to Equation~\eqref{eq:psi_network}, but without $W(I_3^{\bm{C}})$ (cf. \cite{linka2023}).
In summary, our visco-elastic solid neural network contains 3 x 23 (Maxwell element) + 1 x 12 (equilibrium spring) = 81 weights and the state law reads $\bm{S} = \sum_{\alpha=1}^3\left(2\,\bm{U}_{i_\alpha}^{-1}\pfrac{\psi^{\mathrm{Neq}}_{\alpha}(\bar{\bm{C}}_{e_{\alpha}})}{\bar{\bm{C}}_{e_{\alpha}}}\bm{U}_{i_\alpha}^{-1} - p_\alpha\,\bm{C}^{-1}\right) + 2\,\pfrac{\psi^{\mathrm{Eq}}(I_1^{\bm{C}},I_2^{\bm{C}})}{\bm{C}} - p\,\bm{C}^{-1}$\footnote{One could also combine all hydrostatic pressures, $p_\alpha$ and $p$, into one total pressure. However, since the Maxwell elements and the equilibrium spring are distinct from each other, it is convenient to separate these Lagrange functions then it comes to the numerical implementation.}.

\textbf{Time discretization.} Regardless the material behavior of interest, we need to discretize the evolution equation within the time interval $t \in \left[t_n, t_{n+1} \right]$.
As discussed in Section~\ref{sec:potential}, we have to choose an algorithm that preserves the inelastic volume in absence of $I_1^{\bar{\bm{\Sigma}}}$.
Therefore, we will use the class of exponential integrators, and further, we will integrate explicitly in time (cf. \cite{book_compu_inelas}).
For our purpose within this contribution, this choice is sufficient and easier in terms of the numerical implementation.
Exploiting that the exponential satisfies the identity $\exp\left(\bm{B}\bm{A}\bm{B}^{-1}\right)=\bm{B}\exp\left(\bm{A}\right)\bm{B}^{-1}$ if $\bm{B}$ is invertible and further algebraic reformulations (cf. \cite{holthusen2023}), we obtain the following
\begin{equation}
	\bm{C}_{i_{n+1}} = \bm{U}_{i_n}\,\exp\left(\Delta t\, 2\, \bar{\bm{D}}_{i_n} \right)\,\bm{U}_{i_n}, \quad \bm{U}_{i_{n+1}} = +\sqrt{\bm{C}_{i_{n+1}}}
\label{eq:time_discre}
\end{equation}
with the time interval $\Delta t \coloneqq t_{n+1}-t_n$.
We recognize that the derivatives in Equation~\eqref{eq:stress_invars} are all coaxial, and, due to $\exp\left(\bm{A}+\bm{B}\right)=\exp\left(\bm{A}\right)\exp\left(\bm{B}\right)$ for coaxial arguments, we can rewrite the exponential in Equation~\eqref{eq:time_discre} as
\begin{equation}
\begin{aligned}
	\exp\left(\Delta t\, 2\, \bar{\bm{D}}_{i} \right) &= \exp\left(\Delta t\, 2\gamma\, \pfrac{g_1(I_1^{\bar{\bm{\Gamma}}})}{I_1^{\bar{\bm{\Gamma}}}} \bm{I} \right) \ \exp\left(\Delta t\, 2\gamma\, \pfrac{g_2(J_2^{\bar{\bm{\Gamma}}})}{J_2^{\bar{\bm{\Gamma}}}} \dev{\bar{\bm{\Gamma}}} \right) \hdots\\
&\exp\left(\Delta t\, 2\gamma\, \pfrac{g_3(J_3^{\bar{\bm{\Gamma}}})}{J_3^{\bar{\bm{\Gamma}}}} \dev{\dev{\bar{\bm{\Gamma}}}^2} \right)
\end{aligned}
\end{equation}
where the index $n$ was omitted.
Lastly, we take the identity $\dete{\exp\left(\bm{A}\right)}=\exp\left(\tr{\bm{A}}\right)$ into account, and thus, prove that the exponential integrator satisfies \linebreak$\dete{\bm{C}_{i_{n+1}}} = \exp\left(\Delta t\, 6\gamma\, \left(\pfrac{g_1(I_1^{\bar{\bm{\Gamma}}})}{I_1^{\bar{\bm{\Gamma}}}}\right)_n \right) \dete{\bm{C}_{i_n}}$.
\section{Results}
\label{sec:results}
In the following, we apply our Maxwell iCANN (Equations~\eqref{eq:psi_iCANN_Maxwell}-\eqref{eq:potential_iCANN_Maxwell}) as well as our visco-elastic iCANN (Equations~\eqref{eq:psi_eiCANN}-\eqref{eq:potential_eiCANN}) to various experimental and artificially generated data sets.
We implemented our iCANN into the open-source software library \textit{TensorFlow} (see \cite{tensorflow}).
Since this paper focuses on the design of thermodynamically consistent iCANNs and not on questions related to the training procedure of NNs, we have set the L2-regularizer $L_2=0.001$ for all simulations. 
The relationship between potential overfitting and L2-regularization in the context of CANNs is explored, for example, by \cite{stpierre2023} and \cite{wang2023}.
The loss is expressed in terms of least squares, i.e., the sum of the squared differences between the experimentally observed stresses and the stress predicted by the iCANN (cf. \cite{wang2023}).
In addition, we use two error measures, i.e. the normalized root mean squared error, $\epsilon$, and the coefficient of determination, $R^2$, to evaluate the performance and accuracy of our iCANN
\begin{equation}
	\epsilon \coloneqq \frac{1}{|\bar{S}|} \sqrt{\frac{1}{n_{\mathrm{data}}} \sum_{i=1}^{n_{\mathrm{data}}} \left( S_{11_i} - \hat{S}_{11_i} \right)^2 }, \quad R^2 \coloneqq \max\left(0, 1 - \frac{\sum_{i=1}^{n_{\mathrm{data}}} \left( S_{11_i} - \hat{S}_{11_i} \right)^2 }{ \sum_{i=1}^{n_{\mathrm{data}}} \left( \bar{S} - \hat{S}_{11_i} \right)^2 } \right).
\end{equation}
In the latter, $S_{11}$ denotes the iCANN stress response, the experimentally observed stress is indicated by $\hat{S}_{11}$, while $n_{\mathrm{data}}$ is the number of data points.
Further, $|\bar{S}| = \frac{1}{n_{\mathrm{data}}} \sum_{i=1}^{n_{\mathrm{data}}} |\hat{S}_{11_i}|$ denotes the sum of absolute values of the experimentally observed stress and $\bar{S} = \frac{1}{n_{\mathrm{data}}} \sum_{i=1}^{n_{\mathrm{data}}} \hat{S}_{11_i}$ is its mean.
In all examples, the \textit{ADAM} optimizer is used for training the network.
The weights we discover during training for each example  can be found in Appendix~\ref{app:weights_artificial}-\ref{app:weights_vanloocke}.
%
\subsection{Example 1: Artificially generated data}
\label{ex:artificial}
We start by creating artificial data using a Maxwell model.
Consequently, the iCANN used corresponds to the Maxwell network (cf. Equations~\eqref{eq:psi_iCANN_Maxwell}-\eqref{eq:potential_iCANN_Maxwell}).
We use a continuum mechanical model to generate our artificial data.
This constitutive framework is closed by introducing a Helmholtz free energy function of the elastic stretch, a pseudo potential, and the definition of the evolution equation
\begin{align}
	\psi^{\mathrm{Neq}}(\bar{\bm{C}}_e) &= \frac{\mu}{2} \left( \frac{\tr{\bar{\bm{C}}_e}}{\det\left(\bar{\bm{C}}_e\right)^{1/3}} - 3 \right) + \frac{K}{25} \left(\det\left(\bar{\bm{C}}_e\right) - 1 - \ln\left(\det\left(\bar{\bm{C}}_e\right)\right)\right) \\
	g\left( \bar{\bm{\Sigma}}^{\mathrm{Neq}} \right) &= \frac{1}{4\,\mu} \tr{\dev{\bar{\bm{\Sigma}}^{\mathrm{Neq}}}^2} + \frac{1}{18\,K} \tr{\bar{\bm{\Sigma}}^{\mathrm{Neq}}}^2 \\
	\bar{\bm{D}}_i &= \frac{1}{\tau} \frac{\partial g}{\partial \bar{\bm{\Sigma}}^{\mathrm{Neq}}}.
\end{align}
In the latter, $\mu=12.5 \ \lbrack \rm kPa \rbrack$ is the shear modulus, $K=25\ \lbrack \rm kPa \rbrack$ is the material's bulk modulus, and $\tau=10 \ \lbrack \rm s \rbrack$ denotes the relaxation time.
We choose a compressible Neo-Hookean model, while the potential is taken from literature (see \cite{reese1998}).
For the numerical implementation, we use the algorithmic differentiation tool \textit{AceGen} (see \cite{korelc2002}) and calculate the matrix exponential by the closed-form expression provided by \cite{korelc2014}.

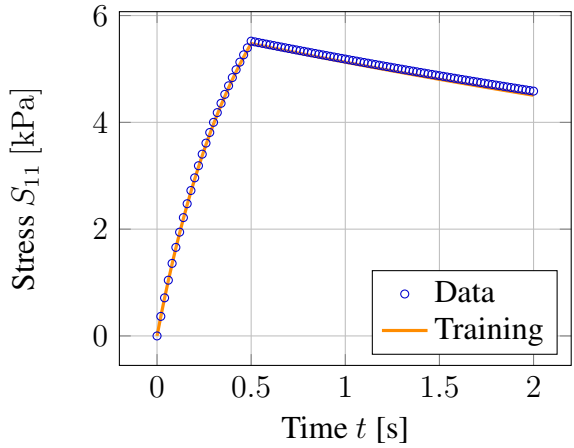
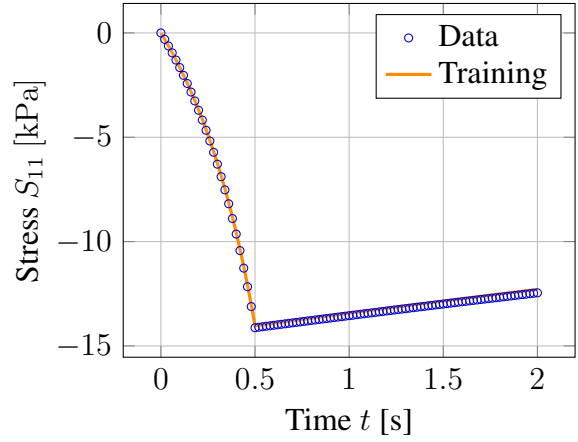
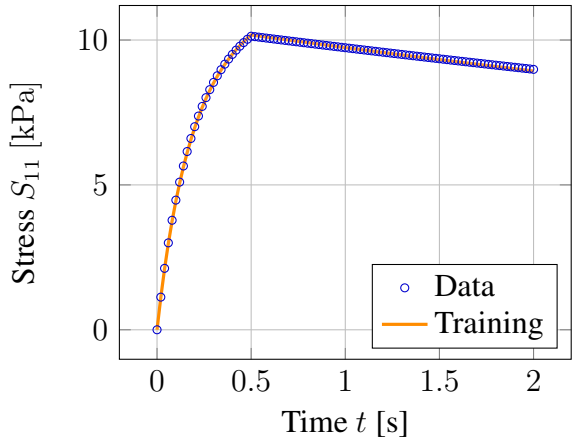
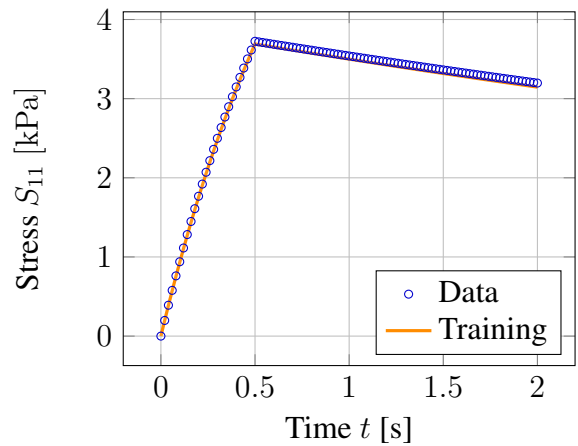
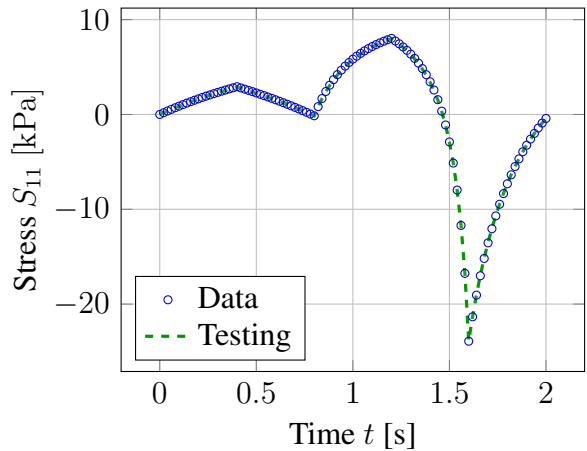
\begin{figure}[H]
	\vspace*{-1.0cm}
	\centering
	\begin{subfigure}[]{0.49\textwidth}
		\begin{tikzpicture}
\pgfplotsset{}
\begin{axis} [grid = major,
			legend pos = south east,
			legend cell align=left,
    			legend entries={Data, Training},
    			xlabel = {Time $t$ [s]},
    			ylabel = {Stress $S_{11}\ \lbrack \rm kPa \rbrack$},
    			width=0.96\textwidth,
    			height=0.8\textwidth  ,
    			/pgf/number format/1000 sep={}			
		]
    		 \addplot[color = blue,fill = white,
    		 		fill opacity=0.1, 
    		 		only marks, mark size=1.5pt,
				error bars/.cd,
				y dir = both,
				y explicit ,
				error bar style={line width=0.5pt}, 
				error mark options={line width=0.5pt, mark size=2pt, rotate=90}
				] table[x expr=\thisrowno{0}, y expr=\thisrowno{5}] {graphs/section06/analytical/Relaxation_Stretch_UT_zsm.txt};
    		 \addplot[orange, very thick
    		 ] table[x expr=\thisrowno{0}, y expr=\thisrowno{6}] {graphs/section06/analytical/Relaxation_Stretch_UT_zsm.txt};
\end{axis}
\end{tikzpicture}
		\caption{Uniaxial tension ($\epsilon=0.01$, $R^2=1.00$,\\ $C_{11}^{\mathrm{max}}=1.5$)}
		\label{fig:artificial_UTt}
	\end{subfigure}
	\begin{subfigure}[]{0.49\textwidth}
		\begin{tikzpicture}
\begin{axis} [grid = major,
			legend pos = north east,
			legend cell align=left,
    			legend entries={Data, Training},
    			xlabel = {Time $t$ [s]},
    			ylabel = {Stress $S_{11}\ \lbrack \rm kPa \rbrack$},
    			width=0.96\textwidth,
    			height=0.8\textwidth  ,
    			/pgf/number format/1000 sep={}			
		]
    		 \addplot[color = blue,fill = white,
    		 		fill opacity=0.1, 
    		 		only marks, mark size=1.5pt,
				error bars/.cd,
				y dir = both,
				y explicit ,
				error bar style={line width=0.5pt}, 
				error mark options={line width=0.5pt, mark size=2pt, rotate=90}
				] table[x expr=\thisrowno{0}, y expr=\thisrowno{5}] {graphs/section06/analytical/Relaxation_Compression_UT_zsm.txt};
    		 \addplot[orange, very thick
    		 ] table[x expr=\thisrowno{0}, y expr=\thisrowno{6}] {graphs/section06/analytical/Relaxation_Compression_UT_zsm.txt};
\end{axis}
\end{tikzpicture}
		\caption{Uniaxial compression ($\epsilon=0.01$, $R^2=1.00$,\\ $C_{11}^{\mathrm{max}}=0.6$)}
		\label{fig:artificial_UTc}
	\end{subfigure}
	\begin{subfigure}[]{0.49\textwidth}
		\begin{tikzpicture}
\begin{axis} [grid = major,
			legend pos = south east,
			legend cell align=left,
    			legend entries={Data, Training},
    			xlabel = {Time $t$ [s]},
    			ylabel = {Stress $S_{11}\ \lbrack \rm kPa \rbrack$},
    			width=0.96\textwidth,
    			height=0.8\textwidth  ,
    			/pgf/number format/1000 sep={}			
		]
    		 \addplot[color = blue,fill = white,
    		 		fill opacity=0.1, 
    		 		only marks, mark size=1.5pt,
				error bars/.cd,
				y dir = both,
				y explicit ,
				error bar style={line width=0.5pt}, 
				error mark options={line width=0.5pt, mark size=2pt, rotate=90}
				] table[x expr=\thisrowno{0}, y expr=\thisrowno{5}] {graphs/section06/analytical/Relaxation_Stretch_ET_zsm.txt};
    		 \addplot[orange, very thick
    		 ] table[x expr=\thisrowno{0}, y expr=\thisrowno{6}] {graphs/section06/analytical/Relaxation_Stretch_ET_zsm.txt};
\end{axis}
\end{tikzpicture}
		\caption{Equibiaxial tension ($\epsilon=0.00$, $R^2=1.00$,\\ $C_{11}^{\mathrm{max}}=1.8$)}
		\label{fig:artificial_ET}
	\end{subfigure}
	\begin{subfigure}[]{0.49\textwidth}
		\begin{tikzpicture}
\begin{axis} [grid = major,
			legend pos = south east,
			legend cell align=left,
    			legend entries={Data, Training},
    			xlabel = {Time $t$ [s]},
    			ylabel = {Stress $S_{11}\ \lbrack \rm kPa \rbrack$},
    			width=0.96\textwidth,
    			height=0.8\textwidth  ,
    			/pgf/number format/1000 sep={}			
		]
    		 \addplot[color = blue,fill = white,
    		 		fill opacity=0.1, 
    		 		only marks, mark size=1.5pt,
				error bars/.cd,
				y dir = both,
				y explicit ,
				error bar style={line width=0.5pt}, 
				error mark options={line width=0.5pt, mark size=2pt, rotate=90}
				] table[x expr=\thisrowno{0}, y expr=\thisrowno{5}] {graphs/section06/analytical/Relaxation_PS_zsm.txt};
    		 \addplot[orange, very thick
    		 ] table[x expr=\thisrowno{0}, y expr=\thisrowno{6}] {graphs/section06/analytical/Relaxation_PS_zsm.txt};
\end{axis}
\end{tikzpicture}
		\caption{Pure shear ($\epsilon=0.01$, $R^2=1.00$,\\ $C_{11}^{\mathrm{max}}=1.2$)}
		\label{fig:artificial_PS}
	\end{subfigure}
	\begin{subfigure}[]{\textwidth}
		\centering
		\begin{tikzpicture}
\begin{axis} [grid = major,
			legend pos = south west,
			legend cell align=left,
    			legend entries={Data, Testing},
    			xlabel = {Time $t$ [s]},
    			ylabel = {Stress $S_{11}\ \lbrack \rm kPa \rbrack$},
    			width=0.48\textwidth,
    			height=0.4\textwidth  ,
    			/pgf/number format/1000 sep={}			
		]
    		 \addplot[color = blue,fill = white,
    		 		fill opacity=0.1, 
    		 		only marks, mark size=1.5pt,
				error bars/.cd,
				y dir = both,
				y explicit ,
				error bar style={line width=0.5pt}, 
				error mark options={line width=0.5pt, mark size=2pt, rotate=90}
				] table[x expr=\thisrowno{0}, y expr=\thisrowno{5}] {graphs/section06/analytical/Cyclic_UT_zsm.txt};
    		 \addplot[green, dashed, very thick
    		 ] table[x expr=\thisrowno{0}, y expr=\thisrowno{6}] {graphs/section06/analytical/Cyclic_UT_zsm.txt};
\end{axis}
\end{tikzpicture}
		\caption{Uniaxial cyclic loading ($\epsilon=0.01, R^2=1.00$)}
		\label{fig:artificial_test}
	\end{subfigure}
\caption{Discovering a model for artificially generated data. For training, $C_{11}$ is increased linearly within the first $0.5$ seconds until $C_{11}^{\mathrm{max}}$ is reached and then kept constant for the rest of the simulation. For testing, the specimen is subjected to cyclic loading within five loading intervals.}
\label{fig:artifical}
\end{figure}

We load the model monotonically using uniaxial tension (Figure~\ref{fig:artificial_UTt}), uniaxial compression (Figure~\ref{fig:artificial_UTc}), equibiaxial tension (Figure~\ref{fig:artificial_ET}), pure shear (Figure~\ref{fig:artificial_PS}), and uniaxial cyclic tension/compression (Figure~\ref{fig:artificial_test}).
The first four loadings are relaxation tests, i.e., the load is applied within $0.5\ \lbrack \rm s \rbrack$ and then kept constant.
For cyclic loading, we utilize three cycles: (i) increase stretch until $C_{11}^{\mathrm{max}}=1.2$ at $t=0.4\ \lbrack \rm s \rbrack$, (ii) increase stretch until $C_{11}^{\mathrm{max}}=2.1$ at $t=1.2\ \lbrack \rm s \rbrack$, decrease stretch until $C_{11}^{\mathrm{max}}=0.5$ at $t=1.6\ \lbrack \rm s \rbrack$.

The iCANN is simultaneously trained using the four monotonic relaxation tests (Figures~\ref{fig:artificial_UTt}-\ref{fig:artificial_PS}).
Here, we used 10,000 epochs for training.
We observe that the iCANN is able to discover a model that explains our training data almost exactly, which can be seen qualitatively in the stress-time curves and quantitatively in terms of $\epsilon$ and $R^2$ (see Figure~\ref{fig:artifical}).
We test the performance of the iCANN using the cyclic data set.
The prediction is in almost perfect agreement with the artificially generated data.
It is important to mention that the loading/unloading rate is not constant.
This property is accurately reproduced by the model, although only relaxation tests are used for training. 
%
\subsection{Example 2: Discovering a model for the polymer VHB 4910 subjected to cyclic loading}
\label{ex:vhb}
Having validated our network on artificially generated data, we are interested in how well our approach can learn and predict experimental data.
Therefore, we take experimental data from \cite{hossain2012} for the very-high-bond (VHB) 4910 polymer.
Note that these data have already been used in a visco-elastic CANN by \cite{abdolazizi2023arxiv}.
However, they use a Prony series approach to account for visco-elasticity.
Since polymers are considered visco-elastic, we use the visco-elastic network approach (Equations~\eqref{eq:psi_eiCANN}-\eqref{eq:potential_eiCANN}).

Within the experimental investigation, \cite{hossain2012} subjected the material to uniaxial loading-unloading cycles at different maximum stretches $C_{11}^{\mathrm{max}} = \{2.25, 4.0, 6.25, 9.0\}$ and three constant stretch rates $\dot{F}_{11} = \{0.01, 0.03, 0.05 \}\ \left[1/s\right]$\footnote{$C_{11}=F_{11}^2$ for uniaxial loading.}.
For the maximum stretch $C_{11}^{\mathrm{max}} = 9.0$, the authors did not provide any experimental data for the rate $\dot{F}_{11} = 0.03\ \left[1/s\right]$.
The experimental data, which we transfered to the second Piola-Kirchhoff stress as well as the right Cauchy-Green tensor, are shown in Figure~\ref{fig:hossain}.
In addition, the authors derived a classical constitutive model using the material model of \cite{arruda1993a}.
In order to identify the corresponding material parameters, additional multi-step relaxation tests were conducted by the authors.

Similar to \cite{abdolazizi2023arxiv}, we are not using any of the multi-step relaxation tests to learn the weights of our iCANN for the polymer.
However, contrary to the latter mentioned authors, we use less experimental data for training\footnote{They use the data for $C_{11}^{\mathrm{max}} = 2.25$ as well, but exclude the rate $\dot{F}_{11} = 0.03\ \left[1/s\right]$.}.
In fact, only the data up to the maximum of $C_{11}^{\mathrm{max}} = 9.0$ is used for training (Figure~\ref{fig:hossain_1}).
We set the epochs to 6,000 for this example.
Therefore, any data with a rate of $\dot{F}_{11} = 0.03\ \left[1/s\right]$ is not seen by the iCANN during training.
The remaining nine experimental curves (Figures~\ref{fig:hossain_2}-\ref{fig:hossain_4}) serve for testing.

We observe a very good agreement with the results of the training process and the corresponding experimental data, not only qualitatively but also quantitatively.
The normalized root squared error for all experiments in provided in Table~\ref{tab:HossainEpsilon}, the coefficient of determination is listed in Table~\ref{tab:HossainR2}.
Furthermore, the results for the unseen data at stretch levels $C_{11}^{\mathrm{max}} = \{2.25, 4.0\}$ are in remarkably good agreement with the experiments.
This is particularly impressive since the data for the rate at which no training was performed is also predicted to a high degree of accuracy.
Only the data for stretch $C_{11}^{\mathrm{max}} = 6.25$ are less closely matched.
However, as noted by \cite{abdolazizi2023arxiv}, the experimental data are inconsistent here, at least from a constitutive point of view:
For constant loading rates but different maximum stretch levels applied, the stress-strain curves should be approximately the same.
This is not the case here (see Appendix~\ref{app:data_hossain}).
Thus, an artificial neural network obeying the laws of constitutive modeling is simply not capable of predicting such results.
However, we see this not as a disadvantage, but rather as an advantage of iCANNs and CANNs, since inconsistencies in the experimentally measured data will not seriously affect the discovered model.

The results shown here give us a groundswell of optimism that iCANNs are a potentially powerful combination of machine learning approaches and thermodynamics.
It is particularly noteworthy that, in contrast to the constitutive material models from the literature, we only used the cyclic experimental data and discovered an extremely accurate model.
\begin{table}[H]
\centering
\caption{Normalized root mean squared error, $\epsilon$, corresponding to the results shown in Figure~\ref{fig:hossain}. The first row represents the results of training, the remaining three rows are the values of testing. In \cite{hossain2012} no experimental data are provided for the middle rate with $C_{11}^{\mathrm{max}}=9.0\ \left[-\right]$.}
\label{tab:HossainEpsilon}
\begin{tabular}{ l l || r r r}
\hline
	\multicolumn{2}{c ||}{\multirow{2}{*}{$\epsilon$}} 	&	\multicolumn{3}{c}{$\dot{F}_{11}\ \left[1/s\right]$} \\
		&		&	0.05		&	0.03		&	0.01 \\
\hline
\hline
		\multirow{4}{*}{$C_{11}^{\mathrm{max}}$}	&	9.0		& 0.02		&	-	&	0.04 \\
												&	2.25		& 0.09		&	0.07	&	0.07 \\
												&	4.0		& 0.04		&	0.05	&	0.14 \\
												&	6.25		& 0.16		&	0.17	&	0.13 \\
\hline	
\end{tabular}%
\end{table}
\begin{table}[H]
\centering
\caption{Coefficient of determination, $R^2$, corresponding to the results shown in Figure~\ref{fig:hossain}. The first row represents the results of training, the remaining three rows are the values of testing. In \cite{hossain2012} no experimental data are provided for the middle rate with $C_{11}^{\mathrm{max}}=9.0\ \left[-\right]$.}
\label{tab:HossainR2}
\begin{tabular}{ l l || r r r}
\hline
			\multicolumn{2}{c ||}{\multirow{2}{*}{$R^2$}}		&	\multicolumn{3}{c}{$\dot{F}_{11}\ \left[1/s\right]$} \\
			&		&	0.05		&	0.03		&	0.01 \\
\hline
\hline
		\multirow{4}{*}{$C_{11}^{\mathrm{max}}$}	&	9.0		& 1.00		&	-	&	0.99 \\
												&	2.25		& 0.96		&	0.97	&	0.97 \\
												&	4.0		& 0.99		&	0.99	&	0.88 \\
												&	6.25		& 0.81		&	0.80	&	0.89 \\
\hline	
\end{tabular}%
\end{table}
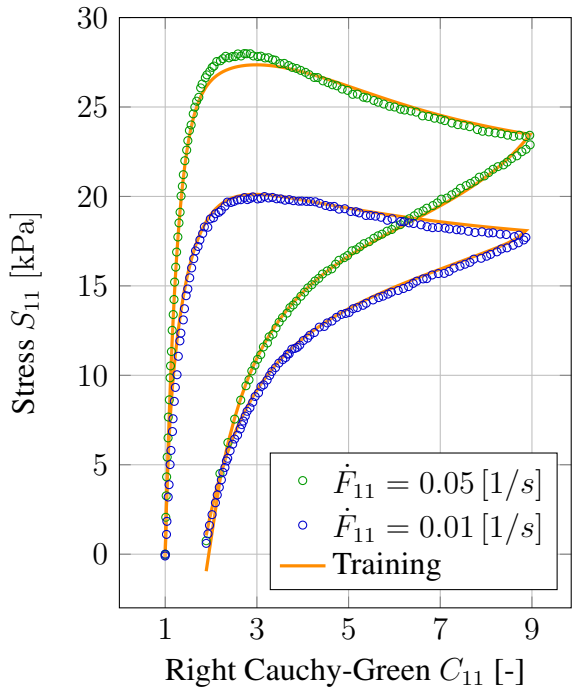
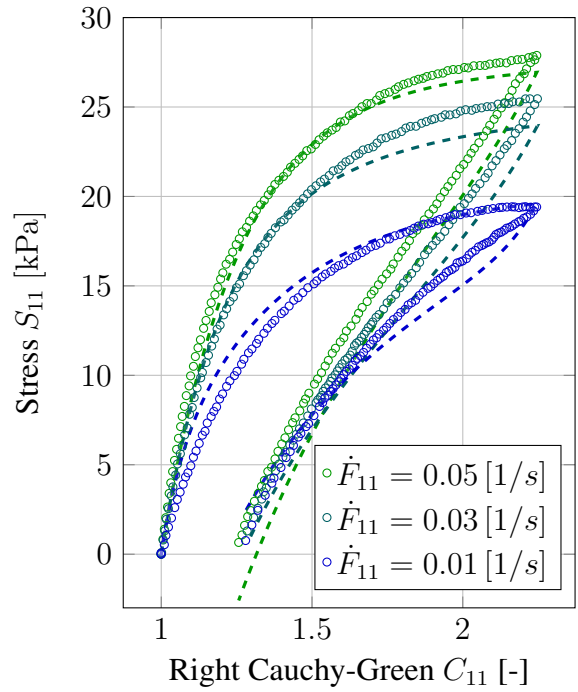
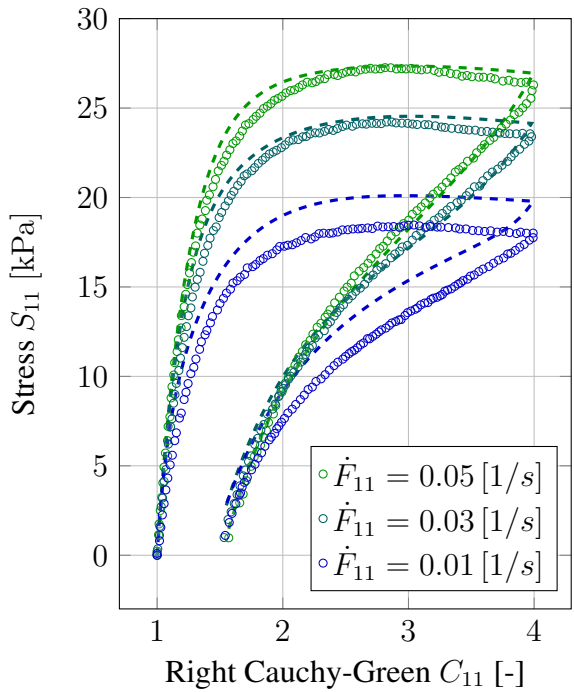
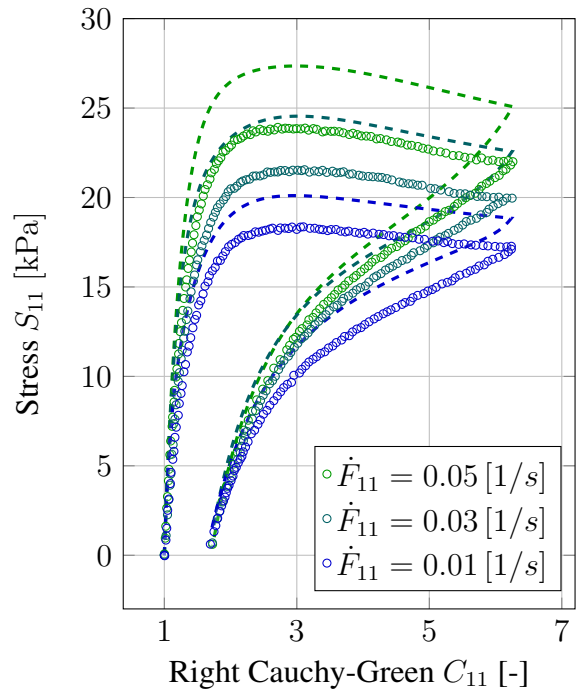
\begin{figure}[H]
	\vspace*{-1.0cm}
	\centering
	\begin{subfigure}[]{0.49\textwidth}
		\begin{tikzpicture}
\begin{axis} [grid = major,
			legend pos = south east,
			legend cell align=left,
    			legend entries={Data, Training},
    			xmin = 0,
    			ymin=-3,
    			ymax=30,
    			xtick={1,3,5,7,9},
    			xlabel = {Right Cauchy-Green $C_{11}$ [-]},
    			ylabel = {Stress $S_{11}\ \lbrack \rm kPa \rbrack$},
    			legend entries={{$\dot{F}_{11}=0.05\left[1/s\right]$}, ,{$\dot{F}_{11}=0.01\left[1/s\right]$},Training},
    			width=0.96\textwidth,
    			height=1.2\textwidth  ,
    			/pgf/number format/1000 sep={}			
		]
    		 \addplot[color = green,fill = white,
    		 		fill opacity=0.1, 
    		 		only marks, mark size=1.5pt,
				error bars/.cd,
				y dir = both,
				y explicit ,
				error bar style={line width=0.5pt}, 
				error mark options={line width=0.5pt, mark size=2pt, rotate=90}
				] table[x expr=\thisrowno{2}, y expr=\thisrowno{5}] {graphs/section06/Hossain2012/Fig4_2_rate3_zsm.txt};
    		 \addplot[orange, very thick
    		 ] table[x expr=\thisrowno{2}, y expr=\thisrowno{6}] {graphs/section06/Hossain2012/Fig4_2_rate3_zsm.txt};		
    		 \addplot[color = blue,fill = white,
    		 		fill opacity=0.1, 
    		 		only marks, mark size=1.5pt,
				error bars/.cd,
				y dir = both,
				y explicit ,
				error bar style={line width=0.5pt}, 
				error mark options={line width=0.5pt, mark size=2pt, rotate=90}
				] table[x expr=\thisrowno{2}, y expr=\thisrowno{5}] {graphs/section06/Hossain2012/Fig4_2_rate1_zsm.txt};
    		 \addplot[orange, very thick
    		 ] table[x expr=\thisrowno{2}, y expr=\thisrowno{6}] {graphs/section06/Hossain2012/Fig4_2_rate1_zsm.txt};
\end{axis}
\end{tikzpicture}
		\caption{Training with $C_{11}^{\mathrm{max}}=9.0\ [-]$ }
		\label{fig:hossain_1}
	\end{subfigure}
	\begin{subfigure}[]{0.49\textwidth}
		\begin{tikzpicture}
\begin{axis} [grid = major,
			legend pos = south east,
			legend cell align=left,
    			legend entries={Data, Training},
    			ymin=-3,
    			ymax=30,
    			xtick={1,1.5,2},
    			xlabel = {Right Cauchy-Green $C_{11}$ [-]},
    			ylabel = {Stress $S_{11}\ \lbrack \rm kPa \rbrack$},
    			legend entries={{$\dot{F}_{11}=0.05\left[1/s\right]$}, ,{$\dot{F}_{11}=0.03\left[1/s\right]$}, ,{$\dot{F}_{11}=0.01\left[1/s\right]$}, },
    			width=0.96\textwidth,
    			height=1.2\textwidth  ,
    			/pgf/number format/1000 sep={}			
		]
    		 \addplot[color = green,fill = white,
    		 		fill opacity=0.1, 
    		 		only marks, mark size=1.5pt,
				error bars/.cd,
				y dir = both,
				y explicit ,
				error bar style={line width=0.5pt}, 
				error mark options={line width=0.5pt, mark size=2pt, rotate=90}
				] table[x expr=\thisrowno{2}, y expr=\thisrowno{5}] {graphs/section06/Hossain2012/Fig3_1_rate3_zsm.txt};
    		 \addplot[green, dashed, very thick
    		 ] table[x expr=\thisrowno{2}, y expr=\thisrowno{6}] {graphs/section06/Hossain2012/Fig3_1_rate3_zsm.txt};
    		 \addplot[color = rwth3,fill = white,
    		 		fill opacity=0.1, 
    		 		only marks, mark size=1.5pt,
				error bars/.cd,
				y dir = both,
				y explicit ,
				error bar style={line width=0.5pt}, 
				error mark options={line width=0.5pt, mark size=2pt, rotate=90}
				] table[x expr=\thisrowno{2}, y expr=\thisrowno{5}] {graphs/section06/Hossain2012/Fig3_1_rate2_zsm.txt};
    		 \addplot[rwth3, dashed, very thick
    		 ] table[x expr=\thisrowno{2}, y expr=\thisrowno{6}] {graphs/section06/Hossain2012/Fig3_1_rate2_zsm.txt};    		
    		 \addplot[color = blue,fill = white,
    		 		fill opacity=0.1, 
    		 		only marks, mark size=1.5pt,
				error bars/.cd,
				y dir = both,
				y explicit ,
				error bar style={line width=0.5pt}, 
				error mark options={line width=0.5pt, mark size=2pt, rotate=90}
				] table[x expr=\thisrowno{2}, y expr=\thisrowno{5}] {graphs/section06/Hossain2012/Fig3_1_rate1_zsm.txt};
    		 \addplot[blue, dashed, very thick
    		 ] table[x expr=\thisrowno{2}, y expr=\thisrowno{6}] {graphs/section06/Hossain2012/Fig3_1_rate1_zsm.txt};
\end{axis}
\end{tikzpicture}
		\caption{Testing with $C_{11}^{\mathrm{max}}=2.25\ [-]$)}
		\label{fig:hossain_2}
	\end{subfigure}
	\begin{subfigure}[]{0.49\textwidth}
		\begin{tikzpicture}
\begin{axis} [grid = major,
			legend pos = south east,
			legend cell align=left,
    			legend entries={Data, Training},
    			ymin=-3,
    			ymax=30,
    			xtick={1,2,3,4},
    			xlabel = {Right Cauchy-Green $C_{11}$ [-]},
    			ylabel = {Stress $S_{11}\ \lbrack \rm kPa \rbrack$},
    			legend entries={{$\dot{F}_{11}=0.05\left[1/s\right]$}, ,{$\dot{F}_{11}=0.03\left[1/s\right]$}, ,{$\dot{F}_{11}=0.01\left[1/s\right]$}, },
    			width=0.96\textwidth,
    			height=1.2\textwidth  ,
    			/pgf/number format/1000 sep={}			
		]
    		 \addplot[color = green,fill = white,
    		 		fill opacity=0.1, 
    		 		only marks, mark size=1.5pt,
				error bars/.cd,
				y dir = both,
				y explicit ,
				error bar style={line width=0.5pt}, 
				error mark options={line width=0.5pt, mark size=2pt, rotate=90}
				] table[x expr=\thisrowno{2}, y expr=\thisrowno{5}] {graphs/section06/Hossain2012/Fig3_2_rate3_zsm.txt};
    		 \addplot[green, dashed, very thick
    		 ] table[x expr=\thisrowno{2}, y expr=\thisrowno{6}] {graphs/section06/Hossain2012/Fig3_2_rate3_zsm.txt};
    		 \addplot[color = rwth3,fill = white,
    		 		fill opacity=0.1, 
    		 		only marks, mark size=1.5pt,
				error bars/.cd,
				y dir = both,
				y explicit ,
				error bar style={line width=0.5pt}, 
				error mark options={line width=0.5pt, mark size=2pt, rotate=90}
				] table[x expr=\thisrowno{2}, y expr=\thisrowno{5}] {graphs/section06/Hossain2012/Fig3_2_rate2_zsm.txt};
    		 \addplot[rwth3, dashed, very thick
    		 ] table[x expr=\thisrowno{2}, y expr=\thisrowno{6}] {graphs/section06/Hossain2012/Fig3_2_rate2_zsm.txt};    		 		
    		 \addplot[color = blue,fill = white,
    		 		fill opacity=0.1, 
    		 		only marks, mark size=1.5pt,
				error bars/.cd,
				y dir = both,
				y explicit ,
				error bar style={line width=0.5pt}, 
				error mark options={line width=0.5pt, mark size=2pt, rotate=90}
				] table[x expr=\thisrowno{2}, y expr=\thisrowno{5}] {graphs/section06/Hossain2012/Fig3_2_rate1_zsm.txt};
    		 \addplot[blue, dashed, very thick
    		 ] table[x expr=\thisrowno{2}, y expr=\thisrowno{6}] {graphs/section06/Hossain2012/Fig3_2_rate1_zsm.txt};
\end{axis}
\end{tikzpicture}
		\caption{Testing with $C_{11}^{\mathrm{max}}=4.0\ [-]$}
		\label{fig:hossain_3}
	\end{subfigure}
	\begin{subfigure}[]{0.49\textwidth}
		\begin{tikzpicture}
\begin{axis} [grid = major,
			legend pos = south east,
			legend cell align=left,
    			legend entries={Data, Training},
    			xmax=7.2,
    			ymin=-3,
    			ymax=30,
    			xtick={1,3,5,7},
    			legend entries={{$\dot{F}_{11}=0.05\left[1/s\right]$}, ,{$\dot{F}_{11}=0.03\left[1/s\right]$}, ,{$\dot{F}_{11}=0.01\left[1/s\right]$}, },
    			xlabel = {Right Cauchy-Green $C_{11}$ [-]},
    			ylabel = {Stress $S_{11}\ \lbrack \rm kPa \rbrack$},
    			width=0.96\textwidth,
    			height=1.2\textwidth  ,
    			/pgf/number format/1000 sep={}			
		]
    		 \addplot[color = green,fill = white,
    		 		fill opacity=0.1, 
    		 		only marks, mark size=1.5pt,
				error bars/.cd,
				y dir = both,
				y explicit ,
				error bar style={line width=0.5pt}, 
				error mark options={line width=0.5pt, mark size=2pt, rotate=90}
				] table[x expr=\thisrowno{2}, y expr=\thisrowno{5}] {graphs/section06/Hossain2012/Fig4_1_rate3_zsm.txt};
    		 \addplot[green, dashed, very thick
    		 ] table[x expr=\thisrowno{2}, y expr=\thisrowno{6}] {graphs/section06/Hossain2012/Fig4_1_rate3_zsm.txt};	
    		 \addplot[color = rwth3,fill = white,
    		 		fill opacity=0.1, 
    		 		only marks, mark size=1.5pt,
				error bars/.cd,
				y dir = both,
				y explicit ,
				error bar style={line width=0.5pt}, 
				error mark options={line width=0.5pt, mark size=2pt, rotate=90}
				] table[x expr=\thisrowno{2}, y expr=\thisrowno{5}] {graphs/section06/Hossain2012/Fig4_1_rate2_zsm.txt};
    		 \addplot[rwth3, dashed, very thick
    		 ] table[x expr=\thisrowno{2}, y expr=\thisrowno{6}] {graphs/section06/Hossain2012/Fig4_1_rate2_zsm.txt};    		 	
    		 \addplot[color = blue,fill = white,
    		 		fill opacity=0.1, 
    		 		only marks, mark size=1.5pt,
				error bars/.cd,
				y dir = both,
				y explicit ,
				error bar style={line width=0.5pt}, 
				error mark options={line width=0.5pt, mark size=2pt, rotate=90}
				] table[x expr=\thisrowno{2}, y expr=\thisrowno{5}] {graphs/section06/Hossain2012/Fig4_1_rate1_zsm.txt};
    		 \addplot[blue, dashed, very thick
    		 ] table[x expr=\thisrowno{2}, y expr=\thisrowno{6}] {graphs/section06/Hossain2012/Fig4_1_rate1_zsm.txt};
\end{axis}
\end{tikzpicture}
		\caption{Testing with $C_{11}^{\mathrm{max}}=6.25\ [-]$}
		\label{fig:hossain_4}
	\end{subfigure}
\caption{Discovering a model for the experimental data of the polymer VHB 4910 taken from \cite{hossain2012}. Each Figure~\ref{fig:hossain_1}-\ref{fig:hossain_4} shows loading/unloading stress-stretch curves at constant deformation rates, $\dot{F}_{11}$, but different maximum applied stretch levels, $C_{11}^{\mathrm{max}}$. The experimental data are indicated by dots, the solid lines correspond to the training results of the network (Figure~\ref{fig:hossain_1}), and dashed lines represent the testing/prediction of the discovered model (Figures~\ref{fig:hossain_2}-\ref{fig:hossain_4}).}
\label{fig:hossain}
\end{figure}
%
\subsection{Example 3: Discovering a model for passive skeletal muscle subjected to relaxation}
\label{ex:muscle}
Finally, we are interested in discovering a model for passive skeletal muscle subjected to uniaxial compression.
We take the data of \cite{vanloocke2008}, who performed relaxation tests.
In total, they performed five different experiments on the same material, i.e. they varied both the maximum stretch $C_{11}^{\mathrm{max}}=\{0.81,0.64,0.49,0.49,0.49\}\,\left[-\right]$ applied for relaxation and the rate of deformation $\dot{F}_{11}=\{0.01,0.01,0.01,0.005,0.05\}\,\left[ s \right]$.
The experimental results used here are the mean of six tests per data set.

\cite{wang2023} studied the performance of a Constitutive Artificial Neural Network combined with a Prony series approach to take visco-elasticity into account.
In addition, they compared their CANN model with the performance of a `vanilla' recurrent neural network, i.e., no physical or constitutive constraints are included in the design of the network.
Although the performance of the training sets was more accurate compared to CANNs, the prediction generally provided non-physical results, which were in strong disagreement with thermodynamics.
Thus, viscoelastic CANNs outperform classical neural network approaches because their results are most likely to violate physics.
Another aspect investigated by the authors was the influence of increased data for training.
One the one hand, they started by training the network using one experimental curve and test their discovered model using the remaining four experiments.
On the other hand, training the network using four experimental data sets and just using one unseen data set for testing should generally increase the network's performance.
In the following, we will study the same using our iCANN approach.
Regardless whether we train on one or on four experiments, we set the epochs to 5,000 in this example.

\textbf{Train on one, test on four.} To begin with, we train our iCANN on one experimental curve and test on four remaining data sets.
Additionally, we vary the training data set with respect to the five different combination of maximum stretch and rate of deformation.
The models we learned as well as the experimental data are shown in Figure~\ref{fig:TrainOne}.
Table~\ref{tab:TrainOneEpsilon} lists the normalized root mean squared error, while the coefficient of determination is given in Table~\ref{tab:TrainOneR2}.
Generally, we can say that the training data sets are fitted very well using our iCANN approach, regardless of which experiment is used for training.
Summarizing each column in Table~\ref{tab:TrainOneEpsilon}, we find the third column to perform the best ($\sum \epsilon = 1.21$).
Nevertheless, both the $\epsilon$ and $R^2$ values of the tests in each column are relatively poor, suggesting that an optimal model has not yet been discovered.
As possible causes we may consider either the limitation of the number of Maxwell elements to three or the insufficient amount of training data.
Regarding the first reason, however, it can be assumed that no better model will be found if the training data remain the same, since the error measures are already very good for the training data.

\textbf{Train on four, test on one.} Consequently, we increase the data used for training.
Our results are illustrated in Figure~\ref{fig:TrainFour} where we use four curves for training and test on the remaining experimental curve.
Again, we calculate the normalized root mean squared error in Table~\ref{tab:TrainFourEpsilon} and the coefficient of determination in Table~\ref{tab:TrainFourR2}.
As expected, we generally observe a way better learning performance compared to the setup `train on one, test on four'.
Overall, both error measurements are in a relatively good range for in case of training and testing.
In contrast to the previous setup, where it was difficult to find a model that explained both training and testing, the results shown in the second and fifth columns in particular agree very well with training and especially with testing.
The best model is discovered in the second column, where we have a summarized normalized root mean squared error of $\sum \epsilon = 0.68$.
As already mentioned above, we might obtain better results if we increase the number of Maxwell elements, however, this is out of the scope of this contribution.

Regardless whether we train with one or four experiments, our results infer that the choice of the training set is by no means unique or trivial.
Different choices influence the accuracy of prediction capabilities.
There are many reasons for this, including the quality of the measurement during the experimental investigation, the choice of optimizer, or how much information about inelastic effects is hidden in the data.
Nevertheless, we observe once more that due to the very general nature of iCANN, models can be found in any combination that not only agree with the training data, as is the case for `vanilla' networks, but also provide extremely good results when tested and, moreover, always deliver to us results that obey the fundamental physical principles of our world.
\begin{figure}[h]
    \centering
    \vspace*{-0mm}
    \begin{subfigure}{.05\textwidth} 
    		\vspace*{-25mm}
				\begin{tikzpicture}[rotate=90,transform shape]
			\begin{axis}[xlabel= Nice $x$ label, ylabel= Nice $y$ label, width=55mm, height=35mm,
				hide axis,
				xmin=10,
				xmax=50,
				ymin=0,
				ymax=0.4,
				legend columns=-1,
				legend style={column sep=1mm}
				]
					\addlegendimage{color = blue,fill = white,
    		 				fill opacity=0.1, 
	    		 			only marks, mark size=1.5pt,
						error bars/.cd,
						y dir = both,
						y explicit,
						error bar style={line width=0.5pt}, 
						error mark options={line width=0.5pt, mark size=2pt, rotate=90}}
					\addlegendentry{Data}
					\addlegendimage{orange, very thick}
					\addlegendentry{Training} 
					\addlegendimage{green, dashed, very thick}
					\addlegendentry{Testing} 
			\end{axis}
		\end{tikzpicture} 
	\end{subfigure}%
    \begin{subfigure}{.18\textwidth} 
		\begin{tikzpicture}[rotate=90,transform shape]
            \begin{axis}[grid = major, width=44mm, height=40mm, xmin=0,xmax=350,ymin=-5,ymax=0,xticklabels=\empty,ytick={0,-2,-4},yticklabels={\vphantom{0},\vphantom{-2},\vphantom{-4}}]
                \addplot[color = blue,fill = white,
    		 		fill opacity=0.1, 
    		 		only marks, mark size=1.5pt,
				error bars/.cd,
				y dir = both,
				y explicit,
				error bar style={line width=0.5pt}, 
				error mark options={line width=0.5pt, mark size=2pt, rotate=90}
				] table[x expr=\thisrowno{0}, y expr=\thisrowno{5}]{graphs/section06/VanLoocke2008/VanLoocke_rate1_zsm.txt};
                \addplot[green, dashed, very thick] table[x expr=\thisrowno{0}, y expr=\thisrowno{10}]{graphs/section06/VanLoocke2008/VanLoocke_rate1_zsm.txt};
            \end{axis}
        \end{tikzpicture}
        \begin{tikzpicture}[rotate=90,transform shape]
            \begin{axis}[grid = major, width=44mm, height=40mm, xmin=0,xmax=350,ymin=-5,ymax=0,xticklabels=\empty,ytick={0,-2,-4},yticklabels={\vphantom{0},\vphantom{-2},\vphantom{-4}}]
                \addplot[color = blue,fill = white,
    		 		fill opacity=0.1, 
    		 		only marks, mark size=1.5pt,
				error bars/.cd,
				y dir = both,
				y explicit,
				error bar style={line width=0.5pt}, 
				error mark options={line width=0.5pt, mark size=2pt, rotate=90}
				] table[x expr=\thisrowno{0}, y expr=\thisrowno{5}]{graphs/section06/VanLoocke2008/VanLoocke_rate1_zsm.txt};
                \addplot[green, dashed, very thick] table[x expr=\thisrowno{0}, y expr=\thisrowno{9}]{graphs/section06/VanLoocke2008/VanLoocke_rate1_zsm.txt};
            \end{axis}
        \end{tikzpicture}
        \begin{tikzpicture}[rotate=90,transform shape]
            \begin{axis}[grid = major, width=44mm, height=40mm, xmin=0,xmax=350,ymin=-5,ymax=0,xticklabels=\empty,ytick={0,-2,-4},yticklabels={\vphantom{0},\vphantom{-2},\vphantom{-4}}]
                \addplot[color = blue,fill = white,
    		 		fill opacity=0.1, 
    		 		only marks, mark size=1.5pt,
				error bars/.cd,
				y dir = both,
				y explicit,
				error bar style={line width=0.5pt}, 
				error mark options={line width=0.5pt, mark size=2pt, rotate=90}
				] table[x expr=\thisrowno{0}, y expr=\thisrowno{5}]{graphs/section06/VanLoocke2008/VanLoocke_rate1_zsm.txt};
                \addplot[green, dashed, very thick] table[x expr=\thisrowno{0}, y expr=\thisrowno{8}]{graphs/section06/VanLoocke2008/VanLoocke_rate1_zsm.txt};
            \end{axis}
        \end{tikzpicture}
        \begin{tikzpicture}[rotate=90,transform shape]
            \begin{axis}[grid = major, width=44mm, height=40mm, xmin=0,xmax=350,ymin=-5,ymax=0,xticklabels=\empty,ytick={0,-2,-4},yticklabels={\vphantom{0},\vphantom{-2},\vphantom{-4}}]
                \addplot[color = blue,fill = white,
    		 		fill opacity=0.1, 
    		 		only marks, mark size=1.5pt,
				error bars/.cd,
				y dir = both,
				y explicit,
				error bar style={line width=0.5pt}, 
				error mark options={line width=0.5pt, mark size=2pt, rotate=90}
				] table[x expr=\thisrowno{0}, y expr=\thisrowno{5}]{graphs/section06/VanLoocke2008/VanLoocke_rate1_zsm.txt};
                \addplot[green, dashed, very thick] table[x expr=\thisrowno{0}, y expr=\thisrowno{7}]{graphs/section06/VanLoocke2008/VanLoocke_rate1_zsm.txt};
            \end{axis}
        \end{tikzpicture}
        \begin{tikzpicture}[rotate=90,transform shape]
            \begin{axis}[grid = major, width=44mm, height=40mm, xmin=0,xmax=350,ymin=-5,ymax=0,xticklabels=\empty,ylabel={\phantom{Stress $S_{11}\ \lbrack \rm kPa \rbrack$}}]
                \addplot[color = blue,fill = white,
    		 		fill opacity=0.1, 
    		 		only marks, mark size=1.5pt,
				error bars/.cd,
				y dir = both,
				y explicit,
				error bar style={line width=0.5pt}, 
				error mark options={line width=0.5pt, mark size=2pt, rotate=90}
				] table[x expr=\thisrowno{0}, y expr=\thisrowno{5}]{graphs/section06/VanLoocke2008/VanLoocke_rate1_zsm.txt};
                \addplot[orange, very thick] table[x expr=\thisrowno{0}, y expr=\thisrowno{6}]{graphs/section06/VanLoocke2008/VanLoocke_rate1_zsm.txt};
            \end{axis}
        \end{tikzpicture} 
    \end{subfigure}%
    \begin{subfigure}{.18\textwidth} 
		\begin{tikzpicture}[rotate=90,transform shape]
            \begin{axis}[grid = major, width=44mm, height=40mm, xmin=0,xmax=350,ymin=-5,ymax=0,xticklabels=\empty,ytick={0,-2,-4},yticklabels={\vphantom{0},\vphantom{-2},\vphantom{-4}}]
                \addplot[color = blue,fill = white,
    		 		fill opacity=0.1, 
    		 		only marks, mark size=1.5pt,
				error bars/.cd,
				y dir = both,
				y explicit,
				error bar style={line width=0.5pt}, 
				error mark options={line width=0.5pt, mark size=2pt, rotate=90}
				] table[x expr=\thisrowno{0}, y expr=\thisrowno{5}]{graphs/section06/VanLoocke2008/VanLoocke_rate2_zsm.txt};
                \addplot[green, dashed, very thick] table[x expr=\thisrowno{0}, y expr=\thisrowno{10}]{graphs/section06/VanLoocke2008/VanLoocke_rate2_zsm.txt};
            \end{axis}
        \end{tikzpicture}
        \begin{tikzpicture}[rotate=90,transform shape]
            \begin{axis}[grid = major, width=44mm, height=40mm, xmin=0,xmax=350,ymin=-5,ymax=0,xticklabels=\empty,ytick={0,-2,-4},yticklabels={\vphantom{0},\vphantom{-2},\vphantom{-4}}]
                \addplot[color = blue,fill = white,
    		 		fill opacity=0.1, 
    		 		only marks, mark size=1.5pt,
				error bars/.cd,
				y dir = both,
				y explicit,
				error bar style={line width=0.5pt}, 
				error mark options={line width=0.5pt, mark size=2pt, rotate=90}
				] table[x expr=\thisrowno{0}, y expr=\thisrowno{5}]{graphs/section06/VanLoocke2008/VanLoocke_rate2_zsm.txt};
                \addplot[green, dashed, very thick] table[x expr=\thisrowno{0}, y expr=\thisrowno{9}]{graphs/section06/VanLoocke2008/VanLoocke_rate2_zsm.txt};
            \end{axis}
        \end{tikzpicture}
        \begin{tikzpicture}[rotate=90,transform shape]
            \begin{axis}[grid = major, width=44mm, height=40mm, xmin=0,xmax=350,ymin=-5,ymax=0,xticklabels=\empty,ytick={0,-2,-4},yticklabels={\vphantom{0},\vphantom{-2},\vphantom{-4}}]
                \addplot[color = blue,fill = white,
    		 		fill opacity=0.1, 
    		 		only marks, mark size=1.5pt,
				error bars/.cd,
				y dir = both,
				y explicit,
				error bar style={line width=0.5pt}, 
				error mark options={line width=0.5pt, mark size=2pt, rotate=90}
				] table[x expr=\thisrowno{0}, y expr=\thisrowno{5}]{graphs/section06/VanLoocke2008/VanLoocke_rate2_zsm.txt};
                \addplot[green, dashed, very thick] table[x expr=\thisrowno{0}, y expr=\thisrowno{8}]{graphs/section06/VanLoocke2008/VanLoocke_rate2_zsm.txt};
            \end{axis}
        \end{tikzpicture}
        \begin{tikzpicture}[rotate=90,transform shape]
            \begin{axis}[grid = major, width=44mm, height=40mm, xmin=0,xmax=350,ymin=-5,ymax=0,xticklabels=\empty,ytick={0,-2,-4},yticklabels={\vphantom{0},\vphantom{-2},\vphantom{-4}}]
                \addplot[color = blue,fill = white,
    		 		fill opacity=0.1, 
    		 		only marks, mark size=1.5pt,
				error bars/.cd,
				y dir = both,
				y explicit,
				error bar style={line width=0.5pt}, 
				error mark options={line width=0.5pt, mark size=2pt, rotate=90}
				] table[x expr=\thisrowno{0}, y expr=\thisrowno{5}]{graphs/section06/VanLoocke2008/VanLoocke_rate2_zsm.txt};
                \addplot[orange, very thick] table[x expr=\thisrowno{0}, y expr=\thisrowno{7}]{graphs/section06/VanLoocke2008/VanLoocke_rate2_zsm.txt};
            \end{axis}
        \end{tikzpicture}
        \begin{tikzpicture}[rotate=90,transform shape]
            \begin{axis}[grid = major, width=44mm, height=40mm, xmin=0,xmax=350,ymin=-5,ymax=0,xticklabels=\empty,ylabel={\phantom{Stress $S_{11}\ \lbrack \rm kPa \rbrack$}}]
                \addplot[color = blue,fill = white,
    		 		fill opacity=0.1, 
    		 		only marks, mark size=1.5pt,
				error bars/.cd,
				y dir = both,
				y explicit,
				error bar style={line width=0.5pt}, 
				error mark options={line width=0.5pt, mark size=2pt, rotate=90}
				] table[x expr=\thisrowno{0}, y expr=\thisrowno{5}]{graphs/section06/VanLoocke2008/VanLoocke_rate2_zsm.txt};
                \addplot[green, dashed, very thick] table[x expr=\thisrowno{0}, y expr=\thisrowno{6}]{graphs/section06/VanLoocke2008/VanLoocke_rate2_zsm.txt};
            \end{axis}
        \end{tikzpicture} 
    \end{subfigure}%
    \begin{subfigure}{.18\textwidth} 
		\begin{tikzpicture}[rotate=90,transform shape]
            \begin{axis}[grid = major, width=44mm, height=40mm, xmin=0,xmax=350,ymin=-5,ymax=0,xticklabels=\empty,ytick={0,-2,-4},yticklabels={\vphantom{0},\vphantom{-2},\vphantom{-4}}]
                \addplot[color = blue,fill = white,
    		 		fill opacity=0.1, 
    		 		only marks, mark size=1.5pt,
				error bars/.cd,
				y dir = both,
				y explicit,
				error bar style={line width=0.5pt}, 
				error mark options={line width=0.5pt, mark size=2pt, rotate=90}
				] table[x expr=\thisrowno{0}, y expr=\thisrowno{5}]{graphs/section06/VanLoocke2008/VanLoocke_rate3_zsm.txt};
                \addplot[green, dashed, very thick] table[x expr=\thisrowno{0}, y expr=\thisrowno{10}]{graphs/section06/VanLoocke2008/VanLoocke_rate3_zsm.txt};
            \end{axis}
        \end{tikzpicture}
        \begin{tikzpicture}[rotate=90,transform shape]
            \begin{axis}[grid = major, width=44mm, height=40mm, xmin=0,xmax=350,ymin=-5,ymax=0,xticklabels=\empty,ytick={0,-2,-4},yticklabels={\vphantom{0},\vphantom{-2},\vphantom{-4}}]
                \addplot[color = blue,fill = white,
    		 		fill opacity=0.1, 
    		 		only marks, mark size=1.5pt,
				error bars/.cd,
				y dir = both,
				y explicit,
				error bar style={line width=0.5pt}, 
				error mark options={line width=0.5pt, mark size=2pt, rotate=90}
				] table[x expr=\thisrowno{0}, y expr=\thisrowno{5}]{graphs/section06/VanLoocke2008/VanLoocke_rate3_zsm.txt};
                \addplot[green, dashed, very thick] table[x expr=\thisrowno{0}, y expr=\thisrowno{9}]{graphs/section06/VanLoocke2008/VanLoocke_rate3_zsm.txt};
            \end{axis}
        \end{tikzpicture}
        \begin{tikzpicture}[rotate=90,transform shape]
            \begin{axis}[grid = major, width=44mm, height=40mm, xmin=0,xmax=350,ymin=-5,ymax=0,xticklabels=\empty,ytick={0,-2,-4},yticklabels={\vphantom{0},\vphantom{-2},\vphantom{-4}}]
                \addplot[color = blue,fill = white,
    		 		fill opacity=0.1, 
    		 		only marks, mark size=1.5pt,
				error bars/.cd,
				y dir = both,
				y explicit,
				error bar style={line width=0.5pt}, 
				error mark options={line width=0.5pt, mark size=2pt, rotate=90}
				] table[x expr=\thisrowno{0}, y expr=\thisrowno{5}]{graphs/section06/VanLoocke2008/VanLoocke_rate3_zsm.txt};
                \addplot[orange, very thick] table[x expr=\thisrowno{0}, y expr=\thisrowno{8}]{graphs/section06/VanLoocke2008/VanLoocke_rate3_zsm.txt};
            \end{axis}
        \end{tikzpicture}
        \begin{tikzpicture}[rotate=90,transform shape]
            \begin{axis}[grid = major, width=44mm, height=40mm, xmin=0,xmax=350,ymin=-5,ymax=0,xticklabels=\empty,ytick={0,-2,-4},yticklabels={\vphantom{0},\vphantom{-2},\vphantom{-4}}]
                \addplot[color = blue,fill = white,
    		 		fill opacity=0.1, 
    		 		only marks, mark size=1.5pt,
				error bars/.cd,
				y dir = both,
				y explicit,
				error bar style={line width=0.5pt}, 
				error mark options={line width=0.5pt, mark size=2pt, rotate=90}
				] table[x expr=\thisrowno{0}, y expr=\thisrowno{5}]{graphs/section06/VanLoocke2008/VanLoocke_rate3_zsm.txt};
                \addplot[green, dashed, very thick] table[x expr=\thisrowno{0}, y expr=\thisrowno{7}]{graphs/section06/VanLoocke2008/VanLoocke_rate3_zsm.txt};
            \end{axis}
        \end{tikzpicture}
        \begin{tikzpicture}[rotate=90,transform shape]
            \begin{axis}[grid = major, width=44mm, height=40mm, xmin=0,xmax=350,ymin=-5,ymax=0,xticklabels=\empty,ylabel={Stress $S_{11}\ \lbrack \rm kPa \rbrack$}]
                \addplot[color = blue,fill = white,
    		 		fill opacity=0.1, 
    		 		only marks, mark size=1.5pt,
				error bars/.cd,
				y dir = both,
				y explicit,
				error bar style={line width=0.5pt}, 
				error mark options={line width=0.5pt, mark size=2pt, rotate=90}
				] table[x expr=\thisrowno{0}, y expr=\thisrowno{5}]{graphs/section06/VanLoocke2008/VanLoocke_rate3_zsm.txt};
                \addplot[green, dashed, very thick] table[x expr=\thisrowno{0}, y expr=\thisrowno{6}]{graphs/section06/VanLoocke2008/VanLoocke_rate3_zsm.txt};
            \end{axis}
        \end{tikzpicture} 
    \end{subfigure}%
    \begin{subfigure}{.18\textwidth} 
		\begin{tikzpicture}[rotate=90,transform shape]
            \begin{axis}[grid = major, width=44mm, height=40mm, xmin=0,xmax=350,ymin=-5,ymax=0,xticklabels=\empty,ytick={0,-2,-4},yticklabels={\vphantom{0},\vphantom{-2},\vphantom{-4}}]
                \addplot[color = blue,fill = white,
    		 		fill opacity=0.1, 
    		 		only marks, mark size=1.5pt,
				error bars/.cd,
				y dir = both,
				y explicit,
				error bar style={line width=0.5pt}, 
				error mark options={line width=0.5pt, mark size=2pt, rotate=90}
				] table[x expr=\thisrowno{0}, y expr=\thisrowno{5}]{graphs/section06/VanLoocke2008/VanLoocke_rate4_zsm.txt};
                \addplot[green, dashed, very thick] table[x expr=\thisrowno{0}, y expr=\thisrowno{10}]{graphs/section06/VanLoocke2008/VanLoocke_rate4_zsm.txt};
            \end{axis}
        \end{tikzpicture}
        \begin{tikzpicture}[rotate=90,transform shape]
            \begin{axis}[grid = major, width=44mm, height=40mm, xmin=0,xmax=350,ymin=-5,ymax=0,xticklabels=\empty,ytick={0,-2,-4},yticklabels={\vphantom{0},\vphantom{-2},\vphantom{-4}}]
                \addplot[color = blue,fill = white,
    		 		fill opacity=0.1, 
    		 		only marks, mark size=1.5pt,
				error bars/.cd,
				y dir = both,
				y explicit,
				error bar style={line width=0.5pt}, 
				error mark options={line width=0.5pt, mark size=2pt, rotate=90}
				] table[x expr=\thisrowno{0}, y expr=\thisrowno{5}]{graphs/section06/VanLoocke2008/VanLoocke_rate4_zsm.txt};
                \addplot[orange, very thick] table[x expr=\thisrowno{0}, y expr=\thisrowno{9}]{graphs/section06/VanLoocke2008/VanLoocke_rate4_zsm.txt};
            \end{axis}
        \end{tikzpicture}
        \begin{tikzpicture}[rotate=90,transform shape]
            \begin{axis}[grid = major, width=44mm, height=40mm, xmin=0,xmax=350,ymin=-5,ymax=0,xticklabels=\empty,ytick={0,-2,-4},yticklabels={\vphantom{0},\vphantom{-2},\vphantom{-4}}]
                \addplot[color = blue,fill = white,
    		 		fill opacity=0.1, 
    		 		only marks, mark size=1.5pt,
				error bars/.cd,
				y dir = both,
				y explicit,
				error bar style={line width=0.5pt}, 
				error mark options={line width=0.5pt, mark size=2pt, rotate=90}
				] table[x expr=\thisrowno{0}, y expr=\thisrowno{5}]{graphs/section06/VanLoocke2008/VanLoocke_rate4_zsm.txt};
                \addplot[green, dashed, very thick] table[x expr=\thisrowno{0}, y expr=\thisrowno{8}]{graphs/section06/VanLoocke2008/VanLoocke_rate4_zsm.txt};
            \end{axis}
        \end{tikzpicture}
        \begin{tikzpicture}[rotate=90,transform shape]
            \begin{axis}[grid = major, width=44mm, height=40mm, xmin=0,xmax=350,ymin=-5,ymax=0,xticklabels=\empty,ytick={0,-2,-4},yticklabels={\vphantom{0},\vphantom{-2},\vphantom{-4}}]
                \addplot[color = blue,fill = white,
    		 		fill opacity=0.1, 
    		 		only marks, mark size=1.5pt,
				error bars/.cd,
				y dir = both,
				y explicit,
				error bar style={line width=0.5pt}, 
				error mark options={line width=0.5pt, mark size=2pt, rotate=90}
				] table[x expr=\thisrowno{0}, y expr=\thisrowno{5}]{graphs/section06/VanLoocke2008/VanLoocke_rate4_zsm.txt};
                \addplot[green, dashed, very thick] table[x expr=\thisrowno{0}, y expr=\thisrowno{7}]{graphs/section06/VanLoocke2008/VanLoocke_rate4_zsm.txt};
            \end{axis}
        \end{tikzpicture}
        \begin{tikzpicture}[rotate=90,transform shape]
            \begin{axis}[grid = major, width=44mm, height=40mm, xmin=0,xmax=350,ymin=-5,ymax=0,xticklabels=\empty,ylabel={\phantom{Stress $S_{11}\ \lbrack \rm kPa \rbrack$}}]
                \addplot[color = blue,fill = white,
    		 		fill opacity=0.1, 
    		 		only marks, mark size=1.5pt,
				error bars/.cd,
				y dir = both,
				y explicit,
				error bar style={line width=0.5pt}, 
				error mark options={line width=0.5pt, mark size=2pt, rotate=90}
				] table[x expr=\thisrowno{0}, y expr=\thisrowno{5}]{graphs/section06/VanLoocke2008/VanLoocke_rate4_zsm.txt};
                \addplot[green, dashed, very thick] table[x expr=\thisrowno{0}, y expr=\thisrowno{6}]{graphs/section06/VanLoocke2008/VanLoocke_rate4_zsm.txt};
            \end{axis}
        \end{tikzpicture} 
    \end{subfigure}%
    \begin{subfigure}{.18\textwidth} 
        \begin{tikzpicture}[rotate=90,transform shape]
            \begin{axis}[grid = major, width=44mm, height=40mm, xmin=0,xmax=350,ymin=-5,ymax=0,ytick={0,-2,-4},yticklabels={\vphantom{0},\vphantom{-2},\vphantom{-4}}]
                \addplot[color = blue,fill = white,
    		 		fill opacity=0.1, 
    		 		only marks, mark size=1.5pt,
				error bars/.cd,
				y dir = both,
				y explicit,
				error bar style={line width=0.5pt}, 
				error mark options={line width=0.5pt, mark size=2pt, rotate=90}
				] table[x expr=\thisrowno{0}, y expr=\thisrowno{5}]{graphs/section06/VanLoocke2008/VanLoocke_rate5_zsm.txt};
                \addplot[orange, very thick] table[x expr=\thisrowno{0}, y expr=\thisrowno{10}]{graphs/section06/VanLoocke2008/VanLoocke_rate5_zsm.txt};
            \end{axis}
        \end{tikzpicture}
        \begin{tikzpicture}[rotate=90,transform shape]
            \begin{axis}[grid = major, width=44mm, height=40mm, xmin=0,xmax=350,ymin=-5,ymax=0,ytick={0,-2,-4},yticklabels={\vphantom{0},\vphantom{-2},\vphantom{-4}}]
                \addplot[color = blue,fill = white,
    		 		fill opacity=0.1, 
    		 		only marks, mark size=1.5pt,
				error bars/.cd,
				y dir = both,
				y explicit,
				error bar style={line width=0.5pt}, 
				error mark options={line width=0.5pt, mark size=2pt, rotate=90}
				] table[x expr=\thisrowno{0}, y expr=\thisrowno{5}]{graphs/section06/VanLoocke2008/VanLoocke_rate5_zsm.txt};
                \addplot[green, dashed, very thick] table[x expr=\thisrowno{0}, y expr=\thisrowno{9}]{graphs/section06/VanLoocke2008/VanLoocke_rate5_zsm.txt};
            \end{axis}
        \end{tikzpicture}
        \begin{tikzpicture}[rotate=90,transform shape]
            \begin{axis}[grid = major, width=44mm, height=40mm, xmin=0,xmax=350,ymin=-5,ymax=0,ytick={0,-2,-4},yticklabels={\vphantom{0},\vphantom{-2},\vphantom{-4}},xlabel={Time $t$ [s]}]
                \addplot[color = blue,fill = white,
    		 		fill opacity=0.1, 
    		 		only marks, mark size=1.5pt,
				error bars/.cd,
				y dir = both,
				y explicit,
				error bar style={line width=0.5pt}, 
				error mark options={line width=0.5pt, mark size=2pt, rotate=90}
				] table[x expr=\thisrowno{0}, y expr=\thisrowno{5}]{graphs/section06/VanLoocke2008/VanLoocke_rate5_zsm.txt};
                \addplot[green, dashed, very thick] table[x expr=\thisrowno{0}, y expr=\thisrowno{8}]{graphs/section06/VanLoocke2008/VanLoocke_rate5_zsm.txt};
            \end{axis}
        \end{tikzpicture}
        \begin{tikzpicture}[rotate=90,transform shape]
            \begin{axis}[grid = major, width=44mm, height=40mm, xmin=0,xmax=350,ymin=-5,ymax=0,ytick={0,-2,-4},yticklabels={\vphantom{0},\vphantom{-2},\vphantom{-4}}]
                \addplot[color = blue,fill = white,
    		 		fill opacity=0.1, 
    		 		only marks, mark size=1.5pt,
				error bars/.cd,
				y dir = both,
				y explicit,
				error bar style={line width=0.5pt}, 
				error mark options={line width=0.5pt, mark size=2pt, rotate=90}
				] table[x expr=\thisrowno{0}, y expr=\thisrowno{5}]{graphs/section06/VanLoocke2008/VanLoocke_rate5_zsm.txt};
                \addplot[green, dashed, very thick] table[x expr=\thisrowno{0}, y expr=\thisrowno{7}]{graphs/section06/VanLoocke2008/VanLoocke_rate5_zsm.txt};
            \end{axis}
        \end{tikzpicture}
        \begin{tikzpicture}[rotate=90,transform shape]
            \begin{axis}[grid = major, width=44mm, height=40mm, xmin=0,xmax=350,ymin=-5,ymax=0,ytick={0,-2,-4},ylabel={\phantom{Stress $S_{11}\ \lbrack \rm kPa \rbrack$}}]
                \addplot[color = blue,fill = white,
    		 		fill opacity=0.1, 
    		 		only marks, mark size=1.5pt,
				error bars/.cd,
				y dir = both,
				y explicit,
				error bar style={line width=0.5pt}, 
				error mark options={line width=0.5pt, mark size=2pt, rotate=90}
				] table[x expr=\thisrowno{0}, y expr=\thisrowno{5}]{graphs/section06/VanLoocke2008/VanLoocke_rate5_zsm.txt};
                \addplot[green, dashed, very thick] table[x expr=\thisrowno{0}, y expr=\thisrowno{6}]{graphs/section06/VanLoocke2008/VanLoocke_rate5_zsm.txt};
            \end{axis}
        \end{tikzpicture} 
    \end{subfigure}
\caption{\textbf{Train on one, test on four.} Experimental data and discovered model for passive skeletal muscle taken from \cite{vanloocke2008}. Each column represents a combination of training and testing, while each row represents one experimental setup. The experimental setup varies with respect to the maximum applied stretch $C_{11}^{\mathrm{max}}=\{0.81,0.64,0.49,0.49,0.49\}\,\left[-\right]$ and the rate of deformation $\dot{F}_{11}=\{0.01,0.01,0.01,0.005,0.05\}\,\left[ s \right]$. The columns vary with respect to the one experiment used for training.}
\label{fig:TrainOne}
\end{figure}
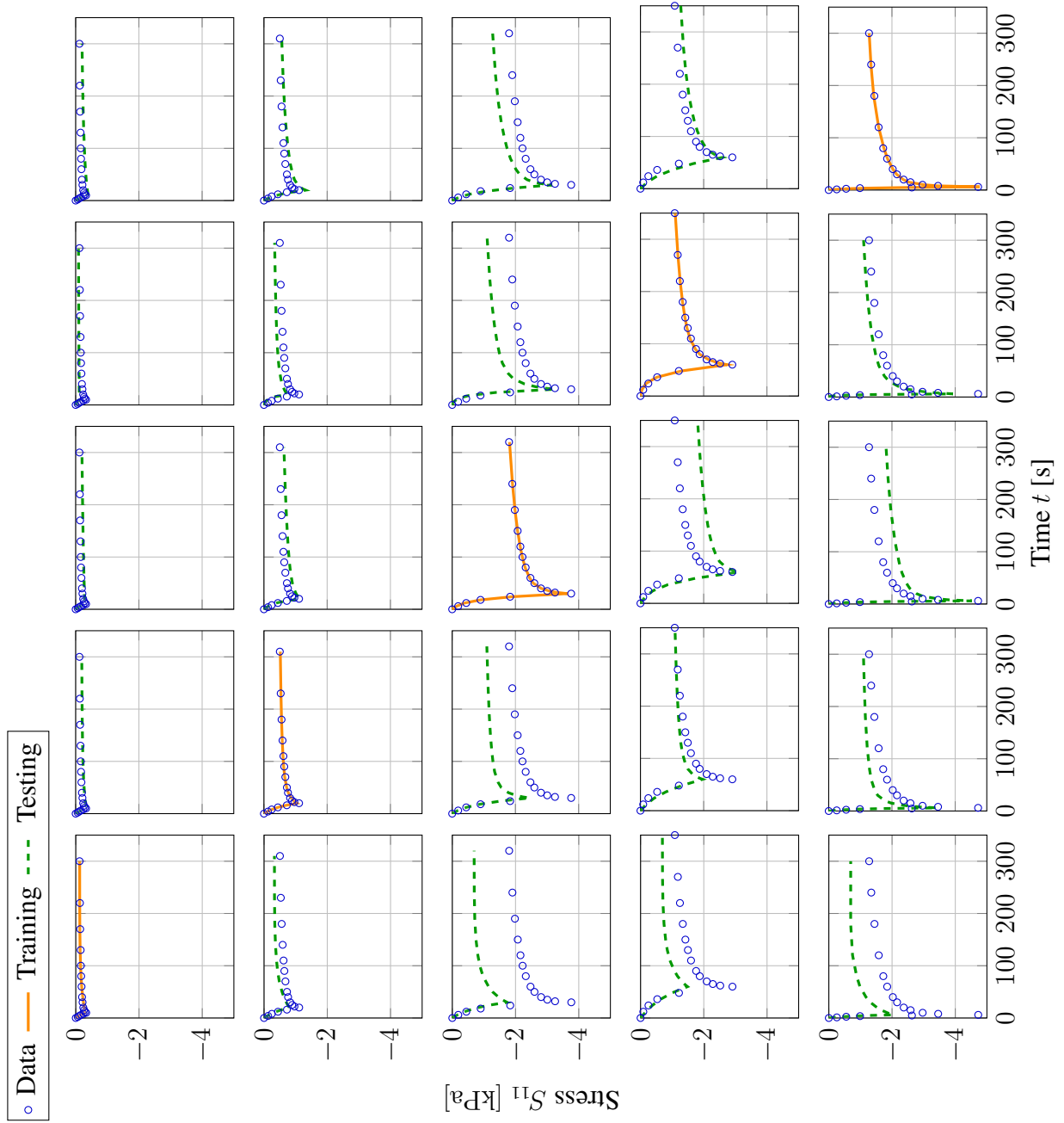
\begin{table}[H]
\centering
\caption{Normalized root mean squared error, $\epsilon$, corresponding to the results shown in Figure~\ref{fig:TrainOne}. Each column represents a combination of training and testing, while each row represents one experimental setup. The colored boxes indicate the experiment used for training, the remaining four tests are utilized for testing.}
\label{tab:TrainOneEpsilon}
\begin{tabular}{ c || r r r r r}
\hline
		\multirow{5}{*}{\shortstack[c]{Stress\\ $S_{11}\ \lbrack \rm kPa \rbrack$}}	& \cellcolor{orange}0.09 & 0.32 & 0.34 & 0.60 & 0.54 \\
		& 0.28 & \cellcolor{orange}0.04 & 0.22 & 0.35 & 0.26 \\
		& 0.61 & 0.41 & \cellcolor{orange}0.02 & 0.28 & 0.21 \\
		& 0.48 & 0.25 & 0.41 & \cellcolor{orange}0.04 & 0.17 \\
		& 0.55 & 0.28 & 0.22 & 0.21 & \cellcolor{orange}0.04 \\		
\hline\hline
$\epsilon$ &	\multicolumn{5}{c}{Time $t$ [s]} \\
\hline
\end{tabular}%
\end{table}
\begin{table}[H]
\centering
\caption{Coefficient of determination, $R^2$, corresponding to the results shown in Figure~\ref{fig:TrainOne}. Each column represents a combination of training and testing, while each row represents one experimental setup. The colored boxes indicate the experiment used for training, the remaining four tests are utilized for testing.}
\label{tab:TrainOneR2}
\begin{tabular}{ c || r r r r r}
\hline
		\multirow{5}{*}{\shortstack[c]{Stress\\ $S_{11}\ \lbrack \rm kPa \rbrack$}}	& \cellcolor{orange}0.96 & 0.47 & 0.40 & 0.00 & 0.00 \\
		& 0.60 & \cellcolor{orange}0.99 & 0.76 & 0.39 & 0.66 \\
		& 0.00 & 0.33 & \cellcolor{orange}1.00 & 0.68 & 0.82 \\
		& 0.29 & 0.79 & 0.49 & \cellcolor{orange}0.99 & 0.91 \\
		& 0.16 & 0.78 & 0.87 & 0.88 & \cellcolor{orange}1.00 \\
\hline\hline
$R^2$ &	\multicolumn{5}{c}{Time $t$ [s]} \\
\hline
\end{tabular}%
\end{table}
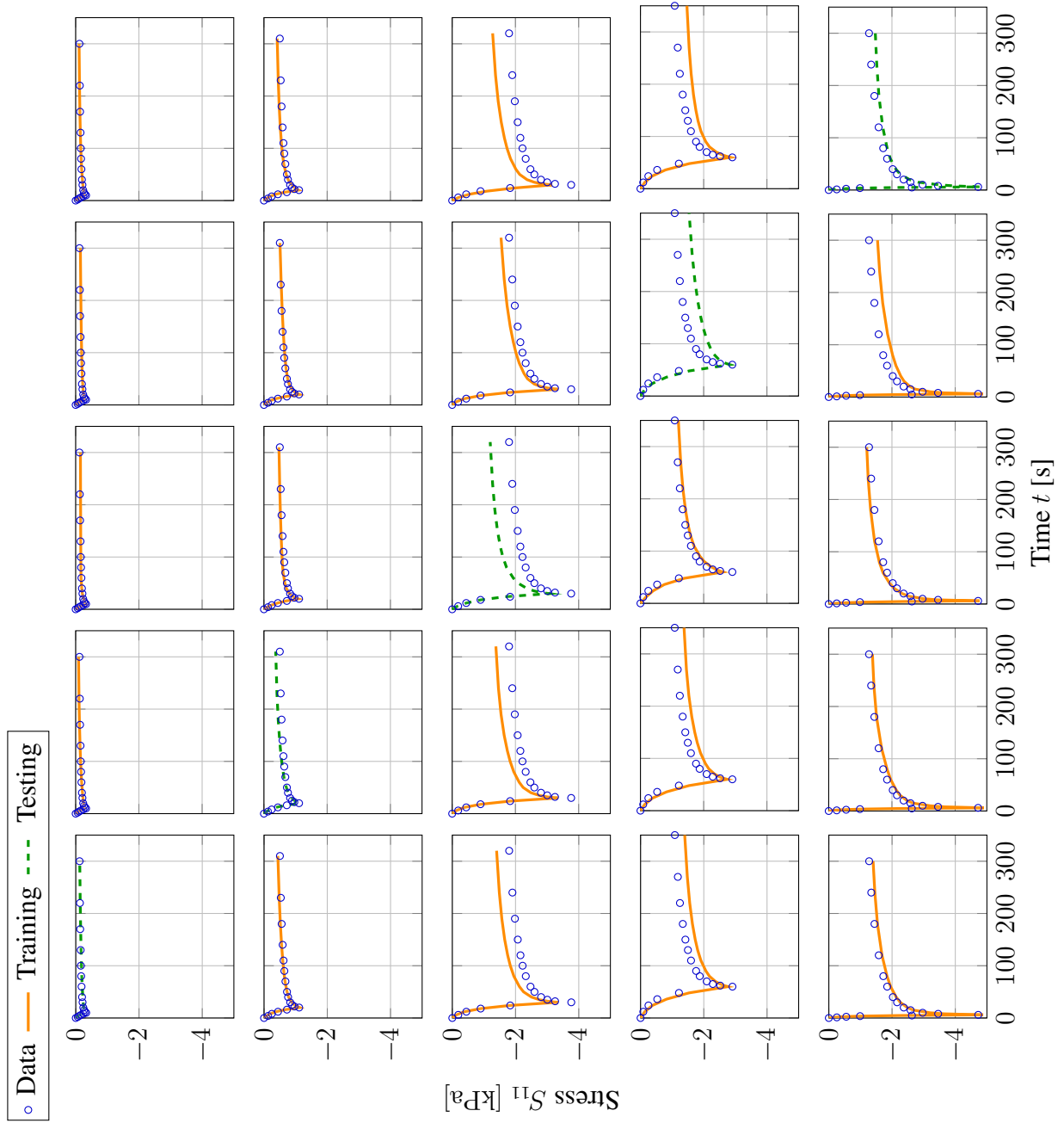
\begin{figure}[h]
    \centering
    \vspace*{-0mm}
    \begin{subfigure}{.05\textwidth} 
    		\vspace*{-25mm}
				\begin{tikzpicture}[rotate=90,transform shape]
			\begin{axis}[xlabel= Nice $x$ label, ylabel= Nice $y$ label, width=55mm, height=35mm,
				hide axis,
				xmin=10,
				xmax=50,
				ymin=0,
				ymax=0.4,
				legend columns=-1,
				legend style={column sep=1mm}
				]
					\addlegendimage{color = blue,fill = white,
    		 				fill opacity=0.1, 
	    		 			only marks, mark size=1.5pt,
						error bars/.cd,
						y dir = both,
						y explicit,
						error bar style={line width=0.5pt}, 
						error mark options={line width=0.5pt, mark size=2pt, rotate=90}}
					\addlegendentry{Data}
					\addlegendimage{orange, very thick}
					\addlegendentry{Training} 
					\addlegendimage{green, dashed, very thick}
					\addlegendentry{Testing} 
			\end{axis}
		\end{tikzpicture} 
	\end{subfigure}%
    \begin{subfigure}{.18\textwidth} 
		\begin{tikzpicture}[rotate=90,transform shape]
            \begin{axis}[grid = major, width=44mm, height=40mm, xmin=0,xmax=350,ymin=-5,ymax=0,xticklabels=\empty,ytick={0,-2,-4},yticklabels={\vphantom{0},\vphantom{-2},\vphantom{-4}}]
                \addplot[color = blue,fill = white,
    		 		fill opacity=0.1, 
    		 		only marks, mark size=1.5pt,
				error bars/.cd,
				y dir = both,
				y explicit,
				error bar style={line width=0.5pt}, 
				error mark options={line width=0.5pt, mark size=2pt, rotate=90}
				] table[x expr=\thisrowno{0}, y expr=\thisrowno{5}]{graphs/section06/VanLoocke2008/VanLoocke_rate1_zsm.txt};
                \addplot[orange, very thick] table[x expr=\thisrowno{0}, y expr=\thisrowno{10}]{graphs/section06/VanLoocke2008/TrainFour/VanLoocke_rate1_zsm.txt};
            \end{axis}
        \end{tikzpicture}
        \begin{tikzpicture}[rotate=90,transform shape]
            \begin{axis}[grid = major, width=44mm, height=40mm, xmin=0,xmax=350,ymin=-5,ymax=0,xticklabels=\empty,ytick={0,-2,-4},yticklabels={\vphantom{0},\vphantom{-2},\vphantom{-4}}]
                \addplot[color = blue,fill = white,
    		 		fill opacity=0.1, 
    		 		only marks, mark size=1.5pt,
				error bars/.cd,
				y dir = both,
				y explicit,
				error bar style={line width=0.5pt}, 
				error mark options={line width=0.5pt, mark size=2pt, rotate=90}
				] table[x expr=\thisrowno{0}, y expr=\thisrowno{5}]{graphs/section06/VanLoocke2008/TrainFour/VanLoocke_rate1_zsm.txt};
                \addplot[orange, very thick] table[x expr=\thisrowno{0}, y expr=\thisrowno{9}]{graphs/section06/VanLoocke2008/TrainFour/VanLoocke_rate1_zsm.txt};
            \end{axis}
        \end{tikzpicture}
        \begin{tikzpicture}[rotate=90,transform shape]
            \begin{axis}[grid = major, width=44mm, height=40mm, xmin=0,xmax=350,ymin=-5,ymax=0,xticklabels=\empty,ytick={0,-2,-4},yticklabels={\vphantom{0},\vphantom{-2},\vphantom{-4}}]
                \addplot[color = blue,fill = white,
    		 		fill opacity=0.1, 
    		 		only marks, mark size=1.5pt,
				error bars/.cd,
				y dir = both,
				y explicit,
				error bar style={line width=0.5pt}, 
				error mark options={line width=0.5pt, mark size=2pt, rotate=90}
				] table[x expr=\thisrowno{0}, y expr=\thisrowno{5}]{graphs/section06/VanLoocke2008/TrainFour/VanLoocke_rate1_zsm.txt};
                \addplot[orange, very thick] table[x expr=\thisrowno{0}, y expr=\thisrowno{8}]{graphs/section06/VanLoocke2008/TrainFour/VanLoocke_rate1_zsm.txt};
            \end{axis}
        \end{tikzpicture}
        \begin{tikzpicture}[rotate=90,transform shape]
            \begin{axis}[grid = major, width=44mm, height=40mm, xmin=0,xmax=350,ymin=-5,ymax=0,xticklabels=\empty,ytick={0,-2,-4},yticklabels={\vphantom{0},\vphantom{-2},\vphantom{-4}}]
                \addplot[color = blue,fill = white,
    		 		fill opacity=0.1, 
    		 		only marks, mark size=1.5pt,
				error bars/.cd,
				y dir = both,
				y explicit,
				error bar style={line width=0.5pt}, 
				error mark options={line width=0.5pt, mark size=2pt, rotate=90}
				] table[x expr=\thisrowno{0}, y expr=\thisrowno{5}]{graphs/section06/VanLoocke2008/TrainFour/VanLoocke_rate1_zsm.txt};
                \addplot[orange, very thick] table[x expr=\thisrowno{0}, y expr=\thisrowno{7}]{graphs/section06/VanLoocke2008/TrainFour/VanLoocke_rate1_zsm.txt};
            \end{axis}
        \end{tikzpicture}
        \begin{tikzpicture}[rotate=90,transform shape]
            \begin{axis}[grid = major, width=44mm, height=40mm, xmin=0,xmax=350,ymin=-5,ymax=0,xticklabels=\empty,ylabel={\phantom{Stress $S_{11}\ \lbrack \rm kPa \rbrack$}}]
                \addplot[color = blue,fill = white,
    		 		fill opacity=0.1, 
    		 		only marks, mark size=1.5pt,
				error bars/.cd,
				y dir = both,
				y explicit,
				error bar style={line width=0.5pt}, 
				error mark options={line width=0.5pt, mark size=2pt, rotate=90}
				] table[x expr=\thisrowno{0}, y expr=\thisrowno{5}]{graphs/section06/VanLoocke2008/TrainFour/VanLoocke_rate1_zsm.txt};
                \addplot[green, dashed, very thick] table[x expr=\thisrowno{0}, y expr=\thisrowno{6}]{graphs/section06/VanLoocke2008/TrainFour/VanLoocke_rate1_zsm.txt};
            \end{axis}
        \end{tikzpicture} 
    \end{subfigure}%
    \begin{subfigure}{.18\textwidth} 
		\begin{tikzpicture}[rotate=90,transform shape]
            \begin{axis}[grid = major, width=44mm, height=40mm, xmin=0,xmax=350,ymin=-5,ymax=0,xticklabels=\empty,ytick={0,-2,-4},yticklabels={\vphantom{0},\vphantom{-2},\vphantom{-4}}]
                \addplot[color = blue,fill = white,
    		 		fill opacity=0.1, 
    		 		only marks, mark size=1.5pt,
				error bars/.cd,
				y dir = both,
				y explicit,
				error bar style={line width=0.5pt}, 
				error mark options={line width=0.5pt, mark size=2pt, rotate=90}
				] table[x expr=\thisrowno{0}, y expr=\thisrowno{5}]{graphs/section06/VanLoocke2008/TrainFour/VanLoocke_rate2_zsm.txt};
                \addplot[orange, very thick] table[x expr=\thisrowno{0}, y expr=\thisrowno{10}]{graphs/section06/VanLoocke2008/TrainFour/VanLoocke_rate2_zsm.txt};
            \end{axis}
        \end{tikzpicture}
        \begin{tikzpicture}[rotate=90,transform shape]
            \begin{axis}[grid = major, width=44mm, height=40mm, xmin=0,xmax=350,ymin=-5,ymax=0,xticklabels=\empty,ytick={0,-2,-4},yticklabels={\vphantom{0},\vphantom{-2},\vphantom{-4}}]
                \addplot[color = blue,fill = white,
    		 		fill opacity=0.1, 
    		 		only marks, mark size=1.5pt,
				error bars/.cd,
				y dir = both,
				y explicit,
				error bar style={line width=0.5pt}, 
				error mark options={line width=0.5pt, mark size=2pt, rotate=90}
				] table[x expr=\thisrowno{0}, y expr=\thisrowno{5}]{graphs/section06/VanLoocke2008/TrainFour/VanLoocke_rate2_zsm.txt};
                \addplot[orange, very thick] table[x expr=\thisrowno{0}, y expr=\thisrowno{9}]{graphs/section06/VanLoocke2008/TrainFour/VanLoocke_rate2_zsm.txt};
            \end{axis}
        \end{tikzpicture}
        \begin{tikzpicture}[rotate=90,transform shape]
            \begin{axis}[grid = major, width=44mm, height=40mm, xmin=0,xmax=350,ymin=-5,ymax=0,xticklabels=\empty,ytick={0,-2,-4},yticklabels={\vphantom{0},\vphantom{-2},\vphantom{-4}}]
                \addplot[color = blue,fill = white,
    		 		fill opacity=0.1, 
    		 		only marks, mark size=1.5pt,
				error bars/.cd,
				y dir = both,
				y explicit,
				error bar style={line width=0.5pt}, 
				error mark options={line width=0.5pt, mark size=2pt, rotate=90}
				] table[x expr=\thisrowno{0}, y expr=\thisrowno{5}]{graphs/section06/VanLoocke2008/TrainFour/VanLoocke_rate2_zsm.txt};
                \addplot[orange, very thick] table[x expr=\thisrowno{0}, y expr=\thisrowno{8}]{graphs/section06/VanLoocke2008/TrainFour/VanLoocke_rate2_zsm.txt};
            \end{axis}
        \end{tikzpicture}
        \begin{tikzpicture}[rotate=90,transform shape]
            \begin{axis}[grid = major, width=44mm, height=40mm, xmin=0,xmax=350,ymin=-5,ymax=0,xticklabels=\empty,ytick={0,-2,-4},yticklabels={\vphantom{0},\vphantom{-2},\vphantom{-4}}]
                \addplot[color = blue,fill = white,
    		 		fill opacity=0.1, 
    		 		only marks, mark size=1.5pt,
				error bars/.cd,
				y dir = both,
				y explicit,
				error bar style={line width=0.5pt}, 
				error mark options={line width=0.5pt, mark size=2pt, rotate=90}
				] table[x expr=\thisrowno{0}, y expr=\thisrowno{5}]{graphs/section06/VanLoocke2008/TrainFour/VanLoocke_rate2_zsm.txt};
                \addplot[green, dashed, very thick] table[x expr=\thisrowno{0}, y expr=\thisrowno{7}]{graphs/section06/VanLoocke2008/TrainFour/VanLoocke_rate2_zsm.txt};
            \end{axis}
        \end{tikzpicture}
        \begin{tikzpicture}[rotate=90,transform shape]
            \begin{axis}[grid = major, width=44mm, height=40mm, xmin=0,xmax=350,ymin=-5,ymax=0,xticklabels=\empty,ylabel={\phantom{Stress $S_{11}\ \lbrack \rm kPa \rbrack$}}]
                \addplot[color = blue,fill = white,
    		 		fill opacity=0.1, 
    		 		only marks, mark size=1.5pt,
				error bars/.cd,
				y dir = both,
				y explicit,
				error bar style={line width=0.5pt}, 
				error mark options={line width=0.5pt, mark size=2pt, rotate=90}
				] table[x expr=\thisrowno{0}, y expr=\thisrowno{5}]{graphs/section06/VanLoocke2008/TrainFour/VanLoocke_rate2_zsm.txt};
                \addplot[orange, very thick] table[x expr=\thisrowno{0}, y expr=\thisrowno{6}]{graphs/section06/VanLoocke2008/TrainFour/VanLoocke_rate2_zsm.txt};
            \end{axis}
        \end{tikzpicture} 
    \end{subfigure}%
    \begin{subfigure}{.18\textwidth} 
		\begin{tikzpicture}[rotate=90,transform shape]
            \begin{axis}[grid = major, width=44mm, height=40mm, xmin=0,xmax=350,ymin=-5,ymax=0,xticklabels=\empty,ytick={0,-2,-4},yticklabels={\vphantom{0},\vphantom{-2},\vphantom{-4}}]
                \addplot[color = blue,fill = white,
    		 		fill opacity=0.1, 
    		 		only marks, mark size=1.5pt,
				error bars/.cd,
				y dir = both,
				y explicit,
				error bar style={line width=0.5pt}, 
				error mark options={line width=0.5pt, mark size=2pt, rotate=90}
				] table[x expr=\thisrowno{0}, y expr=\thisrowno{5}]{graphs/section06/VanLoocke2008/VanLoocke_rate3_zsm.txt};
                \addplot[orange, very thick] table[x expr=\thisrowno{0}, y expr=\thisrowno{10}]{graphs/section06/VanLoocke2008/VanLoocke_rate3_zsm.txt};
            \end{axis}
        \end{tikzpicture}
        \begin{tikzpicture}[rotate=90,transform shape]
            \begin{axis}[grid = major, width=44mm, height=40mm, xmin=0,xmax=350,ymin=-5,ymax=0,xticklabels=\empty,ytick={0,-2,-4},yticklabels={\vphantom{0},\vphantom{-2},\vphantom{-4}}]
                \addplot[color = blue,fill = white,
    		 		fill opacity=0.1, 
    		 		only marks, mark size=1.5pt,
				error bars/.cd,
				y dir = both,
				y explicit,
				error bar style={line width=0.5pt}, 
				error mark options={line width=0.5pt, mark size=2pt, rotate=90}
				] table[x expr=\thisrowno{0}, y expr=\thisrowno{5}]{graphs/section06/VanLoocke2008/TrainFour/VanLoocke_rate3_zsm.txt};
                \addplot[orange, very thick] table[x expr=\thisrowno{0}, y expr=\thisrowno{9}]{graphs/section06/VanLoocke2008/TrainFour/VanLoocke_rate3_zsm.txt};
            \end{axis}
        \end{tikzpicture}
        \begin{tikzpicture}[rotate=90,transform shape]
            \begin{axis}[grid = major, width=44mm, height=40mm, xmin=0,xmax=350,ymin=-5,ymax=0,xticklabels=\empty,ytick={0,-2,-4},yticklabels={\vphantom{0},\vphantom{-2},\vphantom{-4}}]
                \addplot[color = blue,fill = white,
    		 		fill opacity=0.1, 
    		 		only marks, mark size=1.5pt,
				error bars/.cd,
				y dir = both,
				y explicit,
				error bar style={line width=0.5pt}, 
				error mark options={line width=0.5pt, mark size=2pt, rotate=90}
				] table[x expr=\thisrowno{0}, y expr=\thisrowno{5}]{graphs/section06/VanLoocke2008/TrainFour/VanLoocke_rate3_zsm.txt};
                \addplot[green, dashed, very thick] table[x expr=\thisrowno{0}, y expr=\thisrowno{8}]{graphs/section06/VanLoocke2008/TrainFour/VanLoocke_rate3_zsm.txt};
            \end{axis}
        \end{tikzpicture}
        \begin{tikzpicture}[rotate=90,transform shape]
            \begin{axis}[grid = major, width=44mm, height=40mm, xmin=0,xmax=350,ymin=-5,ymax=0,xticklabels=\empty,ytick={0,-2,-4},yticklabels={\vphantom{0},\vphantom{-2},\vphantom{-4}}]
                \addplot[color = blue,fill = white,
    		 		fill opacity=0.1, 
    		 		only marks, mark size=1.5pt,
				error bars/.cd,
				y dir = both,
				y explicit,
				error bar style={line width=0.5pt}, 
				error mark options={line width=0.5pt, mark size=2pt, rotate=90}
				] table[x expr=\thisrowno{0}, y expr=\thisrowno{5}]{graphs/section06/VanLoocke2008/TrainFour/VanLoocke_rate3_zsm.txt};
                \addplot[orange, very thick] table[x expr=\thisrowno{0}, y expr=\thisrowno{7}]{graphs/section06/VanLoocke2008/TrainFour/VanLoocke_rate3_zsm.txt};
            \end{axis}
        \end{tikzpicture}
        \begin{tikzpicture}[rotate=90,transform shape]
            \begin{axis}[grid = major, width=44mm, height=40mm, xmin=0,xmax=350,ymin=-5,ymax=0,xticklabels=\empty,ylabel={Stress $S_{11}\ \lbrack \rm kPa \rbrack$}]
                \addplot[color = blue,fill = white,
    		 		fill opacity=0.1, 
    		 		only marks, mark size=1.5pt,
				error bars/.cd,
				y dir = both,
				y explicit,
				error bar style={line width=0.5pt}, 
				error mark options={line width=0.5pt, mark size=2pt, rotate=90}
				] table[x expr=\thisrowno{0}, y expr=\thisrowno{5}]{graphs/section06/VanLoocke2008/TrainFour/VanLoocke_rate3_zsm.txt};
                \addplot[orange, very thick] table[x expr=\thisrowno{0}, y expr=\thisrowno{6}]{graphs/section06/VanLoocke2008/TrainFour/VanLoocke_rate3_zsm.txt};
            \end{axis}
        \end{tikzpicture} 
    \end{subfigure}%
    \begin{subfigure}{.18\textwidth} 
		\begin{tikzpicture}[rotate=90,transform shape]
            \begin{axis}[grid = major, width=44mm, height=40mm, xmin=0,xmax=350,ymin=-5,ymax=0,xticklabels=\empty,ytick={0,-2,-4},yticklabels={\vphantom{0},\vphantom{-2},\vphantom{-4}}]
                \addplot[color = blue,fill = white,
    		 		fill opacity=0.1, 
    		 		only marks, mark size=1.5pt,
				error bars/.cd,
				y dir = both,
				y explicit,
				error bar style={line width=0.5pt}, 
				error mark options={line width=0.5pt, mark size=2pt, rotate=90}
				] table[x expr=\thisrowno{0}, y expr=\thisrowno{5}]{graphs/section06/VanLoocke2008/TrainFour/VanLoocke_rate4_zsm.txt};
                \addplot[orange, very thick] table[x expr=\thisrowno{0}, y expr=\thisrowno{10}]{graphs/section06/VanLoocke2008/TrainFour/VanLoocke_rate4_zsm.txt};
            \end{axis}
        \end{tikzpicture}
        \begin{tikzpicture}[rotate=90,transform shape]
            \begin{axis}[grid = major, width=44mm, height=40mm, xmin=0,xmax=350,ymin=-5,ymax=0,xticklabels=\empty,ytick={0,-2,-4},yticklabels={\vphantom{0},\vphantom{-2},\vphantom{-4}}]
                \addplot[color = blue,fill = white,
    		 		fill opacity=0.1, 
    		 		only marks, mark size=1.5pt,
				error bars/.cd,
				y dir = both,
				y explicit,
				error bar style={line width=0.5pt}, 
				error mark options={line width=0.5pt, mark size=2pt, rotate=90}
				] table[x expr=\thisrowno{0}, y expr=\thisrowno{5}]{graphs/section06/VanLoocke2008/TrainFour/VanLoocke_rate4_zsm.txt};
                \addplot[green, dashed, very thick] table[x expr=\thisrowno{0}, y expr=\thisrowno{9}]{graphs/section06/VanLoocke2008/TrainFour/VanLoocke_rate4_zsm.txt};
            \end{axis}
        \end{tikzpicture}
        \begin{tikzpicture}[rotate=90,transform shape]
            \begin{axis}[grid = major, width=44mm, height=40mm, xmin=0,xmax=350,ymin=-5,ymax=0,xticklabels=\empty,ytick={0,-2,-4},yticklabels={\vphantom{0},\vphantom{-2},\vphantom{-4}}]
                \addplot[color = blue,fill = white,
    		 		fill opacity=0.1, 
    		 		only marks, mark size=1.5pt,
				error bars/.cd,
				y dir = both,
				y explicit,
				error bar style={line width=0.5pt}, 
				error mark options={line width=0.5pt, mark size=2pt, rotate=90}
				] table[x expr=\thisrowno{0}, y expr=\thisrowno{5}]{graphs/section06/VanLoocke2008/TrainFour/VanLoocke_rate4_zsm.txt};
                \addplot[orange, very thick] table[x expr=\thisrowno{0}, y expr=\thisrowno{8}]{graphs/section06/VanLoocke2008/TrainFour/VanLoocke_rate4_zsm.txt};
            \end{axis}
        \end{tikzpicture}
        \begin{tikzpicture}[rotate=90,transform shape]
            \begin{axis}[grid = major, width=44mm, height=40mm, xmin=0,xmax=350,ymin=-5,ymax=0,xticklabels=\empty,ytick={0,-2,-4},yticklabels={\vphantom{0},\vphantom{-2},\vphantom{-4}}]
                \addplot[color = blue,fill = white,
    		 		fill opacity=0.1, 
    		 		only marks, mark size=1.5pt,
				error bars/.cd,
				y dir = both,
				y explicit,
				error bar style={line width=0.5pt}, 
				error mark options={line width=0.5pt, mark size=2pt, rotate=90}
				] table[x expr=\thisrowno{0}, y expr=\thisrowno{5}]{graphs/section06/VanLoocke2008/TrainFour/VanLoocke_rate4_zsm.txt};
                \addplot[orange, very thick] table[x expr=\thisrowno{0}, y expr=\thisrowno{7}]{graphs/section06/VanLoocke2008/TrainFour/VanLoocke_rate4_zsm.txt};
            \end{axis}
        \end{tikzpicture}
        \begin{tikzpicture}[rotate=90,transform shape]
            \begin{axis}[grid = major, width=44mm, height=40mm, xmin=0,xmax=350,ymin=-5,ymax=0,xticklabels=\empty,ylabel={\phantom{Stress $S_{11}\ \lbrack \rm kPa \rbrack$}}]
                \addplot[color = blue,fill = white,
    		 		fill opacity=0.1, 
    		 		only marks, mark size=1.5pt,
				error bars/.cd,
				y dir = both,
				y explicit,
				error bar style={line width=0.5pt}, 
				error mark options={line width=0.5pt, mark size=2pt, rotate=90}
				] table[x expr=\thisrowno{0}, y expr=\thisrowno{5}]{graphs/section06/VanLoocke2008/TrainFour/VanLoocke_rate4_zsm.txt};
                \addplot[orange, very thick] table[x expr=\thisrowno{0}, y expr=\thisrowno{6}]{graphs/section06/VanLoocke2008/TrainFour/VanLoocke_rate4_zsm.txt};
            \end{axis}
        \end{tikzpicture} 
    \end{subfigure}%
    \begin{subfigure}{.18\textwidth} 
        \begin{tikzpicture}[rotate=90,transform shape]
            \begin{axis}[grid = major, width=44mm, height=40mm, xmin=0,xmax=350,ymin=-5,ymax=0,ytick={0,-2,-4},yticklabels={\vphantom{0},\vphantom{-2},\vphantom{-4}}]
                \addplot[color = blue,fill = white,
    		 		fill opacity=0.1, 
    		 		only marks, mark size=1.5pt,
				error bars/.cd,
				y dir = both,
				y explicit,
				error bar style={line width=0.5pt}, 
				error mark options={line width=0.5pt, mark size=2pt, rotate=90}
				] table[x expr=\thisrowno{0}, y expr=\thisrowno{5}]{graphs/section06/VanLoocke2008/TrainFour/VanLoocke_rate5_zsm.txt};
                \addplot[green, dashed, very thick] table[x expr=\thisrowno{0}, y expr=\thisrowno{10}]{graphs/section06/VanLoocke2008/TrainFour/VanLoocke_rate5_zsm.txt};
            \end{axis}
        \end{tikzpicture}
        \begin{tikzpicture}[rotate=90,transform shape]
            \begin{axis}[grid = major, width=44mm, height=40mm, xmin=0,xmax=350,ymin=-5,ymax=0,ytick={0,-2,-4},yticklabels={\vphantom{0},\vphantom{-2},\vphantom{-4}}]
                \addplot[color = blue,fill = white,
    		 		fill opacity=0.1, 
    		 		only marks, mark size=1.5pt,
				error bars/.cd,
				y dir = both,
				y explicit,
				error bar style={line width=0.5pt}, 
				error mark options={line width=0.5pt, mark size=2pt, rotate=90}
				] table[x expr=\thisrowno{0}, y expr=\thisrowno{5}]{graphs/section06/VanLoocke2008/TrainFour/VanLoocke_rate5_zsm.txt};
                \addplot[orange, very thick] table[x expr=\thisrowno{0}, y expr=\thisrowno{9}]{graphs/section06/VanLoocke2008/TrainFour/VanLoocke_rate5_zsm.txt};
            \end{axis}
        \end{tikzpicture}
        \begin{tikzpicture}[rotate=90,transform shape]
            \begin{axis}[grid = major, width=44mm, height=40mm, xmin=0,xmax=350,ymin=-5,ymax=0,ytick={0,-2,-4},yticklabels={\vphantom{0},\vphantom{-2},\vphantom{-4}},xlabel={Time $t$ [s]}]
                \addplot[color = blue,fill = white,
    		 		fill opacity=0.1, 
    		 		only marks, mark size=1.5pt,
				error bars/.cd,
				y dir = both,
				y explicit,
				error bar style={line width=0.5pt}, 
				error mark options={line width=0.5pt, mark size=2pt, rotate=90}
				] table[x expr=\thisrowno{0}, y expr=\thisrowno{5}]{graphs/section06/VanLoocke2008/TrainFour/VanLoocke_rate5_zsm.txt};
                \addplot[orange, very thick] table[x expr=\thisrowno{0}, y expr=\thisrowno{8}]{graphs/section06/VanLoocke2008/TrainFour/VanLoocke_rate5_zsm.txt};
            \end{axis}
        \end{tikzpicture}
        \begin{tikzpicture}[rotate=90,transform shape]
            \begin{axis}[grid = major, width=44mm, height=40mm, xmin=0,xmax=350,ymin=-5,ymax=0,ytick={0,-2,-4},yticklabels={\vphantom{0},\vphantom{-2},\vphantom{-4}}]
                \addplot[color = blue,fill = white,
    		 		fill opacity=0.1, 
    		 		only marks, mark size=1.5pt,
				error bars/.cd,
				y dir = both,
				y explicit,
				error bar style={line width=0.5pt}, 
				error mark options={line width=0.5pt, mark size=2pt, rotate=90}
				] table[x expr=\thisrowno{0}, y expr=\thisrowno{5}]{graphs/section06/VanLoocke2008/TrainFour/VanLoocke_rate5_zsm.txt};
                \addplot[orange, very thick] table[x expr=\thisrowno{0}, y expr=\thisrowno{7}]{graphs/section06/VanLoocke2008/TrainFour/VanLoocke_rate5_zsm.txt};
            \end{axis}
        \end{tikzpicture}
        \begin{tikzpicture}[rotate=90,transform shape]
            \begin{axis}[grid = major, width=44mm, height=40mm, xmin=0,xmax=350,ymin=-5,ymax=0,ytick={0,-2,-4},ylabel={\phantom{Stress $S_{11}\ \lbrack \rm kPa \rbrack$}}]
                \addplot[color = blue,fill = white,
    		 		fill opacity=0.1, 
    		 		only marks, mark size=1.5pt,
				error bars/.cd,
				y dir = both,
				y explicit,
				error bar style={line width=0.5pt}, 
				error mark options={line width=0.5pt, mark size=2pt, rotate=90}
				] table[x expr=\thisrowno{0}, y expr=\thisrowno{5}]{graphs/section06/VanLoocke2008/TrainFour/VanLoocke_rate5_zsm.txt};
                \addplot[orange, very thick] table[x expr=\thisrowno{0}, y expr=\thisrowno{6}]{graphs/section06/VanLoocke2008/TrainFour/VanLoocke_rate5_zsm.txt};
            \end{axis}
        \end{tikzpicture} 
    \end{subfigure}
\caption{\textbf{Train on four, test on one.} Experimental data and discovered model for passive skeletal muscle taken from \cite{vanloocke2008}. Each column represents a combination of training and testing, while each row represents one experimental setup. The experimental setup varies with respect to the maximum applied stretch $C_{11}^{\mathrm{max}}=\{0.81,0.64,0.49,0.49,0.49\}\,\left[-\right]$ and the rate of deformation $\dot{F}_{11}=\{0.01,0.01,0.01,0.005,0.05\}\,\left[ s \right]$. The columns vary with respect to the four experiments used for training.}
\label{fig:TrainFour}
\end{figure}
\begin{table}[H]
\centering
\caption{Normalized root mean squared error, $\epsilon$, corresponding to the results shown in Figure~\ref{fig:TrainFour}. Each column represents a combination of training and testing, while each row represents one experimental setup. The colored boxes indicate the experiment used for testing, the remaining four tests are utilized for training.}
\label{tab:TrainFourEpsilon}
\begin{tabular}{ c || r r r r r}
\hline
		\multirow{5}{*}{\shortstack[c]{Stress\\ $S_{11}\ \lbrack \rm kPa \rbrack$}}	& \cellcolor{green}0.15 & 0.14 & 0.07 & 0.17 & 0.13 \\
		& 0.08 & \cellcolor{green}0.12 & 0.06 & 0.07 & 0.10 \\
		& 0.16 & 0.16 & \cellcolor{green}0.25 & 0.10 & 0.14 \\
		& 0.19 & 0.18 & 0.10 & \cellcolor{green}0.27 & 0.21 \\
		& 0.08 & 0.08 & 0.05 & 0.12 & \cellcolor{green}0.12 \\		
\hline\hline
$\epsilon$ &	\multicolumn{5}{c}{Time $t$ [s]} \\
\hline
\end{tabular}%
\end{table}
\begin{table}[H]
\centering
\caption{Coefficient of determination, $R^2$, corresponding to the results shown in Figure~\ref{fig:TrainFour}. Each column represents a combination of training and testing, while each row represents one experimental setup. The colored boxes indicate the experiment used for testing, the remaining four tests are utilized for training.}
\label{tab:TrainFourR2}
\begin{tabular}{ c || r r r r r}
\hline
		\multirow{5}{*}{\shortstack[c]{Stress\\ $S_{11}\ \lbrack \rm kPa \rbrack$}}	& \cellcolor{green}0.89 & 0.90 & 0.97 & 0.85 & 0.91 \\
		& 0.97 & \cellcolor{green}0.93 & 0.98 & 0.98 & 0.95 \\
		& 0.90 & 0.90 & \cellcolor{green}0.76 & 0.96 & 0.93 \\
		& 0.89 & 0.90 & 0.97 & \cellcolor{green}0.77 & 0.87 \\
		& 0.98 & 0.98 & 0.99 & 0.96 & \cellcolor{green}0.96 \\ 
\hline\hline
$R^2$ &	\multicolumn{5}{c}{Time $t$ [s]} \\
\hline
\end{tabular}%
\end{table}
\section{Discussion and limitations}
\label{sec:limits}
In this paper, we have designed a general inelastic constitutive artificial neural network that satisfies thermodynamics for all kinds of inelastic phenomena that can be accounted for by the multiplicative decomposition.
To investigate our proposed approach, we limited our study to the effect of visco-elasticity.
In addition, we focused on a thermodynamically sound and flexible theoretical description of the network rather than on aspects related to the machine learning algorithms themselves.
Of course, this leads to some limitations of this work and leaves room for future investigations, some of which we would like to briefly touch upon.

\textbf{iCANNs are able to discover a model for visco-elasticity.} Our results show that our extension of CANNs by multiplicative decomposition and pseudo potentials is a suitable approach for viscoelastic materials.
The sparse amount of data does not hinder the successful discovery.
Since our pseudo potential, and especially the activation functions, are chosen in a general way that satisfies the mathematical requirements of thermodynamics, rather than inspired by existing models in the literature, it would be interesting to compare pseudo potentials used in constitutive modeling to describe certain materials with those discovered by iCANN.
Furthermore, we have fixed the number of Maxwell elements to three.
In future studies, it should be investigated how to design an iCANN that learns whether the number either can be decreased or must be increased.

\textbf{Are our discovered iCANN models unique?} Although we found several models that explain the experimental data well and also perform extremely well in tests compared to standard neural networks, we did not investigate the uniqueness of our solution.
Thus, it is possible that fewer terms in both the Helmholtz free energy and the pseudo potential are needed to explain the data with the same accuracy.
To avoid possible overfitting, it should be investigated to what extent the optimizer's regularization can help us achieve this goal, as has already been done in the literature, e.g. by \cite{wang2023}, \cite{stpierre2023}, and \cite{mcculloch2023arXiv} for the Helmholtz free energy of CANNs.

\textbf{Specialization to further inelastic phenomena.} The network's architecture is formulated in a general way being applicable to almost all kind of inelastic effects.
To investigate the performance of the iCANN, we limited ourselves to the case of visco-elasticity.
Future works should study the ability to explain the data of, for instance, elasto-(visco)plasticity.
Such a phenomenon includes some kind of yield criterion, which distinguishes the elastic from the inelastic regime.
Having in mind that one might need multi-surface models, i.e., several parallel arrangements of iCANNs similar to using several Maxwell elements, we ask ourselves what is the best implementation strategy.
In order to keep the numerical effort during training low and the implementation as simple as possible, we would suggest to keep an explicit time integration, while the equality and/or inequality constrains are satisfied by Fischer-Burmeister approach (see e.g. \cite{fischer1992}).
For example, such an approach was successfully applied in constitutive modeling by \cite{kiefer2012} and \cite{brepols2017}.

\textbf{Initial and induced anisotropy.} We have not considered anisotropy.
However, some materials exhibit a pronounced type of anisotropy, either intrinsic (e.g., fiber reinforced, collagen fibers, etc.) or induced (e.g., anisotropic damage).
Since both phenomena can be constitutively modeled by structural tensors, future work should include this additional tensorial argument.
It would be interesting to study whether the iCANN helps to identify the degree of anisotropy, e.g. the ratio between microvoids and microcracks.
In addition, structural tensors combined with multiplicative decomposition require push-forward operations of the structural tensors to the intermediate configuration.
This mapping is not unique, see \cite{sansour2007} and \cite{holthusen2023} for an overview of possible choices.
This raises the question of whether iCANNs are capable of deciding which mapping is best for certain types of materials.

\textbf{How to include multiphysics?} Up to now, we only have taken mechanical influences into account.
However, almost all engineering material behave temperature dependent, e.g., by thermal expansion and/or thermal softening.
To account for this phenomenon, we suggest to use a multiplicative decomposition of the deformation gradient into a thermal part and a mechanical part (see \cite{stojanovic1964} and \cite{vujosevic2002}).
In addition, the influence of electrochemical fields is of great importance in manufacturing (see \cite{dorfmann2005}, \cite{wulfinghoff2023}, and \cite{vandervelden2021}). 
Moreover, the diffusion of hormones and nutrients plays an important role in the growth and remodeling of living organisms (see \cite{manjunatha2022}).
Hence, it is highly disable to extend the CANN and iCANN approaches to multiphysical problems.
The question is how to find the most general formulation of our iCANN to account for all the interactions between the different fields at the material point level.

\textbf{How to set up experimental investigations?} We have shown that we need less experimental data to discover a model explaining our data compared to the identification of material parameters used in constitutive modeling.
The long-term perspective of iCANN is that by design any inelastic effects are included. 
This will allow the iCANN to automatically learn the inelastic effects hidden in the data and provide us with information about the micromechanical processes.
Up to now, such an identification requires complex experimental investigations, e.g. several cyclic tests, relaxation tests etc.
In view of our results, the question arises whether and how experiments have to be conducted in order to generate a maximum of information for the iCANN with as little effort as possible.
\section{Conclusion and outlook}
\label{sec:conclusion}
The inelastic Constitutive Artificial Neural Networks (iCANN) architecture provides a new family 
of neural networks for the prediction of inelastic material behavior under finite deformations. 
This architecture represents a consistent extension of the original Constitutive Artificial Neural Network (CANN) design and 
inherits its advantages over classical neural network architectures being used for constitutive 
modeling. The overall design of the iCANN is based on sound kinematic and thermodynamic considerations.
It combines well-known principles of continuum mechanics with the power of modern machine learning 
procedures to provide a neural network architecture that does not only offer a-priori thermodynamic 
consistency, objectivity, rigid motion of the reference configuration, independence of the rotational non-uniqueness or polyconvexity but also provides a flexible, modular and 
interpretable machine learning algorithm. Based on the multiplicative decomposition of the 
deformation gradient, this framework uses individual subnets to approximate both, the general 
Helmholtz free energy as well as an pseudo potential function. This idea enables the network 
to flexibly adjust to the data provided, and therefore, represents a general formulation to explain various kinds of inelastic phenomena.
For illustrative purposes, we chose the example of finite visco-elasticity within this publication.
We were able to demonstrate that the iCANN is capable to predict this kind of rate-dependent, inelastic material response properly with a high degree of accuracy, even though the amount of experimental data is sparse.
Due to its general design, the iCANN should be extended in future investigations to inequality constraint material behavior (e.g. elasto-plasticity and damage), equality constrained behavior (e.g. biological growth), multiphysics (e.g. thermoelasticity), and initial anisotropy (structural tensors).
The nature of the inelastic Constitutive Artificial Neural Network enables us to achieve automated model discovery for inelastic constitutive modeling, which is easily accessible now. 
Ideally, this technology might introduce a new age in the field of constitutive modeling away from less flexible user-defined models, which already include assumptions about the inelastic phenomena involved, to automated model selection on the basis of the experimental data provided.

\newpage
\appendix
\section{Appendix}
%
\subsection{Discovered weights for artificially generated data}
\label{app:weights_artificial}
\begin{table}[H]
\centering
\caption{Discovered weights for the Helmholtz free energy feed-forward network. The weights belong to the results for the artificially generated data in Section~\ref{ex:artificial}. The order of the weights corresponds to the numerical implementation provided.}
\label{tab:weights_psi_analytical}
\begin{tabular}{ l r}
\hline
					&	$\psi^{\mathrm{Neq}}$	\\
\hline
\hline
	$w_{1,1}^\psi$	&	0.16197191	\\
	$w_{1,3}^\psi$	&	0	\\
	$w_{1,2}^\psi$	&	0	\\
	$w_{1,4}^\psi$	&	0	\\
	$w_{3,1}^\psi$	&	4.4753417e-33	\\
	$w_{2,1}^\psi$	&	4.5402040e+00	\\
	$w_{2,5}^\psi$	&	9.2217451e-01	\\
	$w_{2,3}^\psi$	&	0	\\
	$w_{2,7}^\psi$	&	0	\\
	$w_{2,2}^\psi$	&	2.2400634e+00	\\
	$w_{2,6}^\psi$	&	0	\\
	$w_{2,4}^\psi$	&	1.5364066e-33	\\
	$w_{2,8}^\psi$	&	1.8214491e-33	\\
	$w_{3,2}^\psi$	&	0	\\
\hline	
\end{tabular}%
\end{table}
\begin{table}[H]
\centering
\caption{Discovered weights for the pseudo potential feed-forward network. The weights belong to the results for the artificially generated data in Section~\ref{ex:artificial}. The order of the weights corresponds to the numerical implementation provided.}
\label{tab:weights_g_analytical}
\begin{tabular}{ l r}
\hline
						&	$g$	\\
\hline
\hline
	$w_{1,1}^g$			&	0	\\
	$w_{1,3}^g$			&	0	\\
	$\tilde{w}_{1,5}^g$	&	0	\\
	$w_{2,1}^g$			&	0	\\
	$w_{2,4}^g$			&	0	\\
	$\tilde{w}_{2,7}^g$	&	0.00107837	\\
	$w_{2,2}^g$			&	0	\\
	$w_{2,5}^g$			&	0	\\
	$w_{2,8}^g$			&	0	\\
\hline	
\end{tabular}%
\end{table}
%
%
\subsection{Discovered weights for VHB 4910 polymer}
\label{app:weights_hossain}
\begin{table}[H]
\centering
\caption{Discovered weights for the Helmholtz free energy feed-forward network. The weights belong to the results for the VHB 4910 polymer in Section~\ref{ex:vhb}. The order of the weights corresponds to the numerical implementation provided.}
\label{tab:weights_psi_hossain}
\begin{tabular}{ l r r r r}
\hline
					&	$\psi^{\mathrm{Eq}}$		&	$\psi^{\mathrm{Neq}}_1$	&	$\psi^{\mathrm{Neq}}_2$	&	$\psi^{\mathrm{Neq}}_3$	\\
\hline
\hline
	$w_{1,1}^\psi$	&	0.08989155	&	0.6867937		&	0.06954185		&	2.9327118	\\
	$w_{1,3}^\psi$	&	0.3871473	&	0.6118265		&	0.28366846		&	2.6145873	\\
	$w_{1,2}^\psi$	&	0			&	0.06561667		&	0				&	0.03622768	\\
	$w_{1,4}^\psi$	&	0.01116255	&	0.21422595		&	0.01034374		&	0.17840806	\\
	$w_{3,1}^\psi$	&	-			&	3.9390977e-33	&	-2.5810559e-33	&	-1.908444e-33	\\
	$w_{2,1}^\psi$	&	2.646479		&	1.0183624		&	2.3602471		&	1.0819899	\\
	$w_{2,5}^\psi$	&	3.1516218	&	0.70885456		&	2.8352766		&	1.2915097	\\
	$w_{2,3}^\psi$	&	0			&	0.35119042		&	0				&	0.21051936	\\
	$w_{2,7}^\psi$	&	0.6164215	&	0.35955966		&	0.1741371		&	0.2551231	\\
	$w_{2,2}^\psi$	&	1.4593441	&	1.894548			&	1.1949594		&	3.15422	\\
	$w_{2,6}^\psi$	&	2.9215317	&	1.354211			&	2.65872			&	2.8406792	\\
	$w_{2,4}^\psi$	&	0.0537339	&	0.1450241		&	0.08411078		&	0.09968195	\\
	$w_{2,8}^\psi$	&	0.24964914	&	0.32216933		&	0.2144338		&	0.23061126	\\
	$w_{3,2}^\psi$	&	-			&	0				&	0				&	0	\\
\hline	
\end{tabular}%
\end{table}
\begin{table}[H]
\centering
\caption{Discovered weights for the pseudo potential feed-forward network. The weights belong to the results for the VHB 4910 polymer in Section~\ref{ex:vhb}. The order of the weights corresponds to the numerical implementation provided.}
\label{tab:weights_g_analytical}
\begin{tabular}{ l r r r}
\hline
						&	$g_1$	&	$g_2$	&	$g_3$	\\
\hline
\hline
	$w_{1,1}^g$			&	0			&	0			&	0	\\
	$w_{1,3}^g$			&	0			&	0			&	0	\\
	$\tilde{w}_{1,5}^g$	&	0			&	0			&	0	\\
	$w_{2,1}^g$			&	0			&	0			&	0	\\
	$w_{2,4}^g$			&	0			&	0			&	0	\\
	$\tilde{w}_{2,7}^g$	&	0.02321647	&	0.00174507	&	0.08019841	\\
	$w_{2,2}^g$			&	0			&	0			&	0	\\
	$w_{2,5}^g$			&	0			&	0			&	0	\\
	$w_{2,8}^g$			&	0			&	0			&	0	\\
\hline	
\end{tabular}%
\end{table}
%
%
%
\subsection{Discovered weights for passive skeletal muscle}
\label{app:weights_vanloocke}
\begin{table}[H]
\centering
\caption{Discovered weights for the Helmholtz free energy feed-forward network. The weights belong to the best results for passive skeletal muscle in Section~\ref{ex:muscle}, i.e., the third column in Figure~\ref{fig:TrainOne}. The order of the weights corresponds to the numerical implementation provided.}
\label{tab:weights_psi_vanloocke_train}
\begin{tabular}{ l r r r r}
\hline
					&	$\psi^{\mathrm{Eq}}$		&	$\psi^{\mathrm{Neq}}_1$	&	$\psi^{\mathrm{Neq}}_2$	&	$\psi^{\mathrm{Neq}}_3$	\\
\hline
\hline
	$w_{1,1}^\psi$	&	0.10420806	&	0.01166395     	&	0.00115344		&	0.01122695 \\
	$w_{1,3}^\psi$	&	0.1841228	&	0.01309146     	&	0.00347792		&	0.01664283 \\
	$w_{1,2}^\psi$	&	0.02560516	&	0.00111786	  	&	0.00250525   	&	0.00101571 \\
	$w_{1,4}^\psi$	&	0.21253704	&	0.0157883	  	&	0.6823113		&	0.01311576 \\
	$w_{3,1}^\psi$	&	-			&	5.497631e-34    &	-2.4178903e-14	&	-1.6557697e-05 \\
	$w_{2,1}^\psi$	&	0.06608981	&	0.04277665     	&	0   				&	4.2320956e-02 \\
	$w_{2,5}^\psi$	&	0.0637252	&	0.03449954	  	&	0   				&	3.4831800e-02 \\
	$w_{2,3}^\psi$	&	0.06350973	&	0.01193809	  	&	0.02947876   	&	1.5011105e-02 \\
	$w_{2,7}^\psi$	&	0.07190274	&	0.01611285	  	&	0.06952783   	&	1.8784763e-02 \\
	$w_{2,2}^\psi$	&	0.09108008	&	0.01163169     	&	0   				&	1.1191454e-02 \\
	$w_{2,6}^\psi$	&	0.14969026	&	0.01267746     	&	0				&	1.6345050e-02 \\
	$w_{2,4}^\psi$	&	0.02551797	&	0     			&	0.0014553		&	7.9535537e-05 \\
	$w_{2,8}^\psi$	&	0.20174257	&	0	  			&	0.52773875   	&	0 \\
	$w_{3,2}^\psi$	&	-			&	0     			&	0				&	0 \\
\hline	
\end{tabular}%
\end{table}
\begin{table}[H]
\centering
\caption{Discovered weights for the pseudo potential feed-forward network. The weights belong to the best results for passive skeletal muscle in Section~\ref{ex:muscle}, i.e., the third column in Figure~\ref{fig:TrainOne}. The order of the weights corresponds to the numerical implementation provided.}
\label{tab:weights_g_vanloocke_train}
\begin{tabular}{ l r r r}
\hline
						&	$g_1$	&	$g_2$	&	$g_3$	\\
\hline
\hline
        $w_{1,1}^g$			&	0 			& 0 		& 0 \\
        $w_{1,3}^g$			&	0 			& 0 		& 0 \\
        $\tilde{w}_{1,5}^g$	&	0 			& 0 		& 0 \\
        $w_{2,1}^g$			&	4.5470802e-06 	& 0 		& 0 \\
        $w_{2,4}^g$			&	7.6929213e-13 	& 0 		& 0 \\
        $\tilde{w}_{2,7}^g$	&	1.7541333e-01 	& 0.5692788 		& 0.15919787 \\
        $w_{2,2}^g$			&	0 	& 0 		& 0 \\
        $w_{2,5}^g$			&	0 	& 0 		& 0 \\
        $w_{2,8}^g$			&	0 	& 0 		& 0 \\
\hline	
\end{tabular}%
\end{table}
\begin{table}[H]
\centering
\caption{Discovered weights for the Helmholtz free energy feed-forward network. The weights belong to the best results for passive skeletal muscle in Section~\ref{ex:muscle}, i.e., the second column in Figure~\ref{fig:TrainFour}. The order of the weights corresponds to the numerical implementation provided.}
\label{tab:weights_psi_vanloocke_test}
\begin{tabular}{ l r r r r}
\hline
					&	$\psi^{\mathrm{Eq}}$		&	$\psi^{\mathrm{Neq}}_1$	&	$\psi^{\mathrm{Neq}}_2$	&	$\psi^{\mathrm{Neq}}_3$	\\
\hline
\hline
	$w_{1,1}^\psi$	&	0.00138299	&	0     			&	0	&	6.044407e-06 \\
	$w_{1,3}^\psi$	&	0.03045796	&	0     			&	0	&	0 \\
	$w_{1,2}^\psi$	&	0.07063455	&	0	  			&	0   &	0 \\
	$w_{1,4}^\psi$	&	0.09885824	&	7.708453e-10	  	&	0.00558791	&	0 \\
	$w_{3,1}^\psi$	&	-			&	3.6285916e-33    &	3.733226e-33	&	3.766066e-33 \\
	$w_{2,1}^\psi$	&	0.02312493	&	3.51029038e-02     &	0.02298807   &	8.7625727e-02 \\
	$w_{2,5}^\psi$	&	0.04258661	&	1.12825185e-02	  &	0.01470857   &	5.3523790e-02 \\
	$w_{2,3}^\psi$	&	0.11215075	&	2.31756195e-01	  &	0.02367271   &	8.5964095e-04 \\
	$w_{2,7}^\psi$	&	0.12233058	&	5.45360267e-01	  &	0.04010505   &	8.0012449e-04 \\
	$w_{2,2}^\psi$	&	0.00133567	&	0     			&	0   &	5.9973318e-06 \\
	$w_{2,6}^\psi$	&	0.02934347	&	0     			&	0	&	0 \\
	$w_{2,4}^\psi$	&	0.0702935	&	0     			&	0	&	0 \\
	$w_{2,8}^\psi$	&	0.0983757	&	7.71025022e-10	  &	0.00552198   &	0 \\
	$w_{3,2}^\psi$	&	-			&	0     			&	0	&	0 \\
\hline	
\end{tabular}%
\end{table}
\begin{table}[H]
\centering
\caption{Discovered weights for the pseudo potential feed-forward network. The weights belong to the best results for passive skeletal muscle in Section~\ref{ex:muscle}, i.e., the second column in Figure~\ref{fig:TrainFour}. The order of the weights corresponds to the numerical implementation provided.}
\label{tab:weights_g_vanloocke_test}
\begin{tabular}{ l r r r}
\hline
						&	$g_1$	&	$g_2$	&	$g_3$	\\
\hline
\hline
        $w_{1,1}^g$			&	0 	& 0 		& 0 \\
        $w_{1,3}^g$			&	0 	& 0 		& 0 \\
        $\tilde{w}_{1,5}^g$	&	0 	& 0 		& 0 \\
        $w_{2,1}^g$			&	0 	& 0 		& 0 \\
        $w_{2,4}^g$			&	4.6914302e-11 	& 0 		& 0 \\
        $\tilde{w}_{2,7}^g$	&	7.3394459e-01 	& 0.24998124 		& 0.29360896 \\
        $w_{2,2}^g$			&	0 	& 0 		& 0 \\
        $w_{2,5}^g$			&	0 	& 0 		& 0 \\
        $w_{2,8}^g$			&	0 	& 0 		& 0 \\
\hline	
\end{tabular}%
\end{table}
%
%
\subsection{Experimental data of polymer VHB 4910}
\label{app:data_hossain}
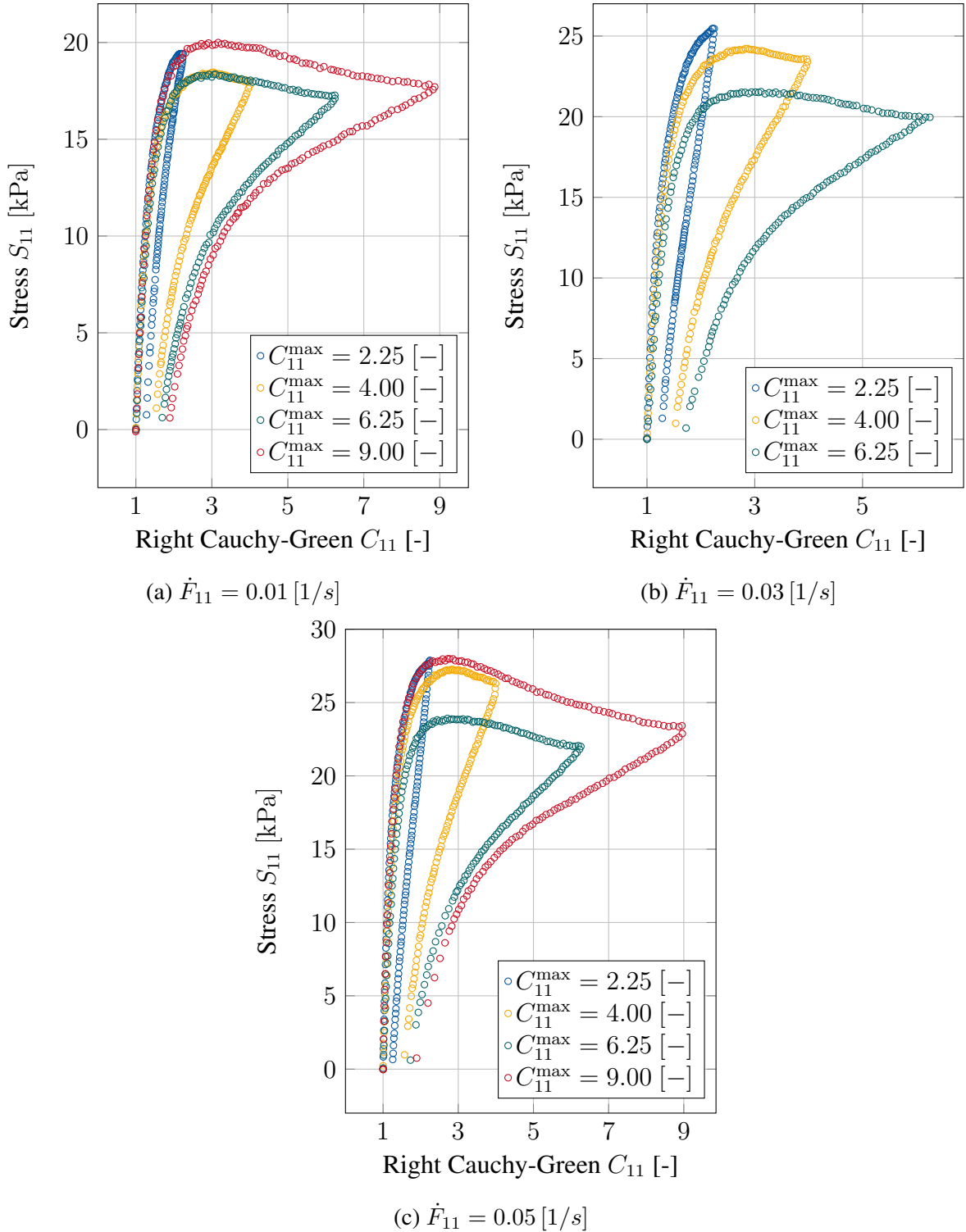
\begin{figure}[H]
	\centering
	\begin{subfigure}[]{0.49\textwidth}
		\begin{tikzpicture}
\begin{axis} [grid = major,
			legend pos = south east,
			legend cell align=left,
    			legend entries={Data, Training},
    			xmin = 0,
    			ymin=-3,
    			ymax=22,
    			xtick={1,3,5,7,9},
    			xlabel = {Right Cauchy-Green $C_{11}$ [-]},
    			ylabel = {Stress $S_{11}\ \lbrack \rm kPa \rbrack$},
    			legend entries={{$C_{11}^{\mathrm{max}}=2.25\ [-]$},{$C_{11}^{\mathrm{max}}=4.00\ [-]$},{$C_{11}^{\mathrm{max}}=6.25\ [-]$},{$C_{11}^{\mathrm{max}}=9.00\ [-]$}},
    			width=0.96\textwidth,
    			height=1.2\textwidth  ,
    			/pgf/number format/1000 sep={}			
		]
    		 \addplot[color = rwth1,fill = white,
    		 		fill opacity=0.1, 
    		 		only marks, mark size=1.5pt,
				error bars/.cd,
				y dir = both,
				y explicit ,
				error bar style={line width=0.5pt}, 
				error mark options={line width=0.5pt, mark size=2pt, rotate=90}
				] table[x expr=\thisrowno{2}, y expr=\thisrowno{5}] {graphs/section06/Hossain2012/Fig3_1_rate1_zsm.txt};	
    		 \addplot[color = rwth2,fill = white,
    		 		fill opacity=0.1, 
    		 		only marks, mark size=1.5pt,
				error bars/.cd,
				y dir = both,
				y explicit ,
				error bar style={line width=0.5pt}, 
				error mark options={line width=0.5pt, mark size=2pt, rotate=90}
				] table[x expr=\thisrowno{2}, y expr=\thisrowno{5}] {graphs/section06/Hossain2012/Fig3_2_rate1_zsm.txt};
    		 \addplot[color = rwth3,fill = white,
    		 		fill opacity=0.1, 
    		 		only marks, mark size=1.5pt,
				error bars/.cd,
				y dir = both,
				y explicit ,
				error bar style={line width=0.5pt}, 
				error mark options={line width=0.5pt, mark size=2pt, rotate=90}
				] table[x expr=\thisrowno{2}, y expr=\thisrowno{5}] {graphs/section06/Hossain2012/Fig4_1_rate1_zsm.txt};
    		 \addplot[color = rwth4,fill = white,
    		 		fill opacity=0.1, 
    		 		only marks, mark size=1.5pt,
				error bars/.cd,
				y dir = both,
				y explicit ,
				error bar style={line width=0.5pt}, 
				error mark options={line width=0.5pt, mark size=2pt, rotate=90}
				] table[x expr=\thisrowno{2}, y expr=\thisrowno{5}] {graphs/section06/Hossain2012/Fig4_2_rate1_zsm.txt};
\end{axis}
\end{tikzpicture}
		\caption{$\dot{F}_{11}=0.01\left[1/s\right]$}
		\label{fig:data_hossain_1}
	\end{subfigure}
	\begin{subfigure}[]{0.49\textwidth}
		\begin{tikzpicture}
\begin{axis} [grid = major,
			legend pos = south east,
			legend cell align=left,
    			legend entries={Data, Training},
    			xmin = 0,
    			ymin=-3,
    			ymax=27,
    			xtick={1,3,5,7,9},
    			xlabel = {Right Cauchy-Green $C_{11}$ [-]},
    			ylabel = {Stress $S_{11}\ \lbrack \rm kPa \rbrack$},
    			legend entries={{$C_{11}^{\mathrm{max}}=2.25\ [-]$},{$C_{11}^{\mathrm{max}}=4.00\ [-]$},{$C_{11}^{\mathrm{max}}=6.25\ [-]$}},
    			width=0.96\textwidth,
    			height=1.2\textwidth  ,
    			/pgf/number format/1000 sep={}			
		]
    		 \addplot[color = rwth1,fill = white,
    		 		fill opacity=0.1, 
    		 		only marks, mark size=1.5pt,
				error bars/.cd,
				y dir = both,
				y explicit ,
				error bar style={line width=0.5pt}, 
				error mark options={line width=0.5pt, mark size=2pt, rotate=90}
				] table[x expr=\thisrowno{2}, y expr=\thisrowno{5}] {graphs/section06/Hossain2012/Fig3_1_rate2_zsm.txt};	
    		 \addplot[color = rwth2,fill = white,
    		 		fill opacity=0.1, 
    		 		only marks, mark size=1.5pt,
				error bars/.cd,
				y dir = both,
				y explicit ,
				error bar style={line width=0.5pt}, 
				error mark options={line width=0.5pt, mark size=2pt, rotate=90}
				] table[x expr=\thisrowno{2}, y expr=\thisrowno{5}] {graphs/section06/Hossain2012/Fig3_2_rate2_zsm.txt};
    		 \addplot[color = rwth3,fill = white,
    		 		fill opacity=0.1, 
    		 		only marks, mark size=1.5pt,
				error bars/.cd,
				y dir = both,
				y explicit ,
				error bar style={line width=0.5pt}, 
				error mark options={line width=0.5pt, mark size=2pt, rotate=90}
				] table[x expr=\thisrowno{2}, y expr=\thisrowno{5}] {graphs/section06/Hossain2012/Fig4_1_rate2_zsm.txt};
\end{axis}
\end{tikzpicture}
		\caption{$\dot{F}_{11}=0.03\left[1/s\right]$}
		\label{fig:data_hossain_2}
	\end{subfigure}
	\begin{subfigure}[]{0.49\textwidth}
		\begin{tikzpicture}
\begin{axis} [grid = major,
			legend pos = south east,
			legend cell align=left,
    			legend entries={Data, Training},
    			xmin = 0,
    			ymin=-3,
    			ymax=30,
    			xtick={1,3,5,7,9},
    			xlabel = {Right Cauchy-Green $C_{11}$ [-]},
    			ylabel = {Stress $S_{11}\ \lbrack \rm kPa \rbrack$},
    			legend entries={{$C_{11}^{\mathrm{max}}=2.25\ [-]$},{$C_{11}^{\mathrm{max}}=4.00\ [-]$},{$C_{11}^{\mathrm{max}}=6.25\ [-]$},{$C_{11}^{\mathrm{max}}=9.00\ [-]$}},
    			width=0.96\textwidth,
    			height=1.2\textwidth  ,
    			/pgf/number format/1000 sep={}			
		]
    		 \addplot[color = rwth1,fill = white,
    		 		fill opacity=0.1, 
    		 		only marks, mark size=1.5pt,
				error bars/.cd,
				y dir = both,
				y explicit ,
				error bar style={line width=0.5pt}, 
				error mark options={line width=0.5pt, mark size=2pt, rotate=90}
				] table[x expr=\thisrowno{2}, y expr=\thisrowno{5}] {graphs/section06/Hossain2012/Fig3_1_rate3_zsm.txt};	
    		 \addplot[color = rwth2,fill = white,
    		 		fill opacity=0.1, 
    		 		only marks, mark size=1.5pt,
				error bars/.cd,
				y dir = both,
				y explicit ,
				error bar style={line width=0.5pt}, 
				error mark options={line width=0.5pt, mark size=2pt, rotate=90}
				] table[x expr=\thisrowno{2}, y expr=\thisrowno{5}] {graphs/section06/Hossain2012/Fig3_2_rate3_zsm.txt};
    		 \addplot[color = rwth3,fill = white,
    		 		fill opacity=0.1, 
    		 		only marks, mark size=1.5pt,
				error bars/.cd,
				y dir = both,
				y explicit ,
				error bar style={line width=0.5pt}, 
				error mark options={line width=0.5pt, mark size=2pt, rotate=90}
				] table[x expr=\thisrowno{2}, y expr=\thisrowno{5}] {graphs/section06/Hossain2012/Fig4_1_rate3_zsm.txt};
    		 \addplot[color = rwth4,fill = white,
    		 		fill opacity=0.1, 
    		 		only marks, mark size=1.5pt,
				error bars/.cd,
				y dir = both,
				y explicit ,
				error bar style={line width=0.5pt}, 
				error mark options={line width=0.5pt, mark size=2pt, rotate=90}
				] table[x expr=\thisrowno{2}, y expr=\thisrowno{5}] {graphs/section06/Hossain2012/Fig4_2_rate3_zsm.txt};
\end{axis}
\end{tikzpicture}
		\caption{$\dot{F}_{11}=0.05\left[1/s\right]$}
		\label{fig:data_hossain_3}
	\end{subfigure}
\caption{Experimental data of polymer VHB 4910 provided by \cite{hossain2012}. The material is subjected to uniaxial loading-unloading at different maximum stretch levels, $C_{11}^{\mathrm{max}}$. For constant loading rates, $\dot{F}_{11}$, but different maximum stretch levels applied, the material's response during loading should be the same.}
\label{fig:data_hossain}
\end{figure}
\section{Declarations}
%
\subsection{Acknowledgements}
Hagen Holthusen would like to thank the Heinrich Hertz-Stiftung for a scholarship that enabled a research stay at Stanford University. 
The foundation had no role in study design, collection, analysis and interpretation of data, in writing the report, or in the decision to submit the article for publication.
Tim Brepols and Stefanie Reese gratefully acknowledge financial support of the projects 453596084 and 453715964 by Deutsche Forschungsgemeinschaft.
Ellen Kuhl acknowledges funding by NSF CMMI Award 2320933 Automated Model Discovery for Soft Matter.
%
%
\subsection{Conflict of interest}
The authors of this work certify that they have no affiliations with or involvement in any organization or entity with any financial interest (such as honoraria; educational grants; participation in speakers’ bureaus; membership, employment, consultancies, stock ownership, or other equity interest; and expert testimony or patent-licensing arrangements), or non-financial interest (such as personal or professional relationships, affiliations, knowledge or beliefs) in the subject matter or materials discussed in this manuscript.
%
\subsection{Availability of data and material}
Our data is available at \cite{iCANN_code} (\url{https://doi.org/10.5281/zenodo.10066805})
%
\subsection{Code availability}
Our source code and examples are available at \cite{iCANN_code} (\url{https://doi.org/10.5281/zenodo.10066805})
%
\subsection{Contributions by the authors}
\textbf{Hagen Holthusen:} Conceptualization, Methodology, Theoretical description, Software, Data Curation, Formal analysis, Investigation, Visualization, Writing – original draft, Writing – review \& editing.\\
\textbf{Lukas Lamm:} Software, Data Curation, Writing – original draft, Writing – review \& editing.\\
\textbf{Tim Brepols:} Funding acquisition, Writing – original draft, Writing – review \& editing.\\
\textbf{Stefanie Reese:} Funding acquisition.\\
\textbf{Ellen Kuhl:} Funding acquisition, Methodology, Writing – original draft, Writing – review \& editing.

\newpage
\clearpage
\bibliographystyle{abbrvnat_modHH}
\bibliography{00_library_HH}

\end{document}